\documentclass[10pt,twocolumn,letterpaper]{article}

\usepackage[pagenumbers]{cvpr} 
\usepackage[toc,page]{appendix} 
\usepackage{titletoc}
\usepackage[dvipsnames]{xcolor}

\definecolor{cvprblue}{rgb}{0.21,0.49,0.74}
\usepackage[pagebackref,breaklinks,colorlinks,allcolors=cvprblue]{hyperref}
\usepackage{graphicx}
\usepackage{float}  
\usepackage{lipsum}
\usepackage{mathtools}
\usepackage{placeins}
\usepackage{algorithm}
\usepackage{algpseudocode}
\usepackage{nicefrac}
\usepackage{csquotes}
\usepackage[accsupp]{axessibility}
\DeclarePairedDelimiter{\ceil}{\lceil}{\rceil}

\usepackage{multirow}
\usepackage{multicol}

\usepackage[subtle, mathdisplays=tight, charwidths=normal, leading=normal]{savetrees}

\hypersetup{colorlinks,citecolor={cvprblue},urlcolor={cvprblue}, linkcolor={Maroon}} 

\usepackage{amsmath,amsfonts,bm}

\def\eqref#1{equation~\ref{#1}}

\def\ceil#1{\lceil #1 \rceil}

\def\1{\bm{1}}

\def\rr{{\textnormal{r}}}

\def\vvr{{\vec{r}}}

\def\vv{{\vec{v}}}

\def\vx{{\vec{x}}}

\DeclareMathAlphabet{\mathsfit}{\encodingdefault}{\sfdefault}{m}{sl}
\SetMathAlphabet{\mathsfit}{bold}{\encodingdefault}{\sfdefault}{bx}{n}

\def\gL{{\mathcal{L}}}

\def\gQ{{\mathcal{Q}}}

\def\gU{{\mathcal{U}}}

\def\sR{{\mathbb{R}}}

\usepackage{amsmath}
\usepackage{amssymb}
\usepackage{mathtools}
\usepackage{amsthm}
\usepackage{bbm}

\usepackage{pifont}

-lmtk10

\theoremstyle{plain}

\theoremstyle{definition}

\theoremstyle{remark}

\usepackage{dsfont}

\newcommand{\mat}[1]{\ensuremath{{\mathbf{\MakeUppercase{{#1}}}}}}
\renewcommand{\vec}[1]{\ensuremath{\mathbf{\MakeLowercase{{#1}}}}}

\newcommand{\Rm}{\mat{R}}

\newcommand{\Vm}{\mat{V}}

\def\gL{{\mathcal{L}}}

\def\gQ{{\mathcal{Q}}}

\def\gU{{\mathcal{U}}}

\def\sR{{\mathbb{R}}}

\newcommand{\xv}{\vec{x}}

\newcommand{\Nt}{\mathrm{N}}
\newcommand{\Dt}{\mathrm{D}}
\newcommand{\Kt}{\mathrm{K}}
\newcommand{\Wt}{\mathrm{W}}
\newcommand{\Ht}{\mathrm{H}}

\newcommand{\Vt}{\mathrm{V}}

\newcommand{\Mt}{\mathrm{M}}

\newcommand{\pr}{{\rm p}}
\newcommand{\xr}{{\rm x}}
\renewcommand{\rr}{{\rm r}}

\def\paperID{9013} 
\def\confName{CVPR}
\def\confYear{2025}

\begin{document}

\newcommand{\initials}{HMAR\xspace}
\title{\initials: Efficient \underline{H}ierarchical \underline{M}asked \underline{A}uto-\underline{R}egressive Image Generation}

\author{Hermann Kumbong $^{1,2}$\textsuperscript{*}\quad 
Xian Liu$^{2,3}$\textsuperscript{*}\quad 
Tsung-Yi Lin$^{2}$\quad 
Ming-Yu Liu$^{2}$\quad 
Xihui Liu$^{4}$\quad \\
Ziwei Liu$^{5}$\quad
Daniel Y. Fu$^{6, 7}$\quad 
Christopher R\'{e}$^{1}$\quad 
David W. Romero$^{2}$\\
$^1$Stanford University\quad 
$^2$NVIDIA\quad 
$^3$CUHK\quad 
$^4$HKU\quad 
$^5$NTU \quad 
$^6$UCSD \quad 
$^7$Together AI\\
}

\twocolumn[{
\vspace{-1cm}
\maketitle
\begin{center}
    \captionsetup{type=figure}
    \vspace{-7mm}
    \includegraphics[width=17.5cm, height=8cm]{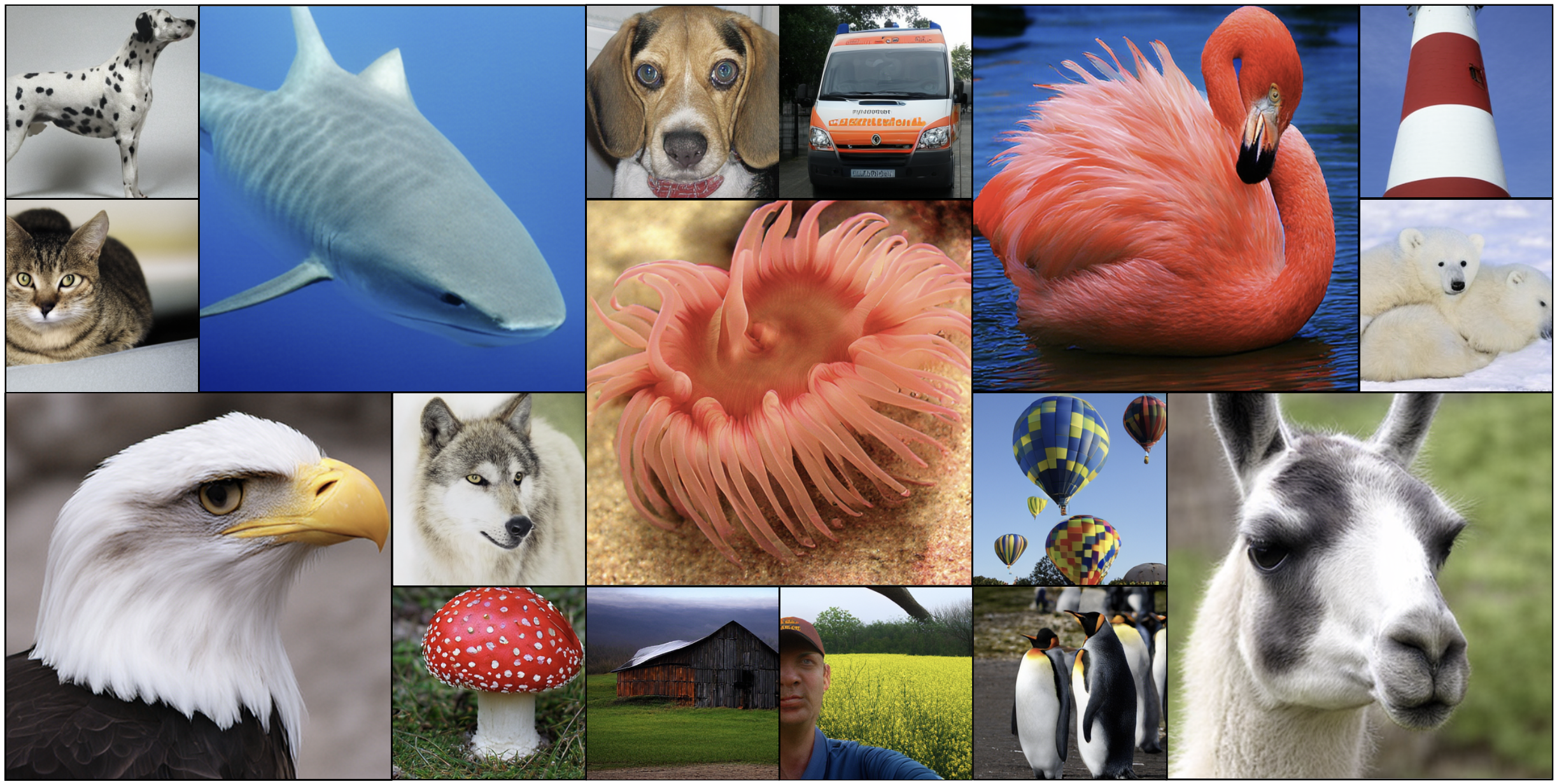}
    \vspace{-7mm}
    \captionof{figure}{\textbf{HMAR Samples}: Class-conditional ImageNet generated samples at $256\times256$ (\initials-$d30$) and $512\times512$ (\initials-$d24$) resolutions.}
    \label{fig:samples-banner}
\end{center}
}]
\maketitle
{
\renewcommand{\thefootnote}{\fnsymbol{footnote}}
\footnotetext[1]{Equal contribution.\hspace{1em}}
}

\setlength\abovedisplayskip{3pt}
\setlength\belowdisplayskip{3pt}

\begin{abstract}
Visual Auto-Regressive modeling (VAR) has shown promise in bridging the speed and quality gap between autoregressive image models and diffusion models. VAR reformulates autoregressive modeling by decomposing an image into successive resolution scales. During inference, an image is generated by predicting all the tokens in the next (higher-resolution) scale, conditioned on all tokens in all previous (lower-resolution) scales. However, this formulation suffers from reduced image quality due to the parallel generation of all tokens in a resolution scale; has sequence lengths scaling superlinearly in image resolution; and requires retraining to change the sampling schedule.

We introduce \underline{H}ierarchical \underline{M}asked \underline{A}uto\underline{R}egressive modeling (HMAR), a new image generation algorithm that alleviates these issues using next-scale prediction and masked prediction to generate high-quality images with fast sampling. HMAR reformulates next-scale prediction as a Markovian process, wherein the prediction of each resolution scale is conditioned only on tokens in its immediate predecessor instead of the tokens in all predecessor resolutions. When predicting a resolution scale, HMAR uses a controllable multi-step masked generation procedure to generate a subset of the tokens in each step. On ImageNet $256 {\times} 256$ and $512 {\times} 512$ benchmarks, HMAR models match or outperform parameter-matched VAR, diffusion, and autoregressive baselines. We develop efficient IO-aware block-sparse attention kernels that allow HMAR to achieve faster training and inference times over VAR by over $2.5\times$ and $1.75 \times$ respectively, as well as over $3 \times$ lower inference memory footprint.
Finally, HMAR yields additional flexibility over VAR; its sampling schedule can be changed without further training, and it can be applied to image editing tasks in a zero-shot manner.
\end{abstract}    
\vspace{-4mm}
\section{Introduction}
\label{sec:intro}

\begin{figure*}
 \centering
\begin{subfigure}[b]{0.3\textwidth}
\centering
\includegraphics[width=0.95\textwidth, height=4.1cm]{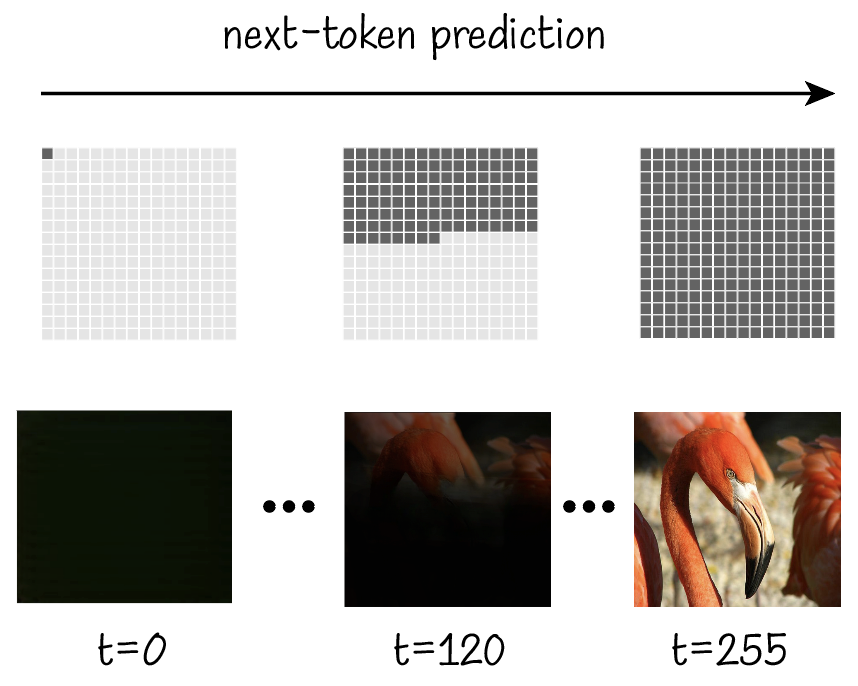}
\caption{Sequential generation -- VQ-GAN \citep{vq-gan}}
\label{fig:image1}
\end{subfigure}
\hfill
\begin{subfigure}[b]{0.3\textwidth}
\centering
\includegraphics[width=0.95\textwidth, height=4cm]{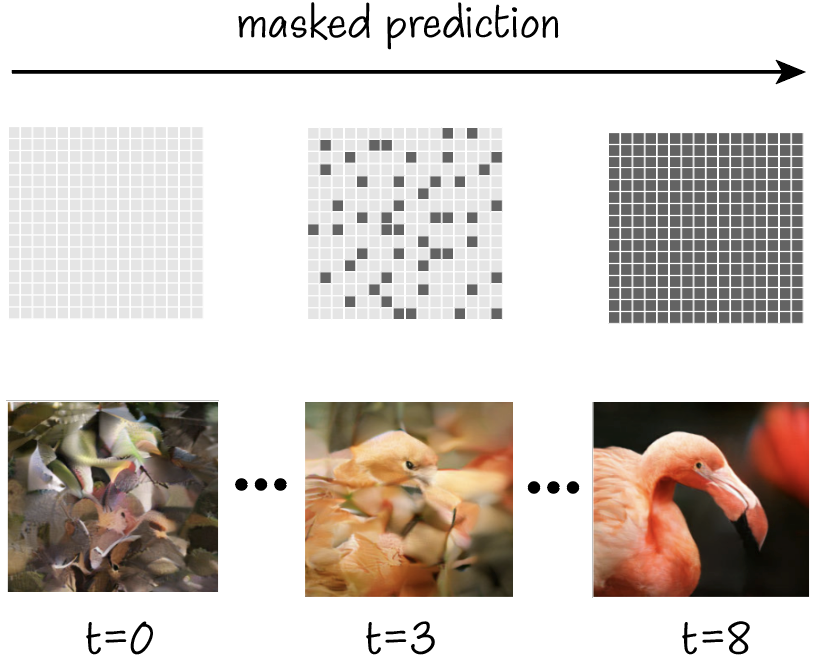}
\caption{Multi-step generation -- MaskGIT \citep{chang2022maskgit}}
\label{fig:image2}
\end{subfigure}
\hfill
\begin{subfigure}[b]{0.3\textwidth}
\centering
\includegraphics[width=0.95\textwidth, height=4.1cm]{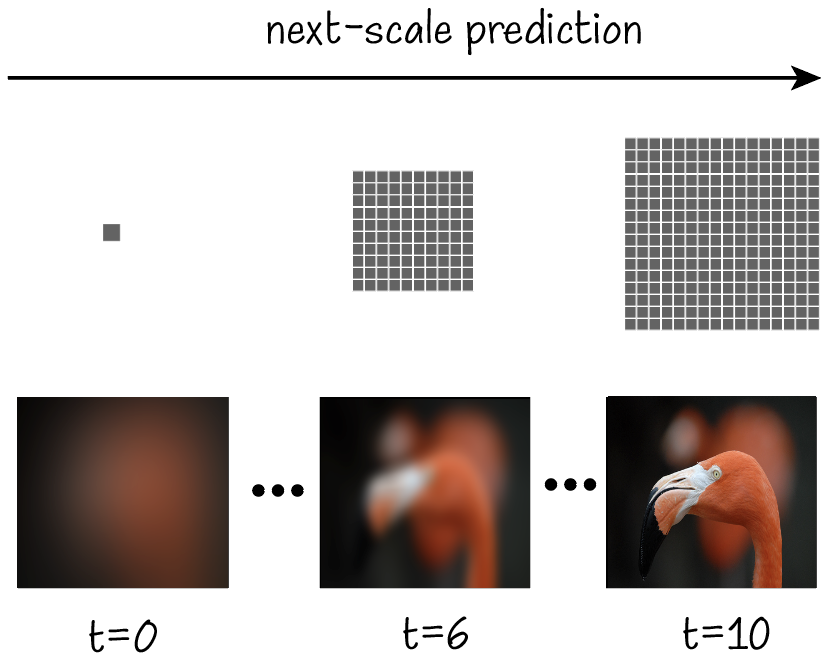}
\caption{Parallel multi-scale generation -- VAR \citep{tian2024visual}}
\label{fig:image3}
\end{subfigure}
\begin{subfigure}[b]{\textwidth}
        \centering
        \includegraphics[width=\textwidth, height=4.5cm]{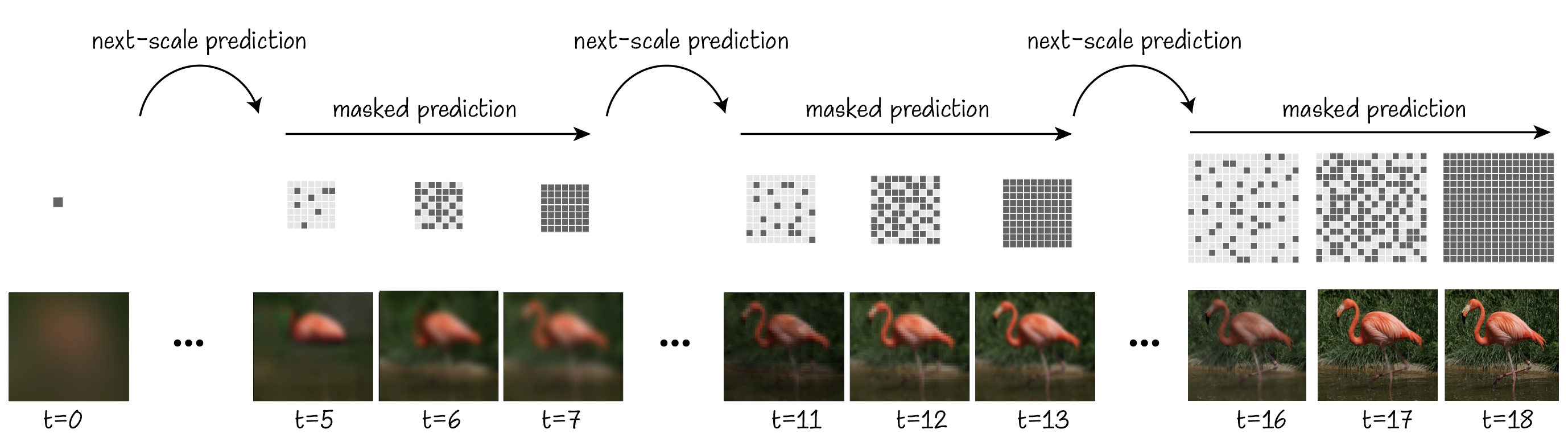}
        \caption{Hierarchical multi-step generation -- HMAR (Ours)}
\end{subfigure}
\vspace{-7mm}
\caption{\textbf{Illustration of the sequential decoding formulation in different methods.} We show the decoding process of next-token prediction \citep{llamagen, vq-gan}, parallel masked prediction \citep{chang2022maskgit}, next-scale prediction \citep{tian2024visual}, and our proposed hierarchical multi-step masked prediction. The dark and light grey squares represent the un-generated and generated tokens, respectively. HMAR generates images in an iterative two-step process by first producing a rough prediction of the next scale, then refining it using multi-step masked prediction until the final scale is reached.
\vspace{-4mm}}
\label{fig:method_banner}
\end{figure*}

Autoregressive modeling is the dominant approach for text generation \cite{achiam2023gpt, radford2018improving, radford2019language}. However, for images and videos, autoregressive models are yet to match diffusion models in speed and quality, making the latter the de-facto generative approach for these modalities \citep{song2020denoising, dhariwal2021diffusion, rombach2022high}.
This disparity raises the question of whether autoregressive models can match diffusion models in speed and quality for image generation.

Adapting the next-token autoregressive generation paradigm from language to images introduces multiple challenges. Images are multi-dimensional, making it difficult to determine an appropriate causal ordering. Orderings like raster-scan \citep{llamagen, pixelcnn, vq-gan} break the natural spatial relationships within images, resulting in lower-quality outputs. Additionally, sequential pixel-by-pixel generation becomes impractically slow, especially at high resolutions.
Masked autoregressive models, such as MaskGIT \citep{chang2022maskgit}, MAR \citep{mar}, and MAE \citep{li2023mage}, do not impose a strict order on the image and instead use global information to progressively fill an empty multi-dimensional canvas. However, the quality of their generation in practice still trails behind diffusion models, leaving diffusion as the preferred approach for image generation.

Recently, Visual Auto-Regressive modeling (VAR) \citep{tian2024visual} has shown promise in bridging the quality and speed gap between diffusion and autoregressive image models. VAR frames image generation as successive coarse-to-fine next-scale prediction over successively higher-resolution scales.
VAR generates higher-resolution scales by conditioning on the tokens across all previous lower-resolution scales. To make the autoregressive generation tractable, VAR generates all the tokens in a resolution scale in a single model iteration (as opposed to generating tokens one at a time).
As a result, VAR achieves faster sampling speeds than diffusion models and delivers the state-of-the-art image quality among autoregressive approaches \citep{tian2024visual}.

However, VAR still faces challenges in terms of achievable \textit{image quality}, \textit{efficiency} and \textit{flexibility}:
\begin{itemize}

    \item \textbf{Quality.} VAR accelerates generation by sampling all the tokens within a given scale in parallel. We hypothesize that this assumes \textit{all the tokens at a given scale are conditionally independent given all previous scales} and does not accurately capture the underlying joint distribution within each scale, which can cause inconsistencies within the same scale and error accumulation across scales, ultimately contributing to degraded sample quality (Fig. \ref{fig:error-accumulation}).
        
    \item \textbf{Efficiency.} Next-scale prediction conditioned on all previous scales leads to longer sequences---up to $5.84\times$ longer than next-token prediction at $256 \times 256$---which grow in both the input resolution and the number of scales (Fig.~\ref{fig:sequence-lengths}). This makes VAR more expensive to train at higher resolutions due to the quadratic time complexity of self-attention with sequence length. Furthermore, efficient self-attention libraries such as FlashAttention do not support the block-causal attention pattern (Fig.~\ref{fig:attn-mask}) in VAR. During inference, caching the lower-resolution scales increases the memory footprint and leads to out-of-memory issues at higher resolutions and larger model sizes.

    \item \textbf{Flexibility.} Next-scale prediction requires defining the number of sampling steps at training. As a result, increasing the number of sampling steps to improve image quality requires retraining the model from scratch with a new set of scales.
\end{itemize}

To address these issues, we introduce \underline{H}ierarchical \underline{M}asked \underline{A}uto\underline{R}egressive modeling (HMAR), a new image generation framework that combines next-scale prediction and masked prediction. HMAR reformulates next-scale prediction as a Markovian process, conditioning the generation of each successive resolution scale only on the tokens of its immediate predecessor (instead of all predecessor scales). The Markovian formulation enables a block-diagonal, windowed attention pattern (Fig.~\ref{fig:attn-mask}) during training, offering up to $5\times$ times more sparsity than VAR's block-causal pattern at $256\times256$. HMAR furthermore replaces the single-step scale generation of VAR with a controllable, multi-step masked generation procedure similar to MaskGIT \citep{chang2022maskgit}, thereby removing the per-scale conditional independence assumption of VAR. Finally, HMAR's hierarchical coarse-to-fine ordering allows reweighting of the training loss to focus the model's capacity on crucial image details at the most important hierarchy levels.

HMAR improves over VAR and autoregressive modeling in terms of \textit{image quality}, \textit{efficiency}, and \textit{flexibility}:

\begin{itemize}

    \item \textbf{Quality.} On ImageNet- $256\times256$ and ImageNet-$512\times512$ benchmarks, our parameter-matched HMAR models match or outperform VAR in FID while improving the Inception Score by up to $\approx30$ points. HMAR outperforms previous AR and Diffusion baselines (DiT) in FID and Inception Score. Qualitatively, HMAR enhances image quality over VAR~\citep{tian2024visual}.
        
    \item \textbf{Efficiency.} Due to its Markovian formulation, HMAR does not need to compute or cache any preceding-scale tokens, resulting in up to $1.75{\times}$ speedup and $3{\times}$ memory reduction during inference. In addition, the block-diagonal attention pattern enables $10\times$ faster attention computation via an I/O-aware window attention kernel. This results in up to $2.5{\times}$ faster end-to-end training time compared to VAR.

    \item \textbf{Flexibility.} The intra-scale masked generation procedure provides flexibility, allowing an increase in the number of sampling steps without retraining the model from scratch. Increasing masked sampling steps at coarser scales improves FID scores while increasing them at finer scales enhances perceptual image quality. HMAR's intra-scale masking makes it easy to adapt HMAR to image editing tasks like inpainting, outpainting, and class-conditional editing. 
\end{itemize}
The remainder of this paper is structured as follows: Section~\ref{sec:related_work_short} gives an abbreviated treatment of related work. Section~\ref{sec:preliminaries} discusses the necessary background. Section~\ref{sec:method} discusses the HMAR method. Section~\ref{sec:experiments} presents experiments. Section~\ref{sec:discussion} concludes and discusses future work. Additional details are provided in the supplementary material.

\section{Related Work}
\label{sec:related_work_short}
We provide an abbreviated discussion of related work. A full treatment is given in Appendix \ref{sec:related_work}.

Diffusion models \citep{sohl2015deep,ho2020denoising,dhariwal2021diffusion, song2020denoising,ramesh2022hierarchical, saharia2022photorealistic} are the dominant class of generative models for image synthesis and are trained to reverse a gradual noising process.
Autoregressive image generation models \citep{pixelcnn, pixelcnn++, pixelrnn} offer an alternative approach by generating images sequentially, typically following a raster scan pattern. Recent work has improved efficiency by using vector-quantized VAEs \citep{vq-gan, oord2018vqvae} to compress images into discrete tokens for autoregressive generation.
Masked image generative models \citep{chang2022maskgit, chang2023muse, li2023mage, weber2024maskbitembeddingfreeimagegeneration} use a masked prediction objective similar to BERT \citep{devlin2019bertpretrainingdeepbidirectional}. By predicting multiple masked tokens in parallel, these models achieve faster generation speeds compared to next-token autoregressive image models.
Visual autoregressive modeling (VAR) \citep{tian2024visual} enhances the efficiency and quality of autoregressive image generation by reframing autoregressive image generation as next-scale prediction instead of next-token prediction.
Finally, efficient attention implementations like FlashAttention \cite{shah2024flashattention, dao2022flashattention, dao2023flashattention2} compute self-attention efficiently on GPU but only support a limited number of attention patterns. Recent work such as FlexAttention \citep{flexattention2024} supports a wider range of attention patterns but currently restricts sequence lengths to multiples of $128$.

\section{Background}
\label{sec:preliminaries}
In this section, we discuss the necessary background on VAR that we build on. We first discuss image generation as next-token prediction and then image generation as next-scale prediction. We then discuss the tokenization scheme used in VAR, which we also adopt in HMAR.
\newline
\noindent\textbf{Image generation as next-token prediction.} An image 
is represented as a sequence of $\Nt$ discrete tokens $\vx = (\xr_1, \xr_2, \ldots, \xr_\Nt)$, flattened according to a specified order, \textit{e.g.}, raster-scan. Each token $\xr_n$ is an integer from a vocabulary of size $\Vt$ and corresponds to a vector in a codebook $\Vm {\in} \sR^{\Vt {\times} \Dt}$ with latent dimension $\Dt$. The probability of the image, $\pr(\xv)$, is then modeled as:
\begin{equation}
\label{eqn:chain_rule}
 \pr(\vx) = \prod_{t=1}^\Nt \pr(\xr_t | \xr_1, \xr_2, \ldots, \xr_{t-1}).
\end{equation}
Flattening an image into a one-dimensional sequence breaks the spatial relationships between neighboring pixels. Closely connected pixels are widely separated in the sequence, making it difficult to capture important local patterns. Moreover, the uni-directional ordering restricts the model's ability to leverage the full image context, resulting in reduced quality and limited flexibility. Finally, the number of required sampling steps grows linearly with image resolution, making high-resolution image generation computationally expensive and often impractical.

\noindent\textbf{Image generation as next-scale prediction.} Visual Auto-Regressive Modeling (VAR) \citep{tian2024visual}  overcomes the limitations of next-token autoregressive image generation by reformulating the task as next-scale rather than next-token prediction. In this approach, an image $\vx$ is decomposed into $\Kt$ sub-images of increasing resolutions $(\vvr_1, \vvr_2, ... \vvr_\Kt)$, and the likelihood is defined over the sequence of scales as:
\begin{equation}
\pr(\xv) = \pr\left(\vvr_1, \vvr_2, ..., \vvr_\Kt\right) = \prod_{k=1}^\Kt \pr\left(\vvr_k | \vvr_1, ..., \vvr_{k-1}\right).
\end{equation}
Each autoregressive step now generates a scale $\vvr_k$ containing $\Ht_k{\times}\Wt_k$ tokens and no flattening like raster-scan is needed. The full context of the image at the preceding scales is available for conditioning. Additionally, the number of autoregressive steps is now controlled by the number of scales, making this method far more scalable.  

A \textit{block-causal} mask (Fig.~\ref{fig:attn-mask}) is used during training to enforce causality across scales while preserving bidirectional dependencies among tokens within each scale. During inference, all tokens within a scale are sampled in parallel, conditioned on the tokens of all previous scales. This leads to fast sampling while providing good visual quality.


\noindent\textbf{Multi-scale vector quantization.}
In order to translate images from a continuous pixel space into a discrete token space, VAR uses a \textit{multi-scale residual quantization} method, where the sub-images $(\vvr_1, \vvr_2, ... \vvr_\Kt)$ progressively add information to a residual approximation of the image $\tilde{\xv}$, such that, after $\Kt$ stages, the approximation resembles the original image as faithfully as possible. VAR uses a VQ-VAE quantization method to quantize continuous vectors into discrete tokens.
In particular, VAR maps each of the $\xr_{i,j}$ values of the latent image representation $\xv$ to one of $\Vt$ learnable vectors $\vv {\in} \Vm$, $\Vm {\in} \sR^{\Vt {\times} \Dt}$ as:
\begin{equation}
    \tilde{\xr}_{i, j} = \gQ(\xr_{i,j}) = \left( \text{argmin}_{v \in [\Vt]} \left\| \Vm_{v, :} - \xr_{i, j} \right\|_2 \right).
\end{equation}

In VAR, the latent image representation is further interpolated across various resolutions corresponding to the scales, $k \in [\Kt]$. At each scale, the residual between the cumulative approximation and the original image is quantized and used as the token map for that level. The associated learnable vector $\vv$ is then used for reconstruction. Encoding and reconstruction in multi-scale vector quantization are depicted in Algs. \ref{algo:multi_scale_enc} and \ref{algo:multi_scale_rec}. We adopt the same approach in HMAR.
\vspace{0.15cm}
\begin{algorithm}[h]
\caption{Multi-scale VQ-VAE Encoding}
\label{algo:multi_scale_enc}
\begin{algorithmic}[1]
\State \textbf{Input:} Latent image representation $\vx$
\State \textbf{Parameters:} Steps $\Kt$, resolutions $\{(\Ht_k, \Wt_k)\}_{k=1}^\Kt$
\State \textbf{Output:} Multi-scale token maps $\Rm$
\State $\Rm = [\;]$
\For{$k = 1, \cdots, \Kt$}
    \State $\vvr_k = \mathcal{Q}\left(\mathtt{interpolate}(\vx, \Ht_k, \Wt_k)\right)$
    \State $\Rm\mathtt{.append}(\vvr_k)$
    \State $\tilde{\xv}_k = \mathtt{interpolate}\left(\mathtt{lookup}(\Vm, \vvr_k), \Ht_\Kt, \Wt_\Kt\right)$
    \State $\vx = \vx - \tilde{\xv}_k$
\EndFor
\State \textbf{return} $\Rm$
\end{algorithmic}
\end{algorithm}
\vspace{-0.025cm}
\begin{algorithm}[h]
\caption{Multi-scale VQ-VAE Reconstruction}
\label{algo:multi_scale_rec}
\begin{algorithmic}[1]
\State \textbf{Input:} Multi-scale token maps $\Rm$
\State \textbf{Parameters:} Steps $\Kt$, resolutions $\{(\Ht_k, \Wt_k)\}_{k=1}^\Kt$
\State \textbf{Output:} Latent image reconstruction $\tilde{\xv}$
\State $\tilde{\xv}_0 = 0$
\For{$k = 1, \cdots, \Kt$}
    \State $\vvr_k = \Rm\mathtt{[}k\mathtt{]}$
    \State $\tilde{\xv}_k = \mathtt{interpolate}\left(\mathtt{lookup}(\Vm, \vvr_k)\right), \Ht_\Kt, \Wt_\Kt)$
    \State $\tilde{\xv}_{1:k} = \tilde{\xv}_{1:k-1} + \tilde{\xv}_k$
\EndFor
\State \textbf{return} $\tilde{\xv}_{1:\Kt}$ 
\end{algorithmic}
\end{algorithm}

\section{Hierarchical Masked Image Generation}\label{sec:method}
In this section, we describe the key components of HMAR. Section \ref{sec:markovian} formulates next-scale prediction with a Markovian assumption, conditioned only on the tokens in the previous scale. We then develop GPU kernels to leverage the resultant block-sparse attention pattern during training. Section \ref{sec:masking} describes HMAR's intra-scale multi-step masked generation process. Section \ref{sec:balancing-scales} describes how HMAR focuses on more important resolution scales during training for higher quality. Finally, Section \ref{sec:hmar} describes the overall HMAR approach.

\subsection{Efficient Markovian Next-Scale Prediction}
\label{sec:markovian}
We reformulate next-scale prediction to be Markovian and develop an efficient, I/O-aware, block-sparse attention GPU kernel that enables faster training.
\newline
\noindent\textbf{Reformulating Next-Scale Prediction to be Markovian.}
In VAR, each resolution scale contains only residual information of the input (Alg.~\ref{algo:multi_scale_rec}, L7). Hence, next-scale prediction is conditioned on the tokens of all previous scales. However, this leads to longer sequences (Fig.~\ref{fig:sequence-lengths}) which are expensive for training and inference. We observe that the running image reconstruction up to the stage $k$, $\Tilde{\vx}_{1:k}$ 
(Alg.~\ref{algo:multi_scale_rec}, L8) \textit{contains the information from all stages up to the stage $k$}. Consequently, conditioning of the running reconstruction is equivalent to conditioning on all previous stages. That is $
\pr\left(\vvr_k | \vvr_1, \vvr_2,..., \vvr_{k-1}\right) = \pr(\vvr_k | \Tilde{\xv}_{1:k-1})$, and therefore, the likelihood of $\vx$ can be rewritten as:
\begin{equation}
\pr(\xv) = \pr\left(\vvr_1, \vvr_2, ..., \vvr_\Kt\right) = \prod_{k=1}^\Kt \pr\left(\vvr_k |\Tilde{\xv}_{1:k-1} \right).
\end{equation}
This equivalence depicts the Markovian nature of next-scale prediction akin to Laplacian and Gaussian pyramids \citep{adelson1980image, burt1987laplacian}. We note that only the conditioning changes in this formulation, and we still predict the residual tokens $\vvr_k$.

In Fig.~\ref{fig:attn-map}, we illustrate the attention pattern in VAR, revealing that the majority of attention is concentrated on the previous scale, which further validates our approach. 
In practice, we make use of the interpolation function used in Alg.~\ref{algo:multi_scale_enc},~L6, to map the running reconstruction $\Tilde{\xv}_{1:k-1}$ to an image of shape $\Ht_{k-1}{\times}\Wt_{k-1}$. This allows us to modify the attention pattern of VAR \citep{tian2024visual} from a lower block-triangular pattern to a block-diagonal pattern (Fig.~\ref{fig:attn-mask}) which is much more sparse. Furthermore, this formulation removes the need for prefix computations and KV-caching during inference, leading to faster inference and reduced inference memory usage.

\noindent\textbf{I/O-Aware Windowed Attention.} Our Markovian formulation \textit{theoretically} enables faster attention computation compared to the original VAR due to its higher sparsity (Fig.~\ref{fig:attn-mask}). However, leveraging this sparsity in practice is not straightforward.  Efficient attention implementations such as FlashAttention~\citep{dao2022flashattention, dao2023flashattention2, shah2024flashattention}, \textit{only support a handful of attention variants} of which our block-diagonal pattern and the original block-causal pattern in VAR are not among.

To address this, we develop a custom GPU kernel using Triton~\citep{tillet2019triton} that extends FlashAttention~\citep{dao2022flashattention, dao2023flashattention2, shah2024flashattention} to support these patterns. Our kernels further leverage the sparsity pattern to accelerate attention computation, leading to more than 10${\times}$ speed-up in attention computation. We provide additional details and micro-benchmarks in Appendix \ref{sec:efficient-attention}.

\subsection{Hierarchical Multi-Step Masked Generation}
\label{sec:masking}
We describe the quality impacts of VAR's single-step generation process for each resolution scale, and we describe the intra-scale multi-step masked generation in HMAR.

\noindent\textbf{Oversmoothing and Error Accumulation in VAR.}
VAR samples all tokens within a scale $\vvr_k$ in parallel given the previous scales from $\pr(\vvr_k | \vvr_{<k})$.
While this approach accelerates sampling, we hypothesize that it implicitly assumes that all tokens $\rr_k^{(i, j)}$ within a scale $k$ are conditionally independent given the previous scales $\vvr_{<k}$. That is, VAR implicitly models $\pr(\vvr_k | \vvr_{<k})$ as:
\begin{align}\label{eq:var_prob_approx}
    \pr(\vvr_k | \vvr_{<k}) = \pr\big(\rr_k^{(1, 1)} \big| \vvr_{<k} \big) ... \pr\big(\rr_k^{(\Ht_k, \Wt_k)} \big| \vvr_{<k}\big).
\end{align}
This is an approximation of the true joint distribution $\pr(\vvr_k | \vvr_{<k})$ given by the chain rule (Equation~\ref{eqn:chain_rule}).
We hypothesize that this is not a very accurate approximation of the underlying distribution and \enquote{oversmooths} the relationship between tokens in the same scale. Oversmoothing potentially degrades image quality, especially when dependencies between tokens are strong. We demonstrate this effect in (Fig~\ref{fig:error-accumulation}), showing how errors generated in earlier scales can propagate during generation to impact the image quality.

\noindent\textbf{Efficient modeling of intra-scale dependencies.} According to the chain rule, the mathematically correct way to model $\pr(\vvr_k | \vvr_{<k})$ entails sampling each token one by one at each scale. However, token-by-token sampling becomes intractable for next-scale prediction.
To strike an optimal trade-off between speed and quality, we instead make use of a \textit{multi-step masked generation} strategy similar to MaskGIT \citep{chang2022maskgit} at each scale. 

Given a number of masking steps $\Mt_k$ at scale $k$, we utilize an iterative process to sample a subset of tokens (at each scale) per step, such that after $\Mt_k$ steps, all the tokens at the corresponding scale are sampled. In HMAR, each step is conditioned on the tokens sampled so far at the current stage as well as the tokens from the previous stage. Formally, let $\vvr_k^m$ be the tokens at the scale $k$ after $m$ intra-scale sampling steps. The probability of the tokens at the scale $\pr(\vvr_k | \vvr_{<k})$ is given as:
\begin{align}
    \hspace{-4mm}\pr(\vvr_k | \vvr_{<k}) = \hspace{-1mm}\prod_{m=1}^\Mt \hspace{-1mm}\pr(\vvr_k^m | \vvr_k^1, ..., \vvr_k^{m-1}, \vvr_k^0, \vvr_{<k})\pr(\vvr_k^0 | \vvr_{<k}),
\end{align}
where $\vvr_k^0$ corresponds to the initial next-scale estimation of the VAR next-scale prediction. $\Mt_k$ offers an adjustable trade-off between quality and speed, where $\Mt_k {=} 0$ yields the VAR approximation in (\ref{eq:var_prob_approx}), and $\Mt_k {=} \Ht_{k}{\times}\Wt_{k}$ corresponds to next-token prediction at each scale. We demonstrate this in Fig.~\ref{fig:method_banner}. While this introduces additional sampling steps, our efficient reformulation of next-scale prediction allows it to still be efficient. 

\subsection{Training Dynamics}
\label{sec:balancing-scales}
The hierarchical generation process in HMAR, similar to Diffusion models, provides a unique advantage; it allows us to prioritize specific detail levels, allocating model capacity accordingly \citep{dieleman2024noise}. We motivate the need to balance the importance of different scales during training and how we achieve this in HMAR.

\noindent\textbf{Multi-Scale Training: Balancing Scale Contributions.}
VAR is trained by optimizing the cross-entropy loss across all tokens at all scales that make up the image. In VAR \citep{tian2024visual}, the loss is simply averaged across all tokens irrespective of the scale.
$\gL_\mathrm{train}$ is given by:
\begin{equation} \label{eq:balance_per_tokens}
\gL_\mathrm{train} =\frac{1}{N} \sum_{k=1}^{\Kt} \sum_{(i, j)}\mathcal{L}\big(\rr_{k}^{(i, j)}\big), 
\end{equation}
where $\mathcal{L}\big(\rr_{k}^{(i, j)}\big)$ denotes the cross-entropy loss for the $(i, j)$-th token at scale $k$ and $N$ is the total number of tokens.

However, this fails to take into account several considerations:  \textbf{1) Number of Tokens per Scale}. For VAR \citep{tian2024visual}, which employs $\Kt{=}10$ levels, the finest scale contributes $256$ times more than the coarsest scale. This imbalance leads the model to prioritize the finer scales, neglecting the coarse scales that capture the global image structure. \textbf{2) Learning Difficulty of each Scale}. We use the minimum test loss at each scale as an indicator of learning difficulty and illustrate in Fig.~\ref{fig:loss-distribution} that it approximately follows a log-normal distribution, suggesting that each level has varied difficulty and this should be incorporated in the learning algorithm. \textbf{3) Perceptual Importance of each Scale}. Each scale plays a distinct role in determining the perceptual quality of an image. Earlier scales focus on capturing the global structure, while later scales refine finer details. Moreover, errors introduced at earlier scales tend to propagate and accumulate during the generation process, emphasizing the critical importance of accurately capturing these early scales (Fig.~\ref{fig:error-accumulation}). 

\noindent\textbf{Loss Reweighting.} To leverage the above insights, we reweight the training loss to account for each scale as follows:
\begin{equation} \label{eq:balance_per_tokens}
\gL_\mathrm{train} = \sum_{k=1}^{\Kt} w(k)\sum_{(i, j)}\mathcal{L}\big(\rr_{k}^{(i, j)}\big), \quad 0 \leq w(k) \leq 1, \quad \sum_{k=1}^{\Kt} w(k) = 1
\end{equation}
We empirically experiment with different loss weighting functions in the Appendix.~\ref{sec:loss-weighting-ablation}. We find that the choice of weighting function significantly impacts quality (Table.~\ref{table:loss-reweighting}). Additionally, we find that a log-normal weighting function (Fig.~\ref{fig:loss-weighting}), which parallels the loss difficulty distribution (Fig.~\ref{fig:loss-distribution}), yields the best FID and Inception Score.

\begin{table*}[t]
\small
\centering
\begin{tabular}{l|l|cc|cc|cc}
\toprule
\textbf{Type} & \textbf{Model} & \textbf{FID} $\downarrow$ & \textbf{IS} $ \uparrow$&  \textbf{Precision} $\uparrow$ & \textbf{Recall} $\uparrow$ & \# \textbf{Params}  & \# \textbf{Steps} \\
\midrule
Diffusion & DiT-XL/2 \citep{dit} & 2.27 & 278.2 & 0.83 & 0.57 & 675M & 250 \\
\midrule
\multirow{3}{*}{Mask.} & MaskGIT \citep{chang2022maskgit} & 6.18 & 182.1 & 0.80 & 0.51 & 227M & 8 \\
 & MAR-L \citep{mar} & 2.35 & 227.8 & 0.79 & 0.62 & 943M & 256 \\
 & MAGE \citep{li2023mage} & 7.04 & 123.5 & -  & - & 439M & 20 \\
\midrule
\multirow{6}{*}{AR} & VQGAN \citep{vq-gan} & 15.8 & 74.3 & - & - & 1.4B & 256\\
 & Llamagen \citep{llamagen} & 2.81 & 263.3 & 0.81 & 0.58 & 3.1B & 256\\
 & VAR-$d16$ \citep{tian2024visual} & 3.36 & 277.8 & 0.84 & 0.51 & 310M & 10 \\
 & VAR-$d20$ \citep{tian2024visual}  & 2.67 & 304.4 & 0.84 & 0.55 & 600M & 10 \\
 & VAR-$d24$ \citep{tian2024visual}  & 2.15  & 312.4 & 0.82 & 0.58 & 1.0B & 10 \\
 & VAR-$d30$ \citep{tian2024visual} &  1.95  &  303.6 & 0.81 & 0.59 & 2.0B & 10 \\
\midrule
Hybrid AR & HART \citep{tang2024hart} & \textbf{1.77} & 330.3  & - & - & 2.0B & 10 \\
\midrule
\multirow{4}{*}{\textbf{HMAR (Ours)}} & HMAR-$d16$ & 3.01 & 288.6 & 0.84 & 0.55 & 465M & 14 \\
 & HMAR-$d20$ & 2.50 & 319.0 & \textbf{0.85} & 0.57 & 840M & 14 \\
 & HMAR-$d24$ & 2.10 & 324.3 & 0.83 & 0.60 & 1.3B & 14 \\
 & HMAR-$d30$ & 1.95 & \textbf{334.5} & 0.82& \textbf{0.62} & 2.4B & 14 \\
\bottomrule
\end{tabular}
\vspace{-2mm}
\caption{\textbf{Quantitative evaluation on class-conditional ImageNet $256\times 256$.} $\downarrow$ and $\uparrow$ indicate whether lower or higher values are better. We report numerical results on commonly used metrics of FID, IS, Precision, and Recall, which are comprehensive to cover generation quality and diversity. \# Steps indicate the number of model runs needed to generate an image. The $-d$ notation in VAR and HMAR indicates the number of layers in the model.
\vspace{-3mm}}
\label{table:imagenet-256}
\end{table*}

\subsection{The HMAR Architecture}\label{sec:hmar}

HMAR consists of two sub-modules: the next-scale prediction module and the intra-scale refining module. The next-scale model corresponds to a Markovian VAR model (Sec.~\ref{sec:markovian}), and the intra-scale refining module corresponds to a multi-step masked generation module as presented in Sec.~\ref{sec:masking}. The whole HMAR architecture is shown in Fig.~\ref{fig:method_banner}.
The remainder of this section describes the training and inference of HMAR.

\noindent\textbf{Training.} HMAR is trained in two steps. First, the next-scale prediction module is trained using an IO-aware windowed attention mask for each image, as described in Section \ref{sec:markovian}. Then, a finetuning step is started for the training of the intra-scale masked prediction module. To this end, we add a masked prediction head and finetune it with a masked prediction objective similar to MaskGIT \citep{chang2022maskgit}. 
In this phase, we uniformly sample a ratio $\gamma_k \sim \gU(0, 1)$, and randomly select $\ceil{\gamma \Ht_k \Wt_k}$ tokens from each $\vvr_k$ and replace them with a special $[\texttt{MASK}]$ token. Then, given the unmasked tokens, the model is trained to predict the value of the masked tokens at each scale. We find that using the same masking ratio $\gamma_k = \gamma$ across scales leads to more stable training.

Let $\gamma_k\vvr_k$ and $\bar{\gamma}_k\vvr_k$ depict the masked and unmasked tokens at a scale $k$. Then, the intra-scale refining module is trained to minimize the cross entropy of the masked tokens given the unmasked tokens. That is:
\begin{equation}
    \gL_{\mathrm{mask}} = \sum_{k=1}^\Kt \gL(\gamma_k\vvr_k | \bar{\gamma}_k\vvr_k) = \sum_{k=1}^\Kt \gL(\gamma\vvr_k | \bar{\gamma}\vvr_k) 
\end{equation}
We condition on both the unmasked tokens within that scale and the accumulated reconstruction of the image from previous scales. Doing so allows us to preserve all the incoming information from the next-scale module during refinement, which gives us image generation of higher quality. 

\noindent\textbf{Inference.} Just as for training, HMAR follows a two-stage process during generation as well. First, we iteratively obtain a coarse estimation of the next scale using the next-scale prediction module, and then we iteratively refine these predictions using the intra-scale masked refinement module. At this point, we generate the initial tokens based only on the estimations of the next-scale module, and then we mask out some of them and then generate them again, conditioning on the accumulated reconstruction of the image and the unmasked tokens at that scale.
\begin{figure*}[t]
  \centering
  \includegraphics[width=17.5cm]{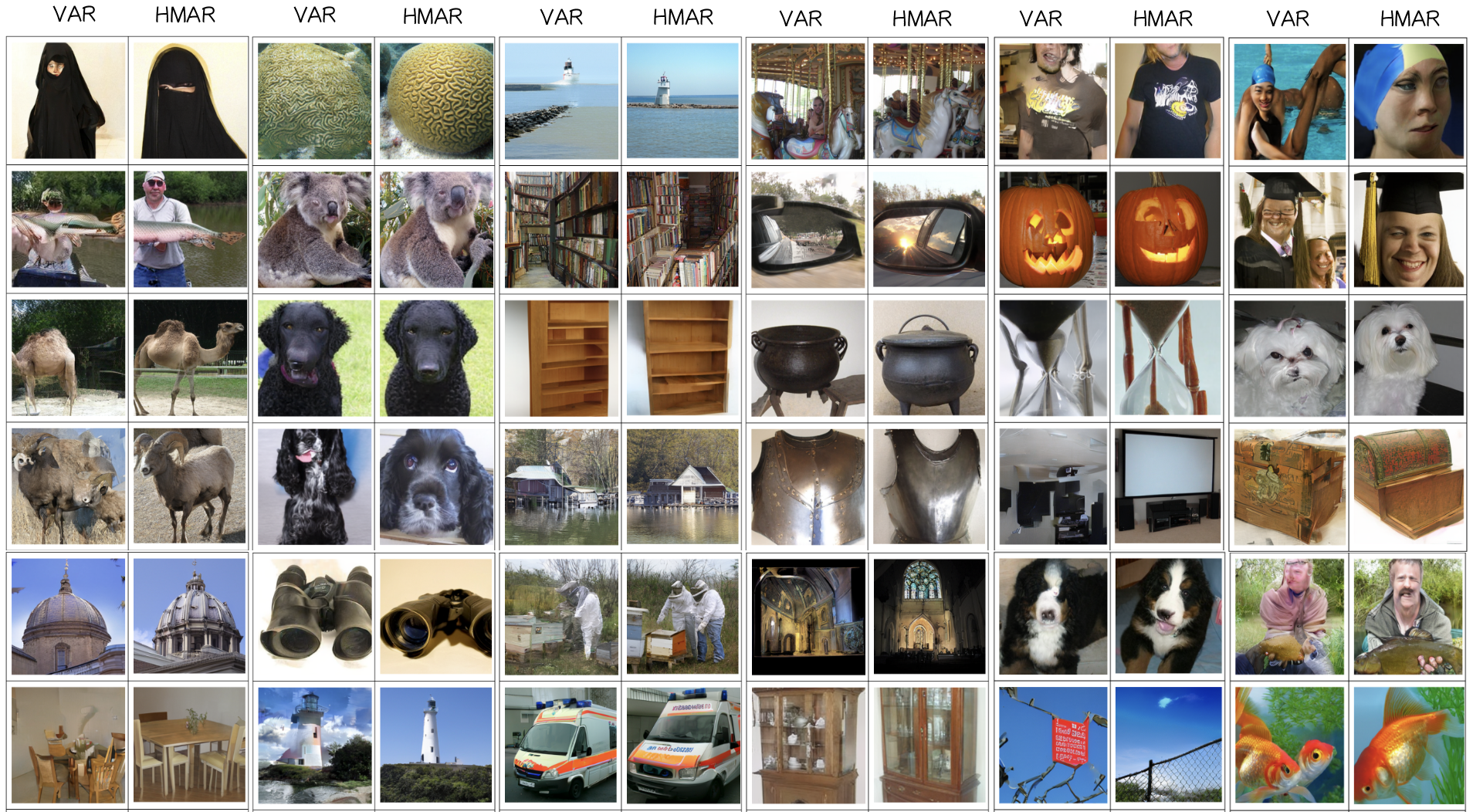}
  \vspace{-6mm}
  \caption{\textbf{Visual Comparisons of Samples from VAR-$d16$ and HMAR-$d16$}. Selected samples highlighting how HMAR's multi-step generation at each scale can enhance image quality compared to using only \textit{next-scale} prediction in VAR. \vspace{-4mm}}
  \label{fig:visual-comparison}
\end{figure*}

\section{Experiments}\label{sec:experiments}
We evaluate HMAR on quality, efficiency, and flexibility. 
\newline
\textbf{Quality.} We evaluate HMAR on ImageNet $256\times256$ and $512\times512$ for class-conditional image generation. HMAR achieves better or comparable FID scores and significantly higher Inception Scores compared to VAR, AR, and diffusion baselines. We also provide qualitative analysis of generated samples.
\newline
\textbf{Efficiency.} We benchmark HMAR models for both training and inference efficiency, showing that HMAR achieves both faster training and inference than VAR, with the efficiency gains increasing as we scale to higher resolutions. 
\newline
\textbf{Flexibility.} We demonstrate HMAR's flexibility, showing that its sampling can be changed without any additional training to improve image quality, and it can be applied to image editing tasks like in-painting, out-painting, and class-conditional image editing. We end with an ablation study evaluating the effect of the individual components of HMAR on image quality.

\noindent\textbf{Experimental Setup.}
We align our experimental setup with VAR \citep{tian2024visual}. We train all our models from scratch with similar parameters and number of transformer layers as VAR. For each scale, we maintain consistency with VAR by adopting identical hyperparameters, number of scales, and training durations. For image tokenization, we employ the pre-trained multi-scale VQ-VAE tokenizer from VAR \citep{tian2024visual}. During the inference phase, we implement top-$k$ top-$p$ sampling. For comparison with VAR models, we utilize open-source pre-trained checkpoints for evaluation. We use the same setup to evaluate both efficiency and quality performance.

\begin{figure}
    \centering
    \begin{subfigure}[b]{0.49\textwidth}
        \centering
         \includegraphics[width=0.49\textwidth]{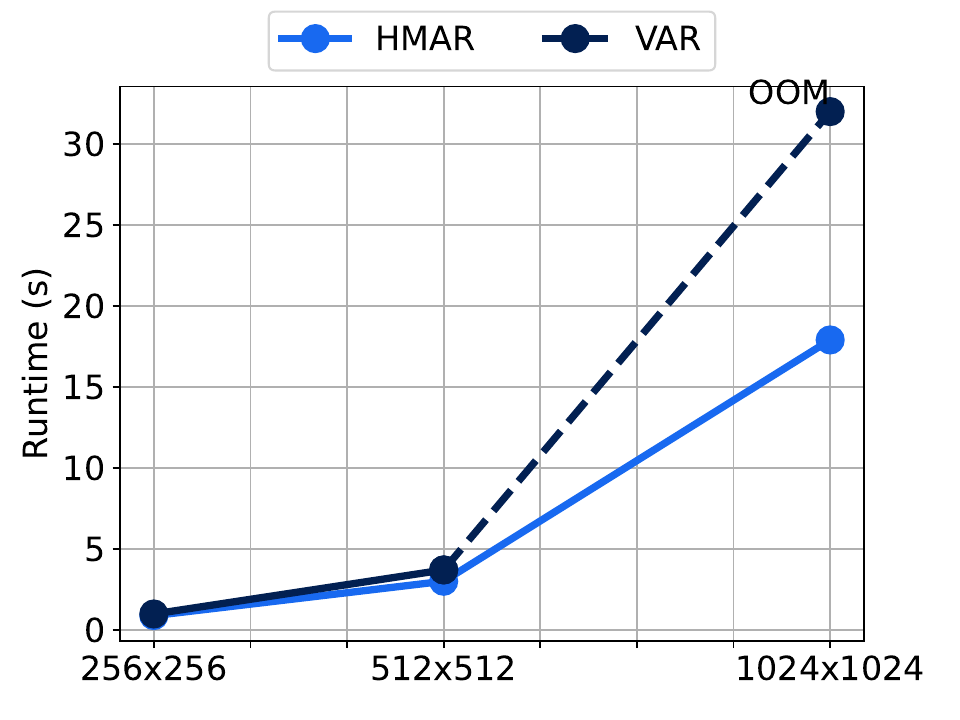}
        \includegraphics[width=0.49\textwidth]{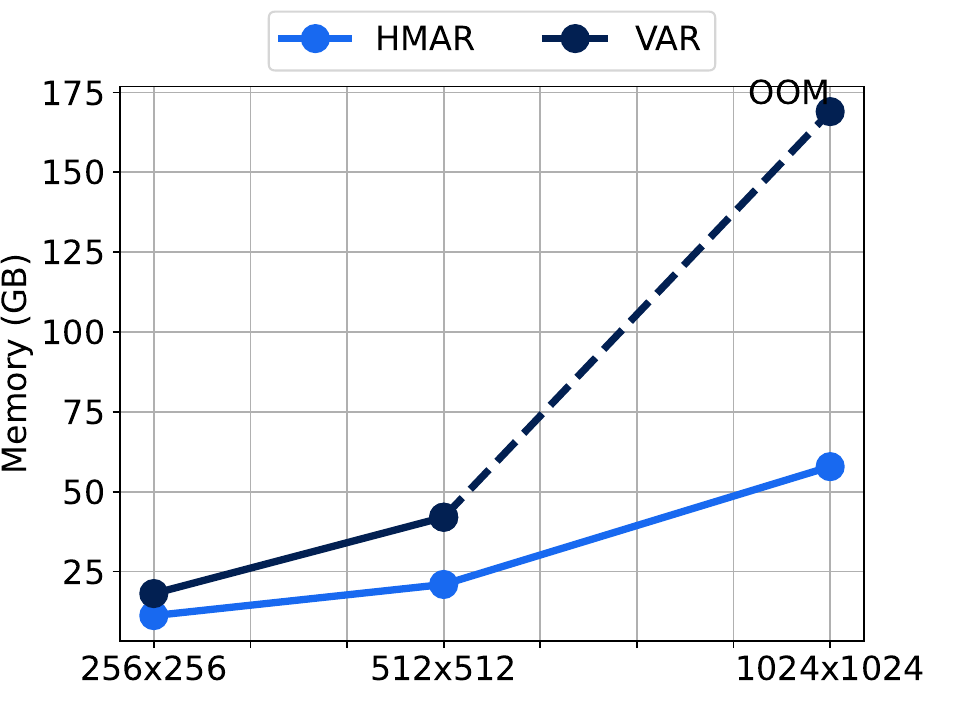}
        \caption{\textbf{Inference} Runtime and Memory Footprint vs Resolution,  d-$24$, bs = $16$}
        \label{fig:inference_memory_and_runtime}
    \end{subfigure}
    \hfill
   \begin{subfigure}[b]{0.49\textwidth}
        \centering
        \includegraphics[width=0.49\textwidth]{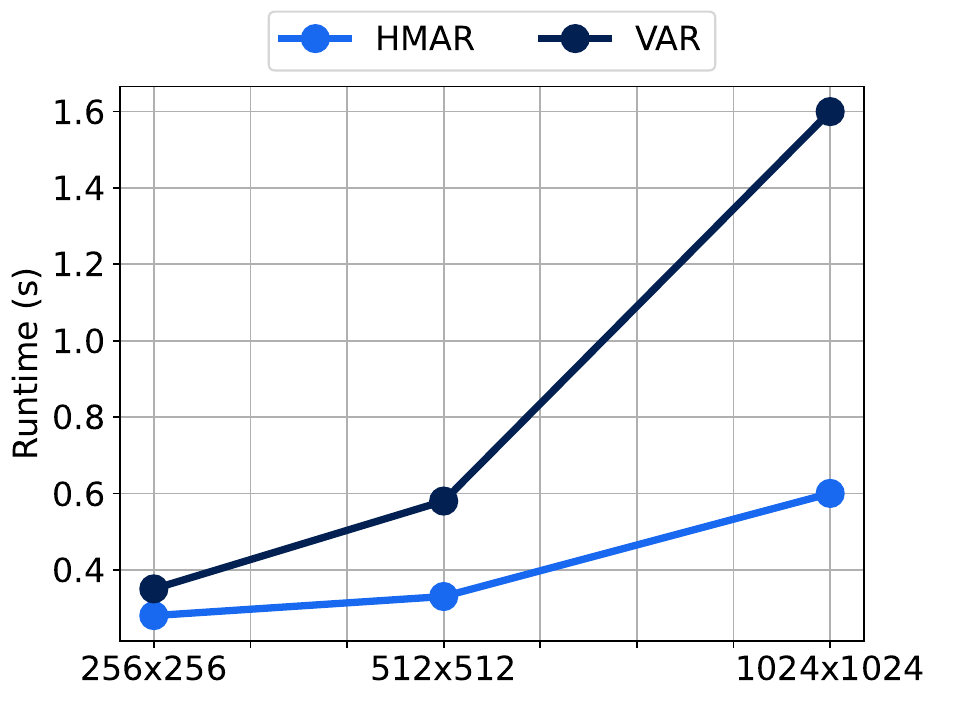}
        \includegraphics[width=0.49\textwidth]{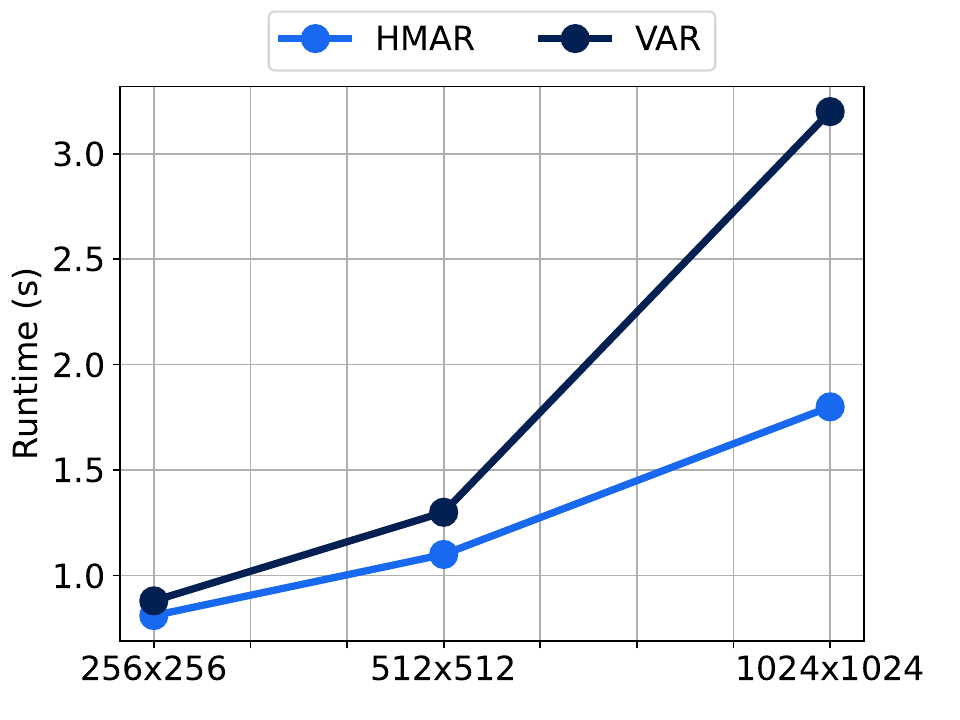}
        \caption{\textbf{Training} FWD (left) and BWD (right), d-$24$, largest bs at each resolution.}
        \label{fig:training_e2e}
    \end{subfigure}
    \vspace{-7mm}
    \caption{\textbf{Inference and Training Efficiency.} HMAR enables more efficient training and inference compared to VAR, with the efficiency gap becoming more pronounced as we scale to higher resolutions.
    \vspace{-4mm}}
    \label{fig:efficiency_analysis}
\end{figure}
\subsection{Quality}
In this section, we evaluate the quality of HMAR image generation using both quantitative metrics and qualitative analysis.

\noindent\textbf{Quantitative Metrics.}
We evaluate class-conditional image generation on ImageNet at $255\times256$ (Table~\ref{table:imagenet-256}) and $512\times512$ (Table ~\ref{table:imagenet-512}) resolutions. Using standard metrics (FID, Inception Score, Precision, and Recall), we find that HMAR consistently matches or outperforms baselines in FID scores while significantly surpassing them in Inception Score. This demonstrates HMAR's ability to generate high-quality, diverse images.

\begin{table}[h]
\centering
\begin{tabular}{l|l|ccc}
\toprule
\multicolumn{5}{c}{\textbf{ImageNet 512x512 Benchmark}}\\
\toprule
\textbf{Type} & \textbf{Model} & \textbf{FID} $\downarrow$ & \textbf{IS} $ \uparrow$ & \#\textbf{Para}  \\
\midrule
Diff. & DiT-XL/2 \citep{dit} & 3.04 & 240.8 & 675M \\
\midrule
Mask. AR & MaskGIT \citep{chang2022maskgit} & 7.32 & 156.0 & -  \\
Mask. AR & MAR-L \citep{mar} & 2.74 & 205.2  & 481M  \\
\midrule
VAR & VAR-$d36$ \citep{tian2024visual}  & 2.63 & 303.2 & 2.5B\\
\midrule
HMAR & HMAR-$d24$ & 2.99 & \textbf{304.1} & 1.3B\\
\bottomrule
\end{tabular} 
\caption{\textbf{ImageNet 512x512 Benchmark}. Due to limited computational resources, we train our HMAR model with $\approx2\times$ fewer parameters compared to VAR and find it to be competitive.} 
\label{table:imagenet-512}
\end{table}
\noindent\textbf{Qualitative Analysis.}
We show class conditional samples from HMAR on ImageNet $256\times256$ and $512\times512$ in Fig.~\ref{fig:samples-banner}. In Fig.~\ref{fig:visual-comparison}, we compare selected samples from HMAR against samples generated from VAR\citep{zhang2024var}. In Appendix \ref{sec:additional-qualitative-results}, we provide additional qualitative comparisons against other baselines, as well as additional samples from HMAR. Our results show that HMAR generates images with comparable or better visual quality compared to baseline methods.

\subsection{Efficiency}
\label{subsec:efficiency}
We benchmark the training speed, the inference speed, and the memory footprint of HMAR compared to VAR. All benchmarks are on a single A100 80GB and averaged over 25 repetitions.

\noindent\textbf{Training.} We benchmark the end-to-end runtime of HMAR (using our custom block diagonal attention kernel) and compare it against the VAR baseline (Fig. \ref{fig:training_e2e}). HMAR demonstrates consistently faster performance, with the speed advantage growing more pronounced at higher resolutions. At the $1024\times1024$ resolution, HMAR achieves a 2.5$\times$ end-to-end speedup over VAR. We provide additional micro-benchmarks on the performance of our attention implementation in Appendix \ref{sec:efficient-attention}.

\noindent\textbf{Inference.}
Fig.~\ref{fig:inference_memory_and_runtime} compares the inference runtime and memory footprint of our HMAR model to VAR. HMAR demonstrates faster inference, with the speed advantage increasing at higher resolutions, primarily due to avoiding prefix computations. The memory footprint of HMAR is significantly lower than VAR, which requires a KV-cache. This performance gap widens as we scale to higher resolutions and larger model sizes. 

\subsection{Flexibility}
\begin{figure}
    \centering
    \includegraphics[width=1.0\linewidth]{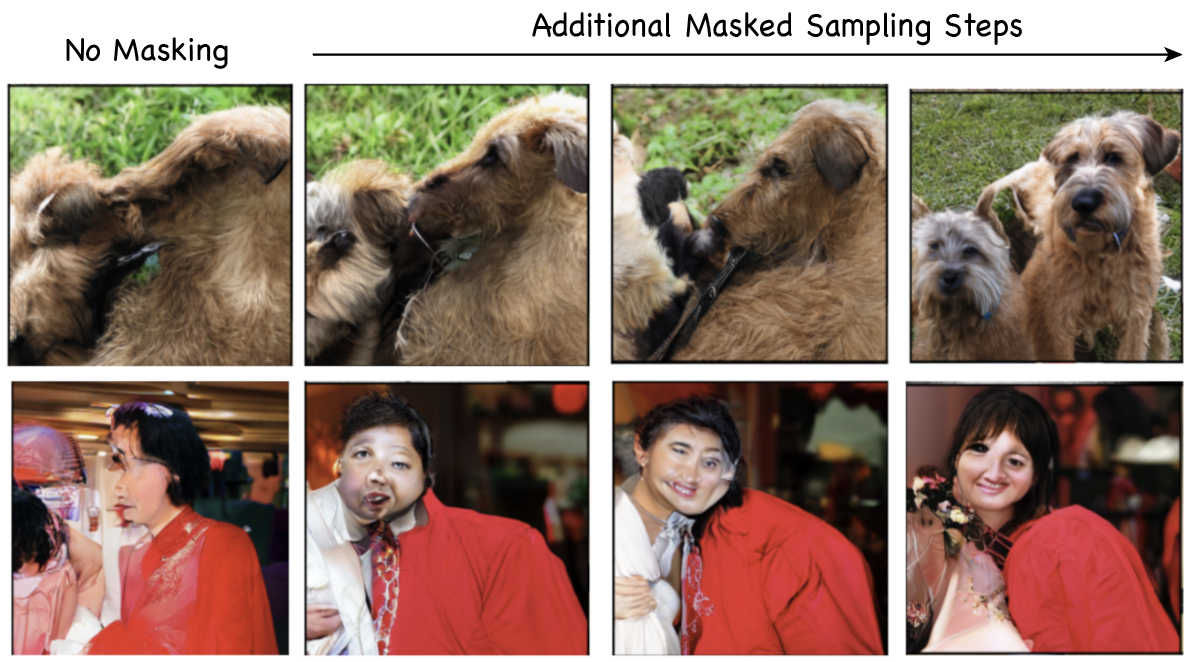}
    \caption{\textbf{Impact of Masking on Visual Quality HMAR-$d16$.} Increasing masked sampling steps can yield improved visual quality.}
    \label{fig:masking-qualitative}
\end{figure}

\begin{figure}
    \centering
    \includegraphics[width=\linewidth]{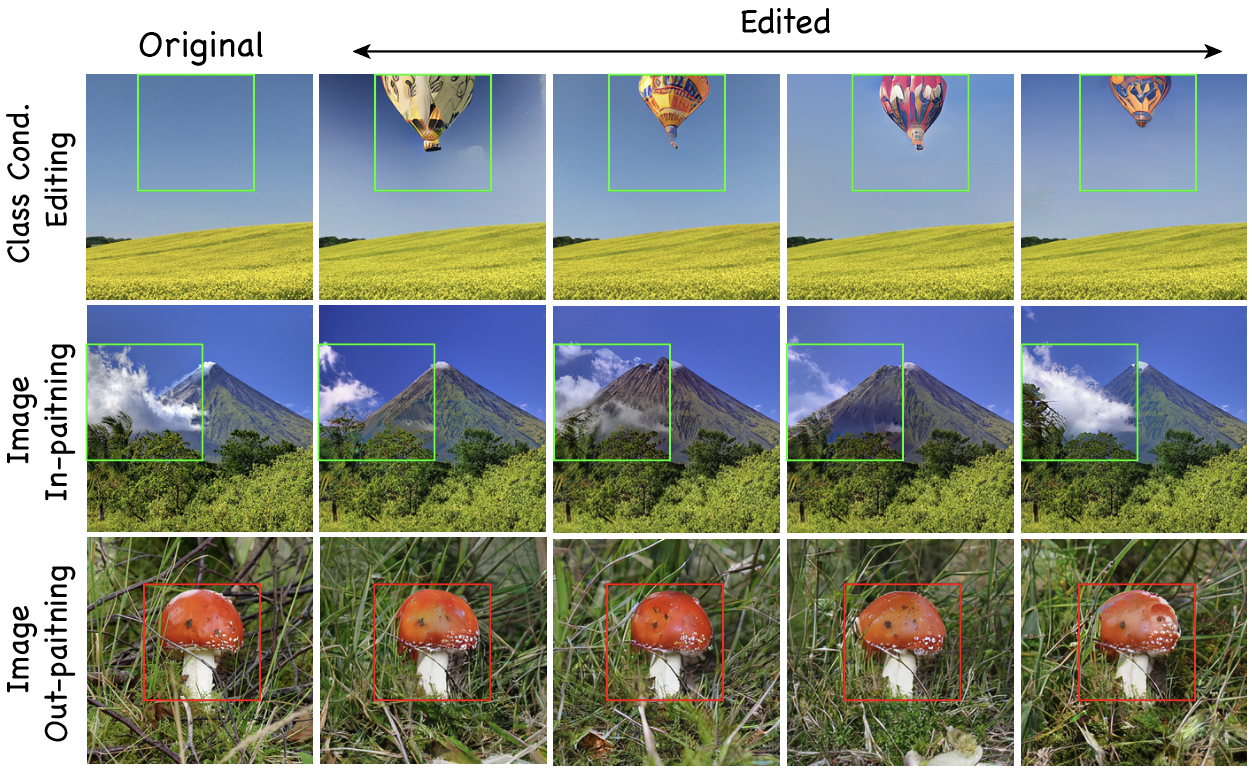}
   \caption{\textbf{Image Editing}. Applying HMAR zero-shot to editing tasks}
    \label{fig:zero-shot}
\end{figure}

In Fig.~\ref{fig:masking-qualitative}, we demonstrate how HMAR's flexible sampling strategy can help improve quality by increasing the number of sampling steps at inference time. In Fig. ~ref{fig:masking-quantitative} we show how increasing the number of sampling steps at inference time can improve the FID score.
We show HMAR's generalization to zero-shot image editing tasks in Fig.~\ref{fig:zero-shot}.
\subsection{Ablation Study}
We ablate the key components in HMAR and quantify their impact in Table \ref{table:ablation-methods}. In Appendix \ref{sec:loss-weighting-ablation}, we provide a detailed ablation on different loss-weighting choices. Fig.~\ref{fig:masking-quantitative} demonstrates that increasing the number of sampling steps through masking enhances the FID score in our HMAR-d$16$ model. We find that a few additional sampling steps at lower resolution scales improve the FID score; while additional steps at higher scales don't meaningfully improve FID, they can enhance visual quality, as illustrated in Figure \ref{fig:masking-qualitative}.

\begin{table}[!h]
\small
\centering
\begin{tabular}{l|l|cc|c}
\toprule
\textbf{Model} & \textbf{Method} & \textbf{FID} $\downarrow$ & \textbf{IS} $ \uparrow$ & \# \textbf{Steps}  \\
\midrule
\multirow{2}{*}{VAR-$d16$} & Reported \citep{tian2024visual}  & 3.30 & 274.4 & 10 \\
 & Our run  & 3.50 & 276.0 & 10 \\
\midrule
\multirow{4}{*}{HMAR-$d16$} & Markov Assumption & 3.76 & 293.3 & 10  \\
 & Loss Weighting  & 3.42 & 307.9  & 10  \\
 & Masked Prediction & 3.01 & 288.6 & 14 \\
\midrule
HMAR-$d30$ & Scale-up & \textbf{1.95} & \textbf{334.5} & 14 \\
\bottomrule
\end{tabular}
\vspace{-2mm}
\caption{\textbf{Ablation study comparing successive HMAR enhancements compared to VAR.} We show that each of our proposed methods improves both the image generation quality and diversity metrics.
\vspace{-4mm}}
\label{table:ablation-methods}
\end{table}

\section{Discussion and Conclusion}\label{sec:discussion}
\textbf{Conclusion}.~This paper introduces \underline{H}ierarchical \underline{M}asked \underline{A}uto\underline{R}egressive Image Generation (HMAR), a new image generation algorithm that improves upon Visual Autoregressive Modeling (VAR) in quality, efficiency and flexibility. HMAR enhances the efficiency of next-scale prediction by conditioning only on the immediate past scale instead of all previous scales. This accelerates inference, reduces memory usage, and enables a sparser attention pattern. We develop sparse attention kernels to leverage the sparse attention pattern, enabling faster training compared to VAR.  HMAR then incorporates masked prediction within each scale, providing flexible sampling while enhancing image quality. HMAR demonstrates superior performance on ImageNet benchmarks at $256\times 256$ and $512\times 512$ resolutions, matching or exceeding the quality of VAR, AR, and diffusion models while providing substantial improvements in training speed, inference speed, and memory efficiency.
\newline
\textbf{Limitations and Future Work.} While this work focuses on class-conditional image generation, we believe HMAR's framework can be naturally extended to Text-to-Image synthesis, offering another promising direction for future investigation. 
In future work, we also plan to investigate further improvements to the overall pipeline, including improvements to the multi-scale VQ-VAE tokenizer (Appendix~\ref{sec:tokenizer-failure}). 

\section{Acknowledgements}
 We thank Sabri Eyuboglu, Chris Fifty, Dan Biderman, Kelly Buchanan, Jerry Liu, Mayee Chen, Michael Zhang, Benjamin Spector, Simran Arora, Alycia Unell, Ben Viggiano, Zekun Hao, Siddharth Kumar, J.P. Lewis and Prithvijit Chattopadhyay for helpful feedback and discussions during this work. This work was supported by NIH under No. U54EB020405 (Mobilize), NSF under Nos. CCF2247015 (Hardware-Aware), CCF1763315 (Beyond Sparsity), CCF1563078 (Volume to Velocity), and 1937301 (RTML); US DEVCOM ARL under Nos. W911NF-23-2-0184 (Long-context) and W911NF-21-2-0251 (Interactive Human-AI Teaming); ONR under Nos. N000142312633 (Deep Signal Processing); Stanford HAI under No. 247183; NXP, Xilinx, LETI-CEA, Intel, IBM, Microsoft, NEC, Toshiba, TSMC, ARM, Hitachi, BASF, Accenture, Ericsson, Qualcomm, Analog Devices, Google Cloud, Salesforce, Total, the HAI-GCP Cloud Credits for Research program, the Stanford Data Science Initiative (SDSI), and members of the Stanford DAWN project: Meta, Google, and VMWare. The U.S. Government is authorized to reproduce and distribute reprints for Governmental purposes notwithstanding any copyright notation thereon. Any opinions, findings, and conclusions or recommendations expressed in this material are those of the authors and do not necessarily reflect the views, policies, or endorsements, either expressed or implied, of NIH, ONR, or the U.S. Government.

{
    \small
    \bibliographystyle{ieeenat_fullname}
    \bibliography{main}
} 
\clearpage
 
\twocolumn[{%
\noindent\rule{17.5cm}{0.8pt}
{\center
\section*{\huge HMAR \\ \Large Supplementary Material}
}
\appendix
\noindent\rule{17.5cm}{1.2pt}

\vspace{0.35cm}
{\Large \textbf{Contents}}

\addcontentsline{toc}{section}{Appendices}
\hypersetup{linkcolor=cvprblue}
\setcounter{section}{0}
\renewcommand\twocolumn[1][]{#1}%
\startcontents[appendix]
\printcontents[appendix]{1}{1}{}
}]
\clearpage
\setcounter{section}{0} 
\renewcommand\thesection{\Alph{section}}
\section{Extended Related Work}
\label{sec:related_work}
We highlight the key trade-offs between diffusion models, autoregressive image generation, masked generative models, and Visual Auto-Regressive Modeling (VAR), which represents the latest evolution in efficient autoregressive generation. We conclude by discussing efficient attention implementations.

\textbf{Diffusion Models} are the dominant class of generative models for image synthesis. Introduced by \citep{sohl2015deep} and further developed by \citep{ho2020denoising}, these models learn to reverse a gradual noising process, enabling high-quality image generation. Subsequent works have improved their efficiency \citep{dhariwal2021diffusion, song2020denoising,liu2023hyperhuman}, extended them to conditional generation \citep{dhariwal2021diffusion}, and applied them to various domains including text-to-image synthesis \citep{ramesh2022hierarchical, saharia2022photorealistic}. Diffusion models are preferred over previous image generation methods \citep{kingma2013auto, razavi2019generating, goodfellow2014generative} due to their superior image quality, diversity, and training stability, despite higher computational costs.

\textbf{Autoregressive Image Generation} models offer an alternative approach to image synthesis, drawing inspiration from the remarkable success of next-token prediction in natural language processing \citep{brown2020languagemodelsfewshotlearners, radford2019language,  touvron2023llama}. These models generate images sequentially, predicting each new token based on all previous ones, typically following a raster scan pattern. Early works \citep{pixelcnn, pixelcnn++, pixelrnn} operated directly in pixel space but faced significant computational challenges. More recent approaches have improved efficiency by using Vector Quantized VAEs \citep{vq-gan, oord2018vqvae, lee2022autoregressive} to compress images into discrete tokens for autoregressive generation, inspiring works like Parti \citep{parti} and LlamaGen \citep{llamagen}. While these models benefit from conceptual simplicity and potential transfer learning from language models, they still face challenges in generation speed and quality compared to diffusion models.

\textbf{Masked Generative Models} provide an alternative approach to improve the sampling speed of autoregressive models. These models generate images using a masked prediction objective similar to BERT \citep{devlin2019bertpretrainingdeepbidirectional, he2021masked, bao2022beitbertpretrainingimage}. By predicting multiple masked tokens in parallel, these models achieve faster generation speeds compared to next-token autoregressive image models.  The inherent independence assumption between masked tokens during parallel prediction can lead to inconsistencies or artifacts in the generated images. This approach has been explored in various recent works \citep{chang2022maskgit, chang2023muse, li2023mage, weber2024maskbitembeddingfreeimagegeneration}, and a unifying framework for these models is presented in MAR \citep{mar}, categorizing them as Masked Autoregressive models.

\textbf{Visual Auto-Regressive Modeling (VAR)} \citep{tian2024visual} enhances the efficiency and quality of autoregressive image generation.  Its versatility is demonstrated by successful adaptations for various tasks, including text-to-image generation \citep{ma2024star, zhang2024var, voronov2024switti, han2024infinity}, controllable image generation \citep{li2024controlvar}, depth estimation \citep{gabdullin2024depthart}, and video generation \citep{jin2024pyramidal}.  Furthermore, VAR has been effectively integrated with other techniques, such as residual diffusion \citep{tang2024hart} for improved image quality, speculative decoding \citep{chen2024collaborative,teng2024accelerating}, foldable tokens \citep{li2024imagefolder} for enhanced efficiency, solidifying its position as a powerful backbone for autoregressive image generation. However, these models still suffer from quality, efficiency, and flexibility issues.

\textbf{Efficient Attention Implementations} like FlashAttention \cite{shah2024flashattention, dao2022flashattention, dao2023flashattention2} allow to compute self-attention efficiently on GPU but only support a limited number of attention patterns. Xformer's Memory Efficient Attention \citep{rabe2021self} supports a wider range of attention patterns but provides memory optimization with limited speedup. Recent work, FlexAttention \citep{flexattention2024} supports a wider range of attention patterns while providing speedup. However, FlexAttention currently only supports sequence lengths that are multiples of 128, is not optimized for H100 GPUs \citep{flexattention2024}, and finally, its flexibility comes at a $10\%$ to $20\%$ performance cost \citep{flexattention2024}.
\section{Efficient Attention Computation}
We demonstrate how next-scale prediction is adversely affected by longer sequences in comparison to AR models, and how increasing the number of sampling steps results in even longer sequences relative to HMAR. Furthermore, we analyze the attention patterns in VAR and HMAR, highlighting why HMAR performs effectively when conditioned solely on the previous scale. Finally, we present microbenchmarks to evaluate the performance of attention computation using our optimized kernels.
\label{sec:efficient-attention}
\subsection{Long Sequences in Next-Scale Prediction}
\begin{figure}[H]
    \centering
    \includegraphics[scale=0.39]{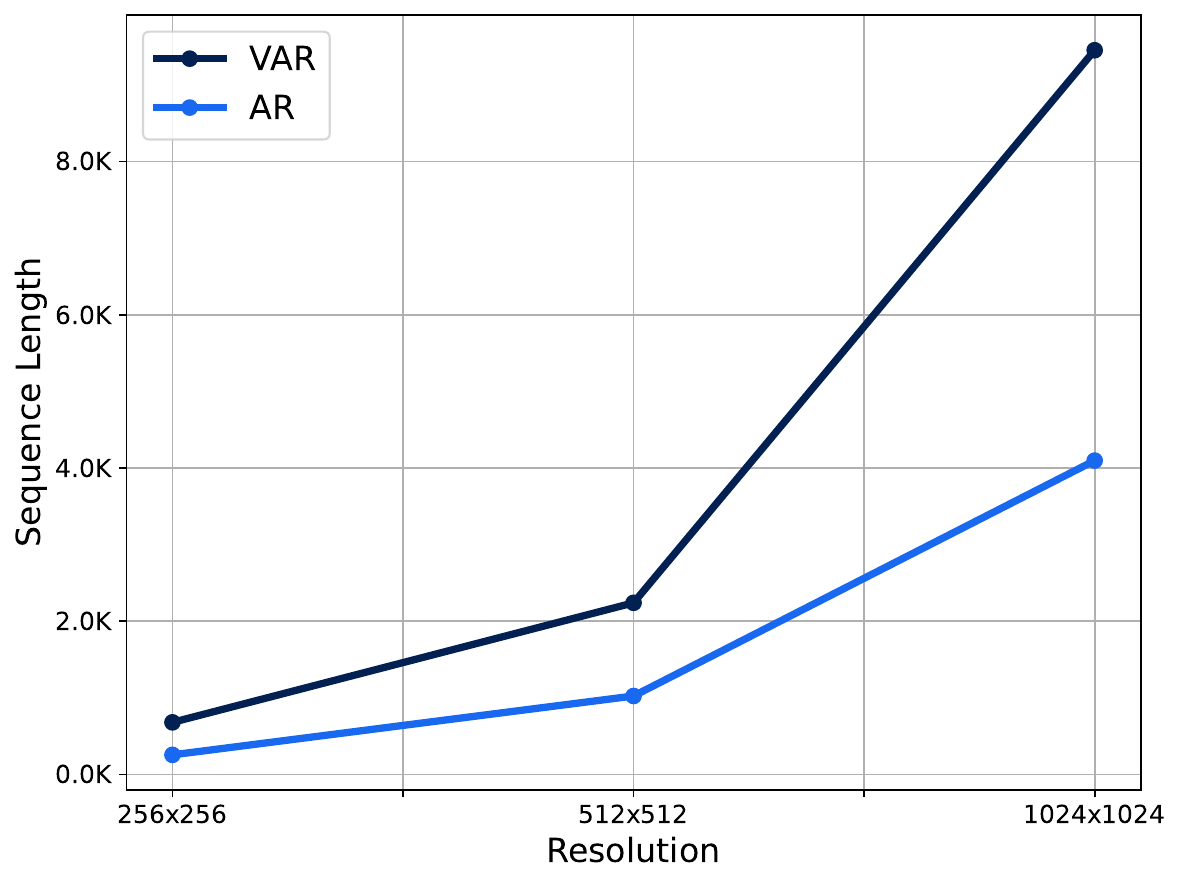}
  \caption{\textbf{Sequence Length vs Resolution for next-scale (VAR) and next-token (AR) prediction}. Next-scale prediction requires longer sequences compared to next-token prediction.}
    \label{fig:sequence-lengths-var-ar}
\end{figure}
\begin{figure}
    \centering
    \includegraphics[scale=0.375]{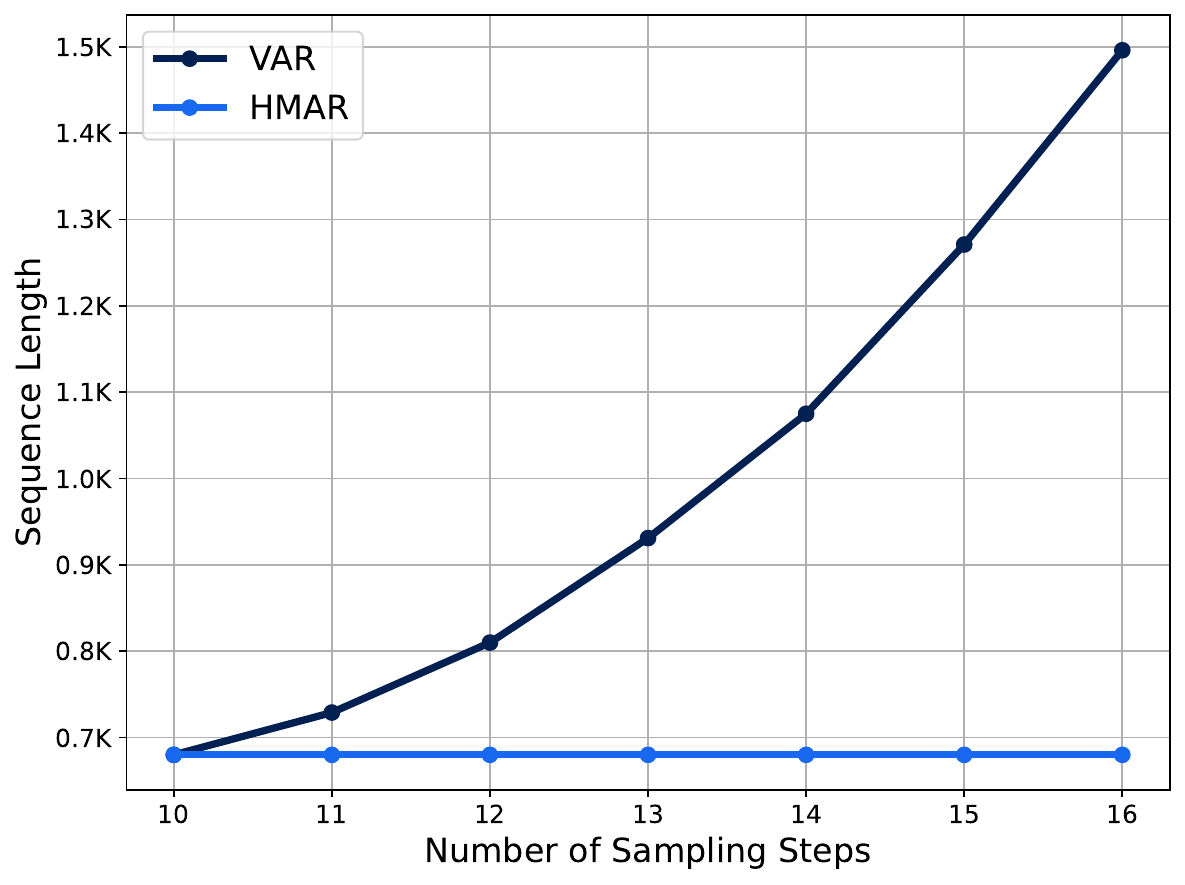}
    \caption{\textbf{Impact of Additional Sampling Steps on Sequence Length}: In VAR compared to HMAR, each additional sampling step leads to longer sequence lengths. We show comparisons for $256\times256$}
    \label{fig:sequence-lengths}
\end{figure}
Fig.~\ref{fig:sequence-lengths-var-ar} compares the sequence lengths in the next-scale prediction of VAR and next-token autoregressive image generation algorithms like VQ-GAN \citep{vq-gan} and LlamaGen\citep{llamagen}. As we grow to higher resolutions, the context length grows making it expensive to train VAR models compared to next-token prediction models. Fig.~\ref{fig:masking-qualitative} illustrates the positive impact of increased sampling steps through masking on generation quality.  While beneficial, achieving this with (VAR) presents several drawbacks.  Each additional sampling step requires a correspondingly longer sequence length, as shown in Fig.~\ref{fig:sequence-lengths}. This increased sequence length has consequences for VAR training, leading to higher computational costs, longer inference times, and greater memory requirements.  Our Hierarchical Masked Autoregressive (HMAR) formulation, in contrast, allows for a flexible increase in the number of sampling steps without necessitating any changes to the sequence length.  Furthermore, VAR models are inherently limited in their maximum number of sampling steps by the number of available scales. As a concrete example, a $16\times16$ latent space restricts VAR to a maximum of 16 sampling steps. HMAR overcomes this limitation, enabling up to 256 sampling steps without requiring the re-masking of previously unmasked tokens.  If re-masking is allowed, HMAR can theoretically accommodate an arbitrary number of sampling steps.

\subsection{Attention Pattern Analysis}
\begin{figure}
    \centering
    \begin{subfigure}{0.49\textwidth}
        \centering
        \begin{tabular}[c]{@{}c@{}c@{}c@{}}
            \includegraphics[scale=0.115]{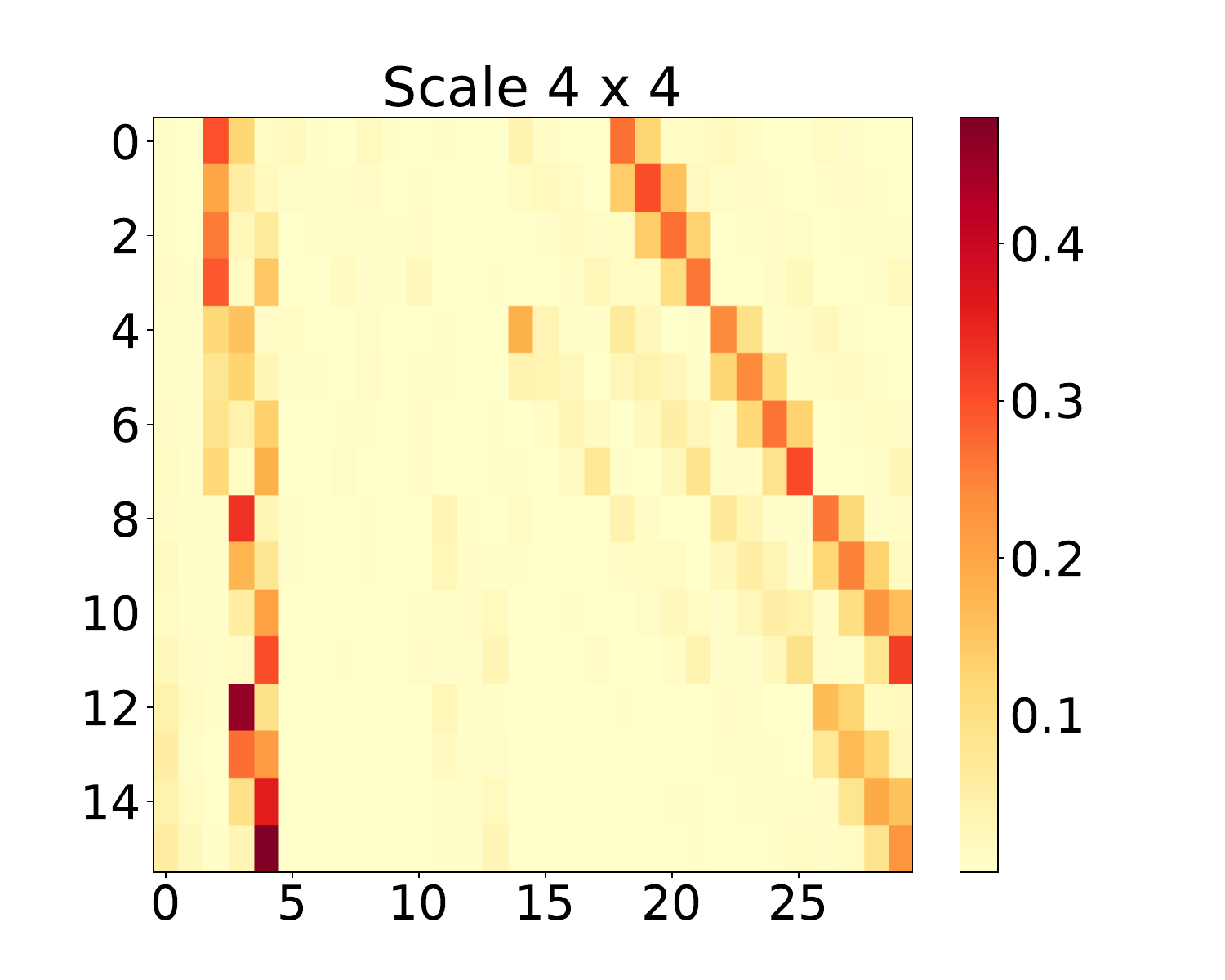} &
            \includegraphics[scale=0.115]{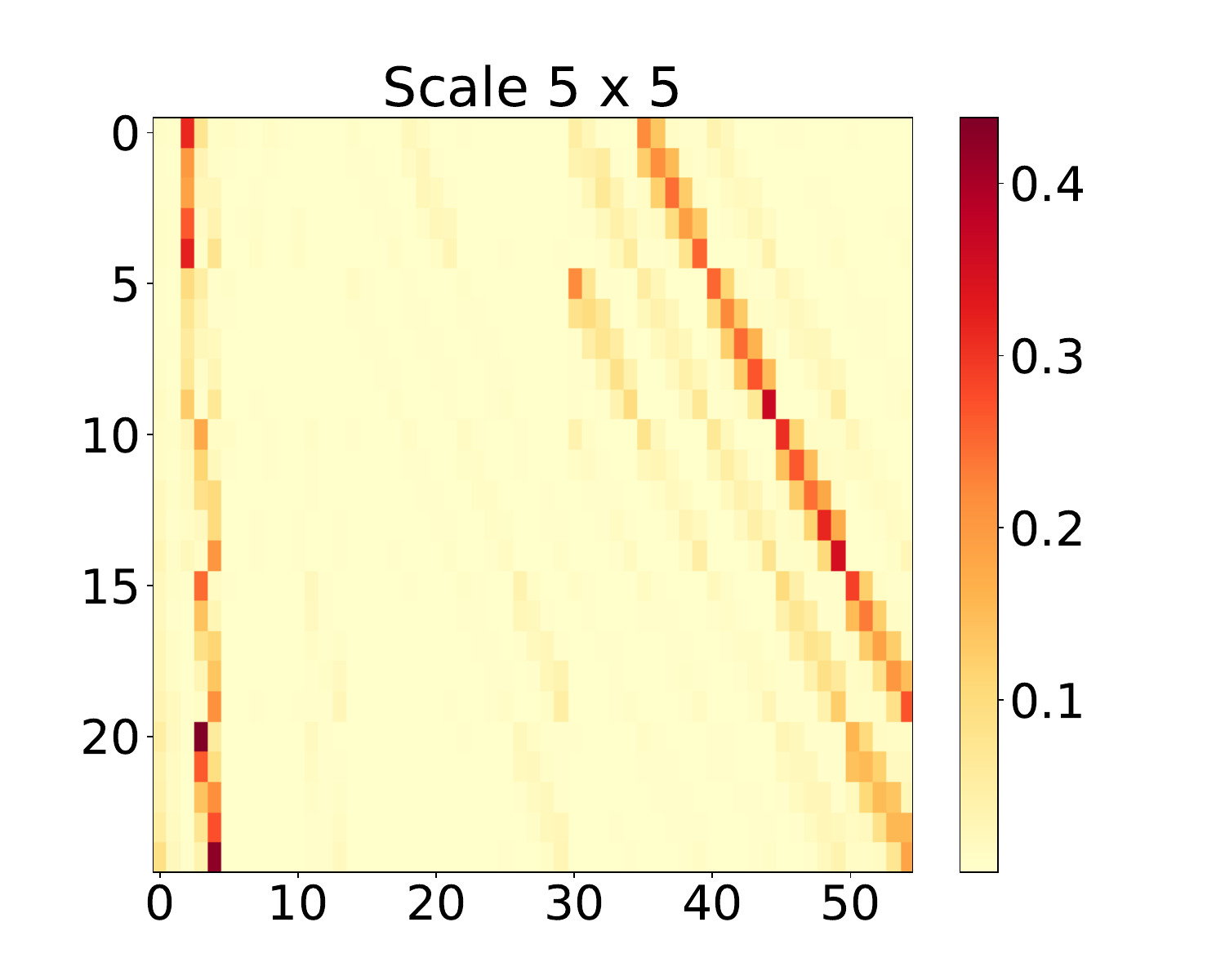} &
            \includegraphics[scale=0.115]{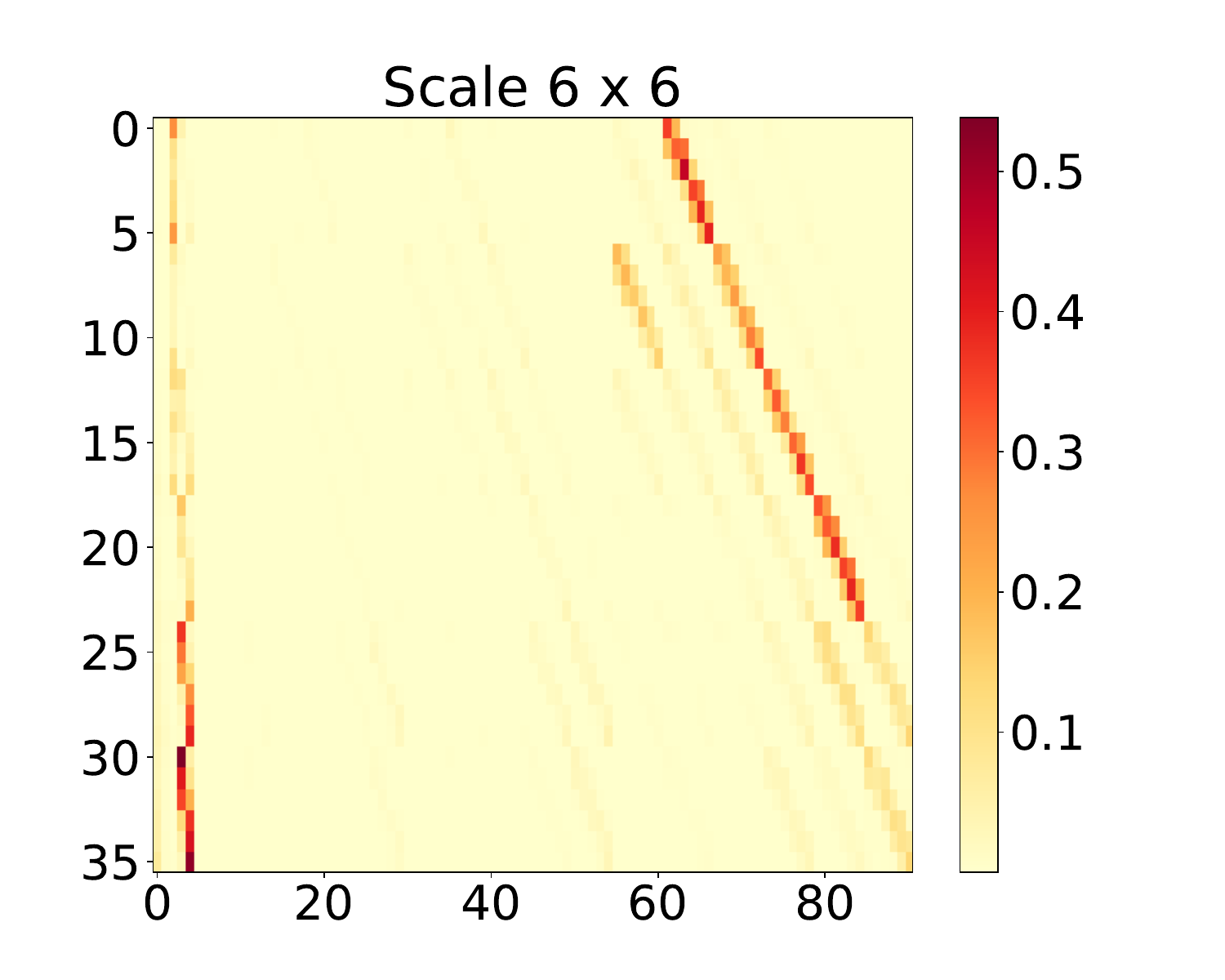} \\[-2pt]
            \includegraphics[scale=0.115]{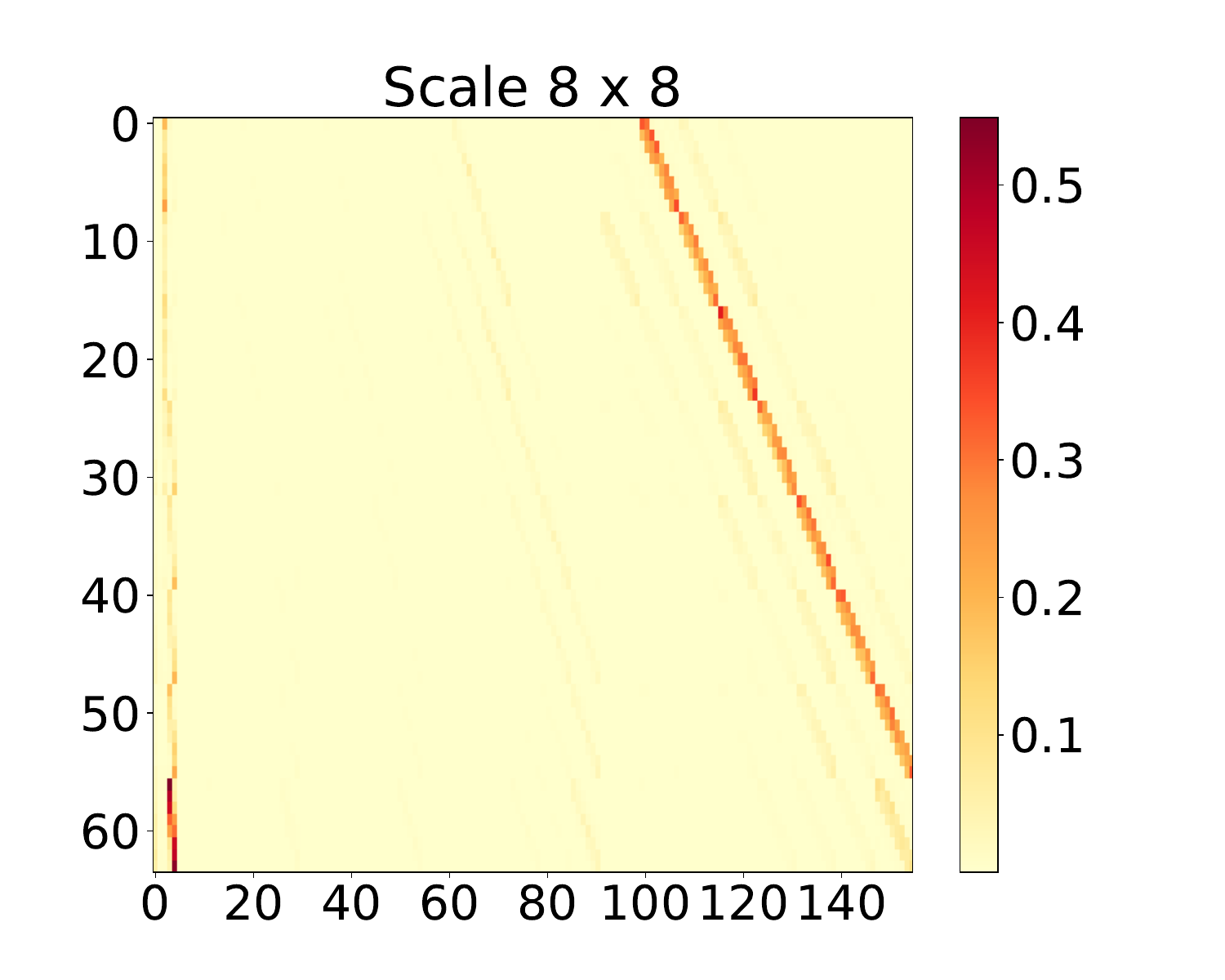} &
            \includegraphics[scale=0.115]{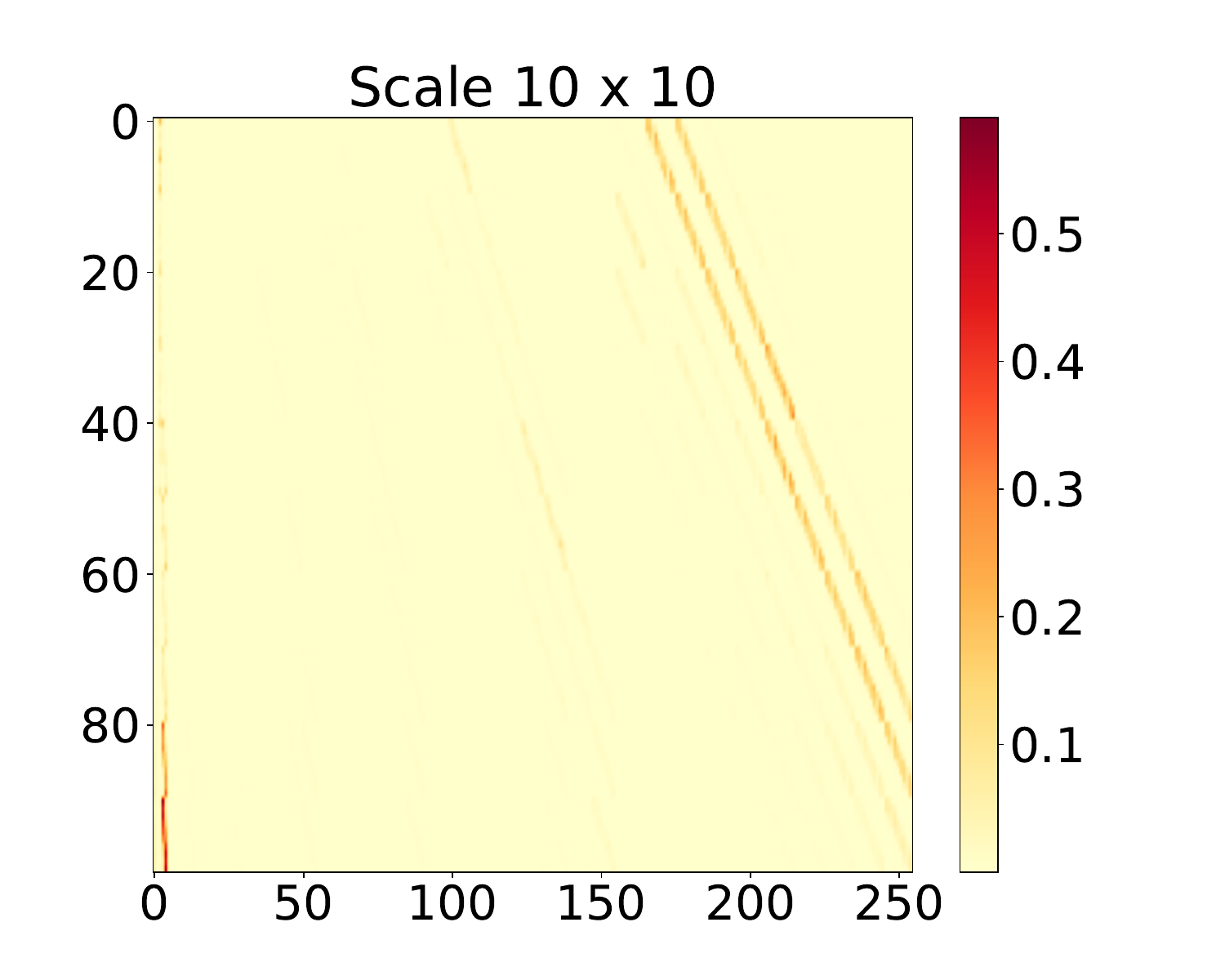} &
            \includegraphics[scale=0.115]{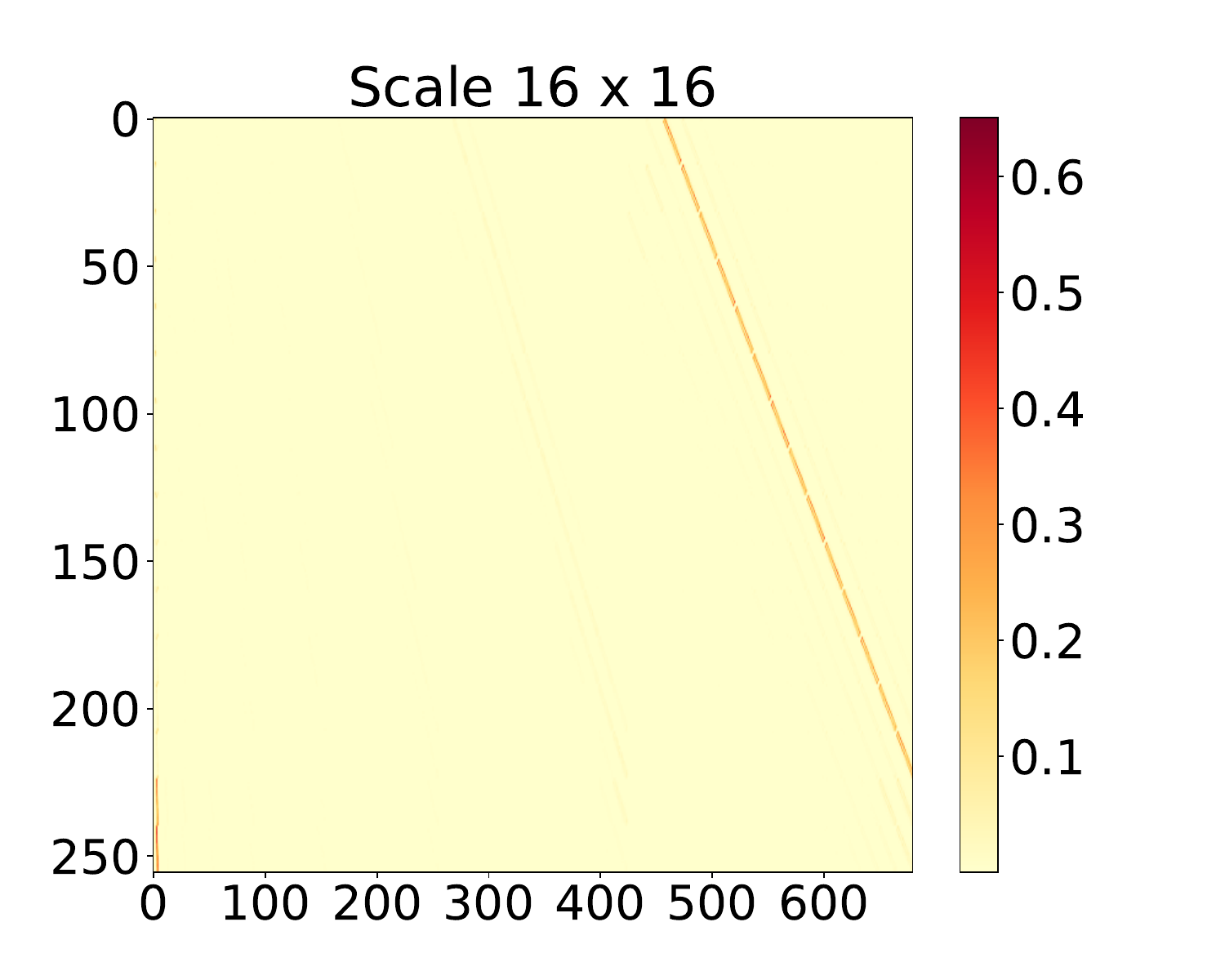}
        \end{tabular}
        \caption{Normalized Attention Scores VAR}
    \end{subfigure}%
    
\begin{subfigure}{0.49\textwidth}
    \centering
    \begin{tabular}[c]{@{}c@{}c@{}c@{}}
        \includegraphics[scale=0.115]{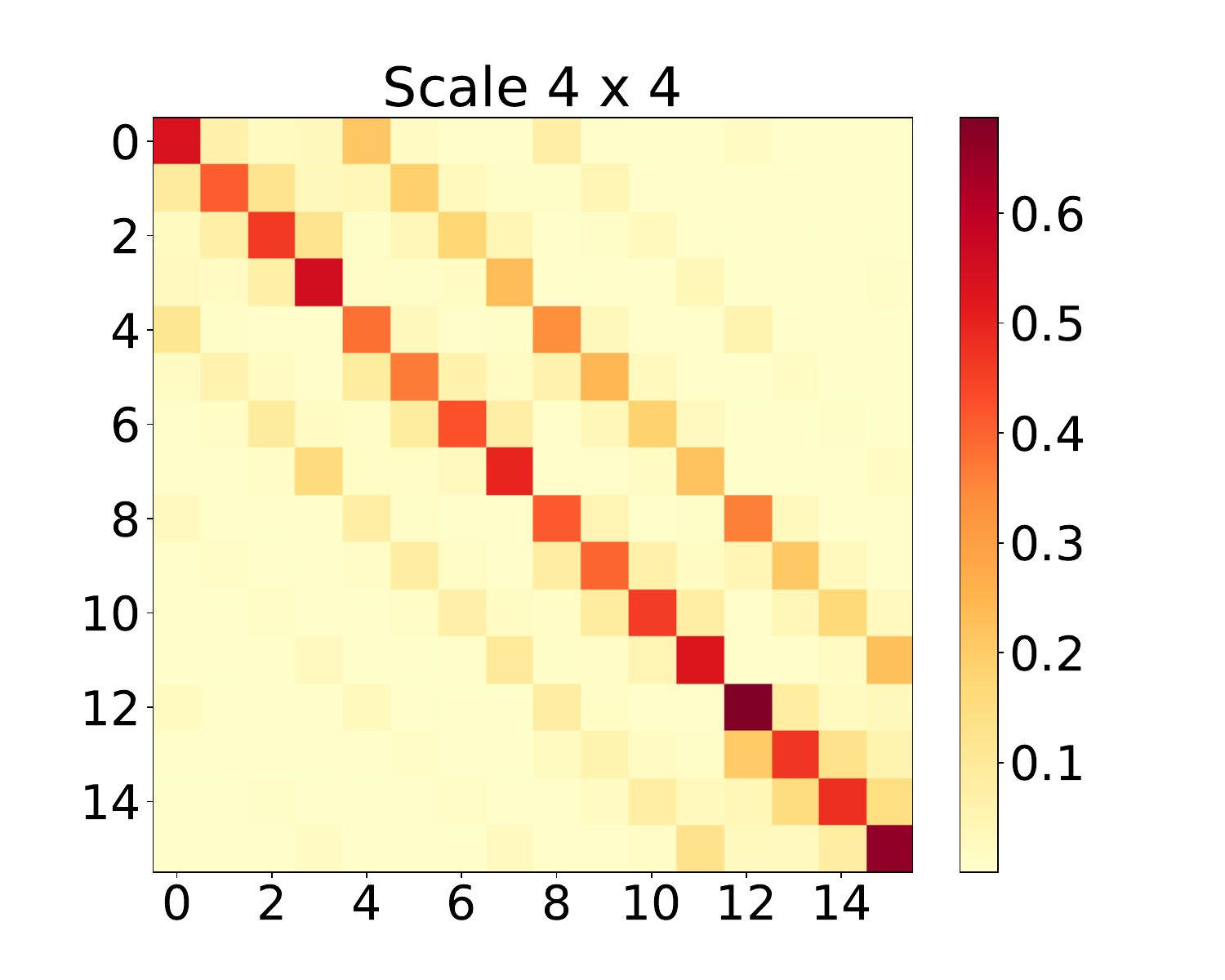} &
        \includegraphics[scale=0.115]{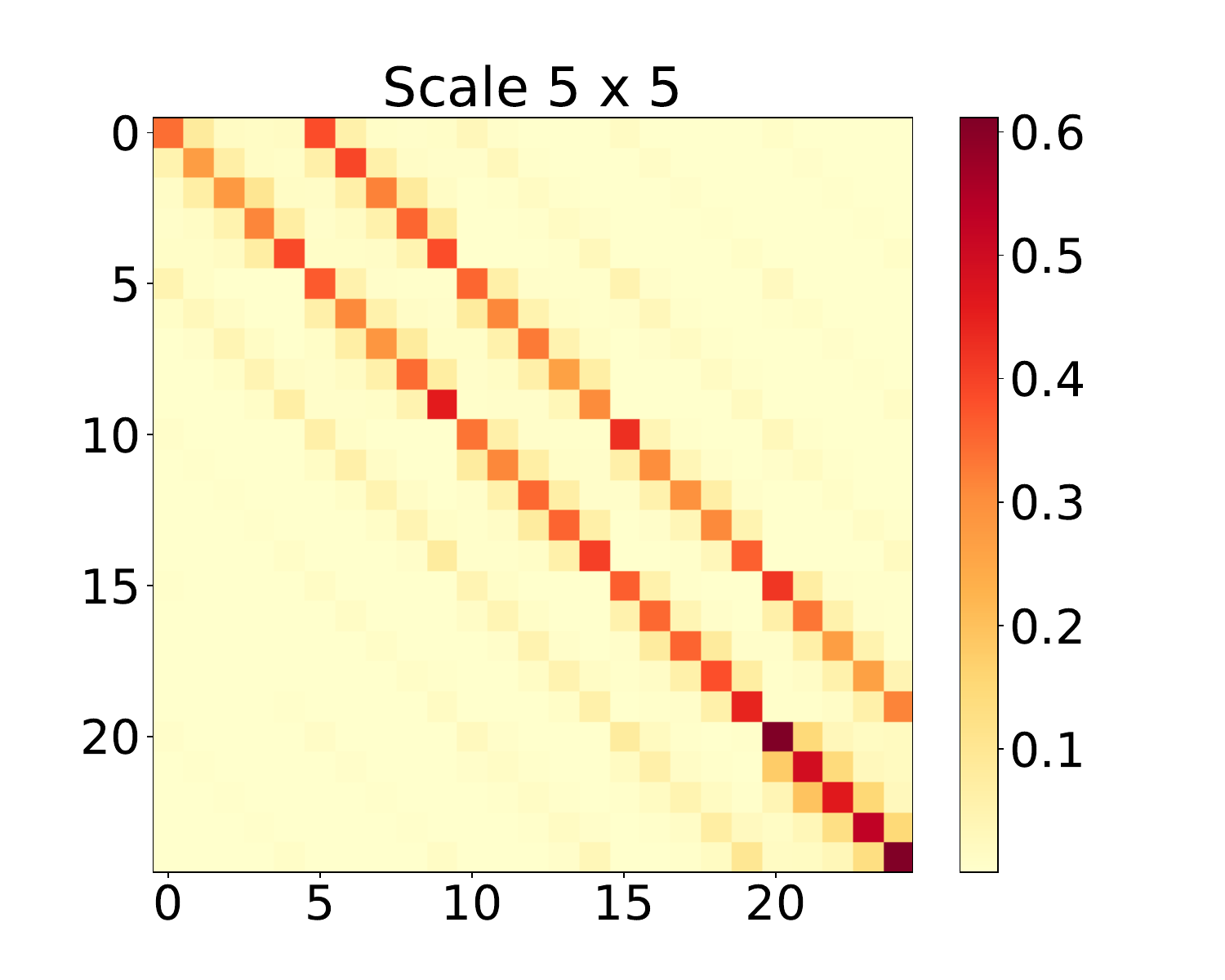} &
        \includegraphics[scale=0.115]{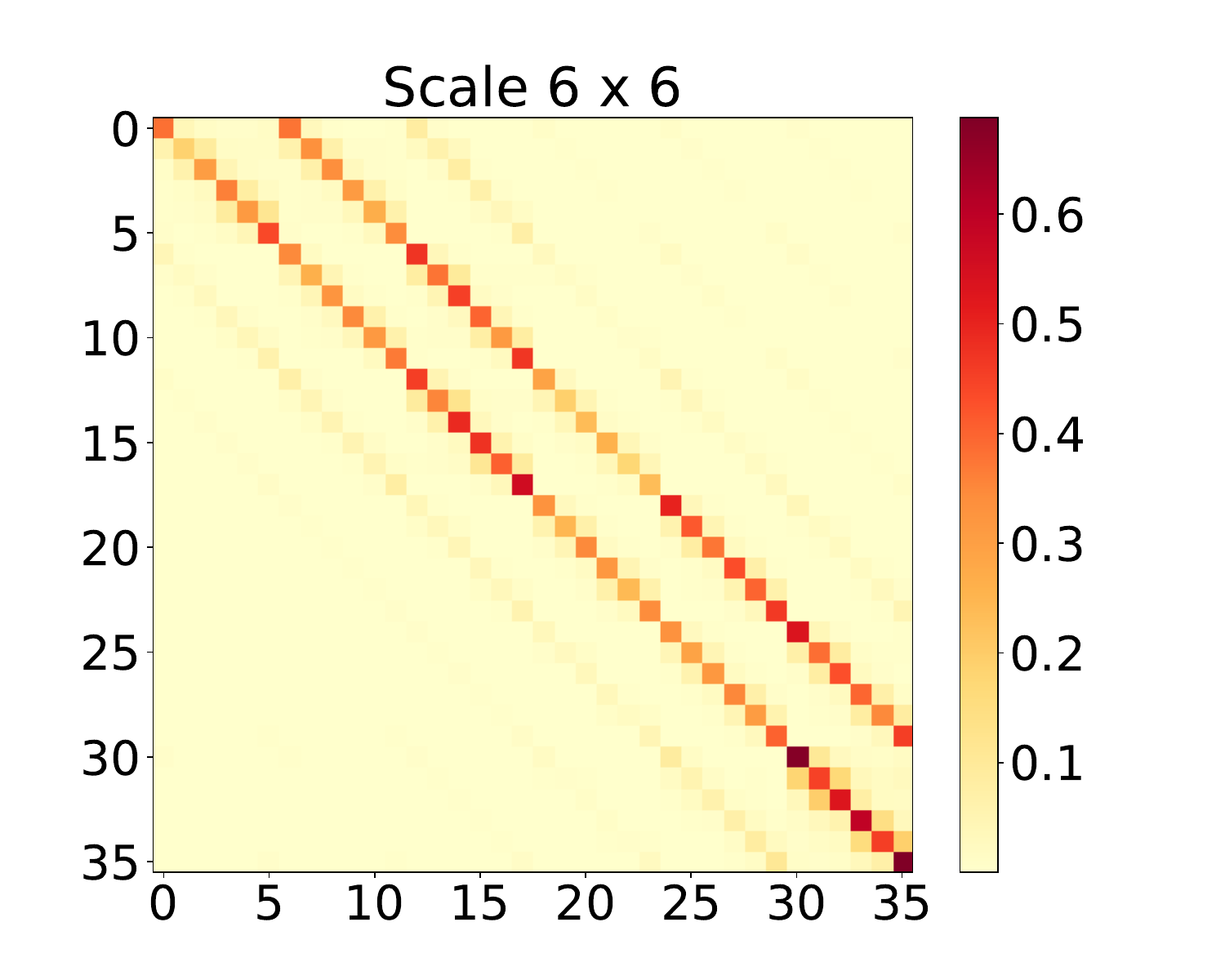} \\[-2pt]
        \includegraphics[scale=0.115]{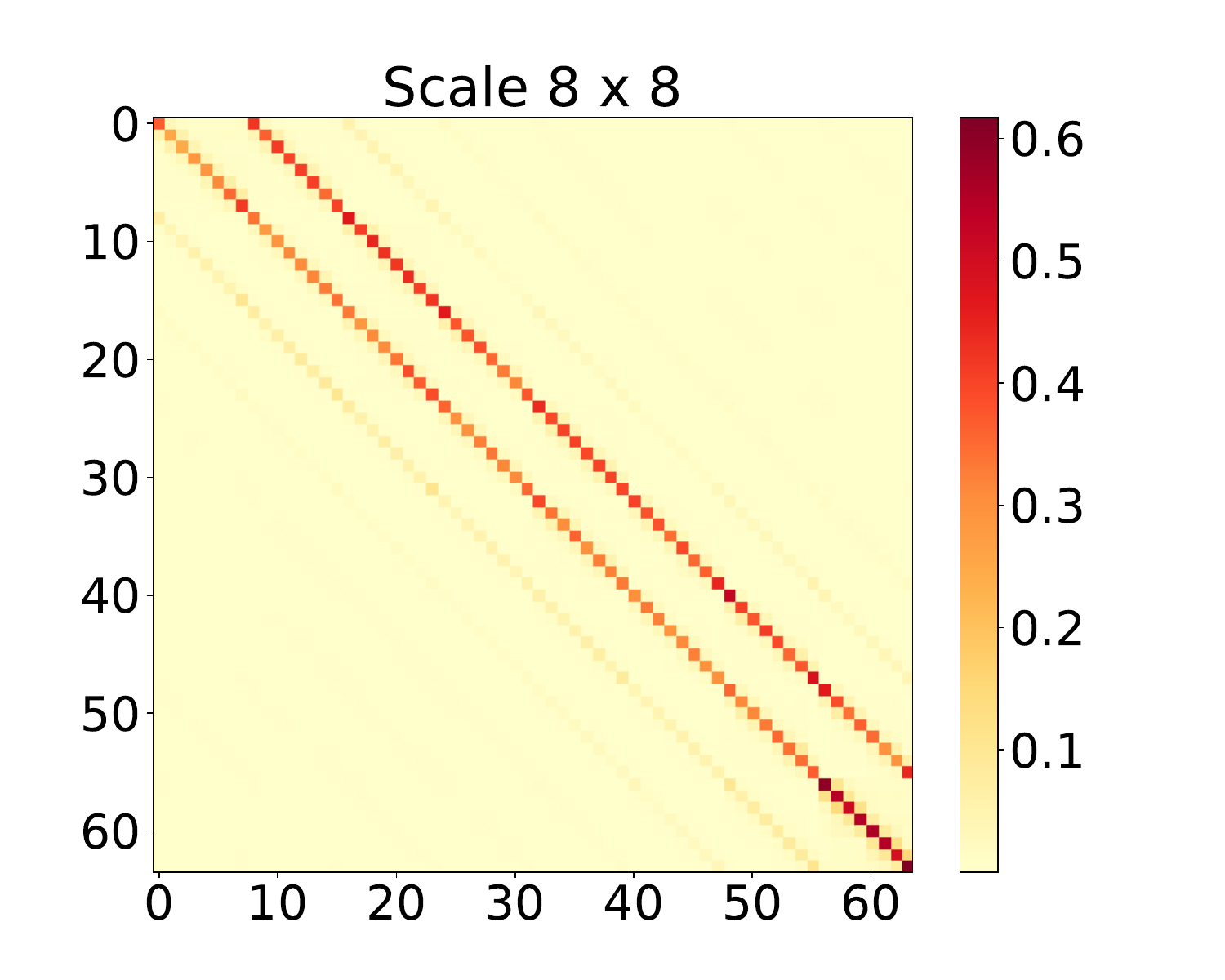} &
        \includegraphics[scale=0.115]{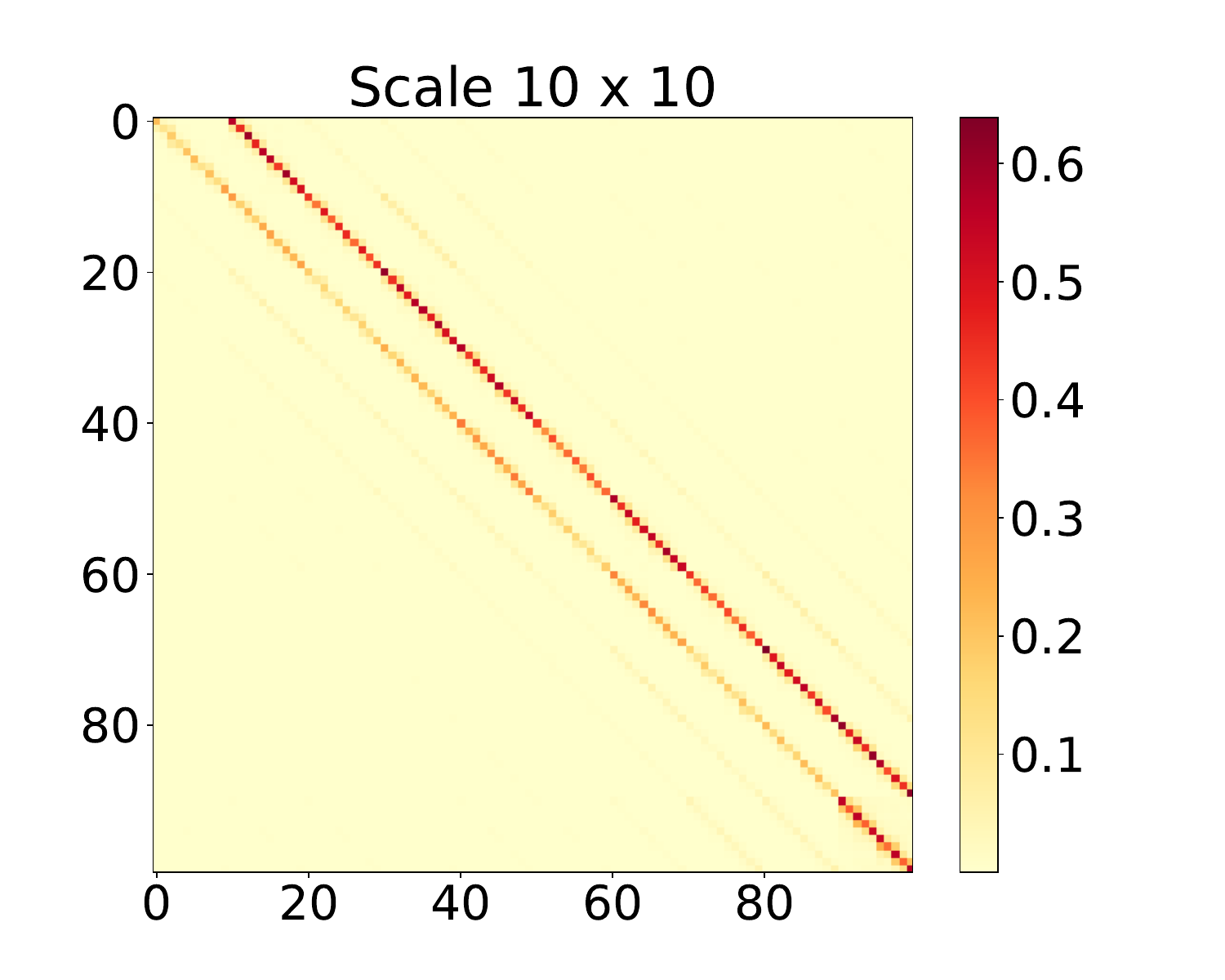} &
        \includegraphics[scale=0.115]{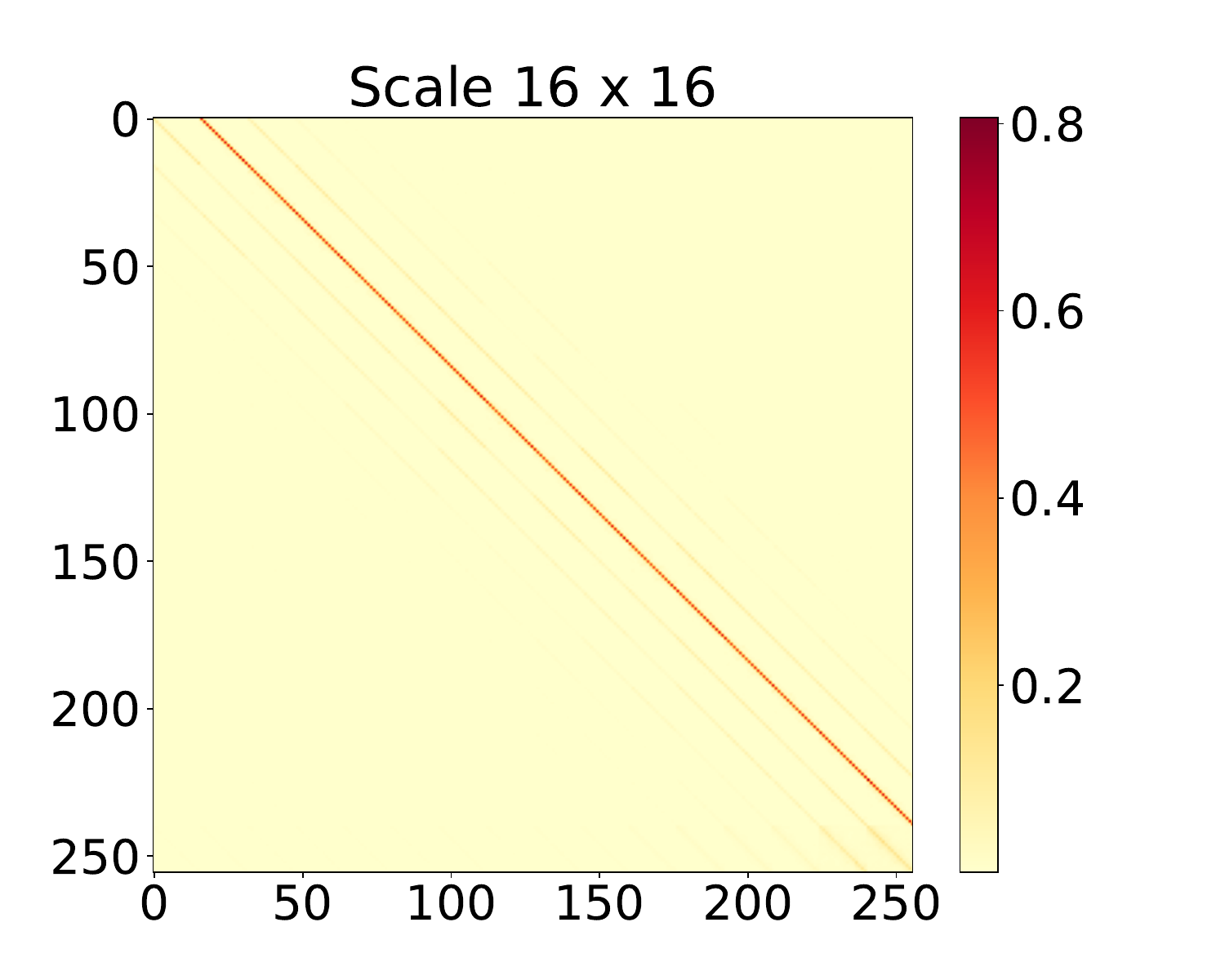}
    \end{tabular}
    \caption{Normalized Attention Scores in our HMAR}
\end{subfigure}%
    \caption{\textbf{Comparing Attention Patterns in VAR and  HMAR}}
    \label{fig:attn-map}
\end{figure}
The attention patterns visualized in Fig. \ref{fig:attn-map} reveal that tokens primarily attend to their local neighborhoods. This localized attention behavior provides strong empirical support for our hypothesis regarding next-scale prediction, demonstrating that the most relevant information for predicting the next scale is predominantly contained in the immediately preceding resolution level. This finding led us to streamline our model by replacing the full prefix conditioning mechanism with a simpler approach, where each scale only depends on its direct predecessor, notably preserving the model's predictive performance while achieving significant computational efficiency gains.
\subsection{Attention Patterns}
\begin{figure}
    \centering
    \includegraphics[scale=0.285]{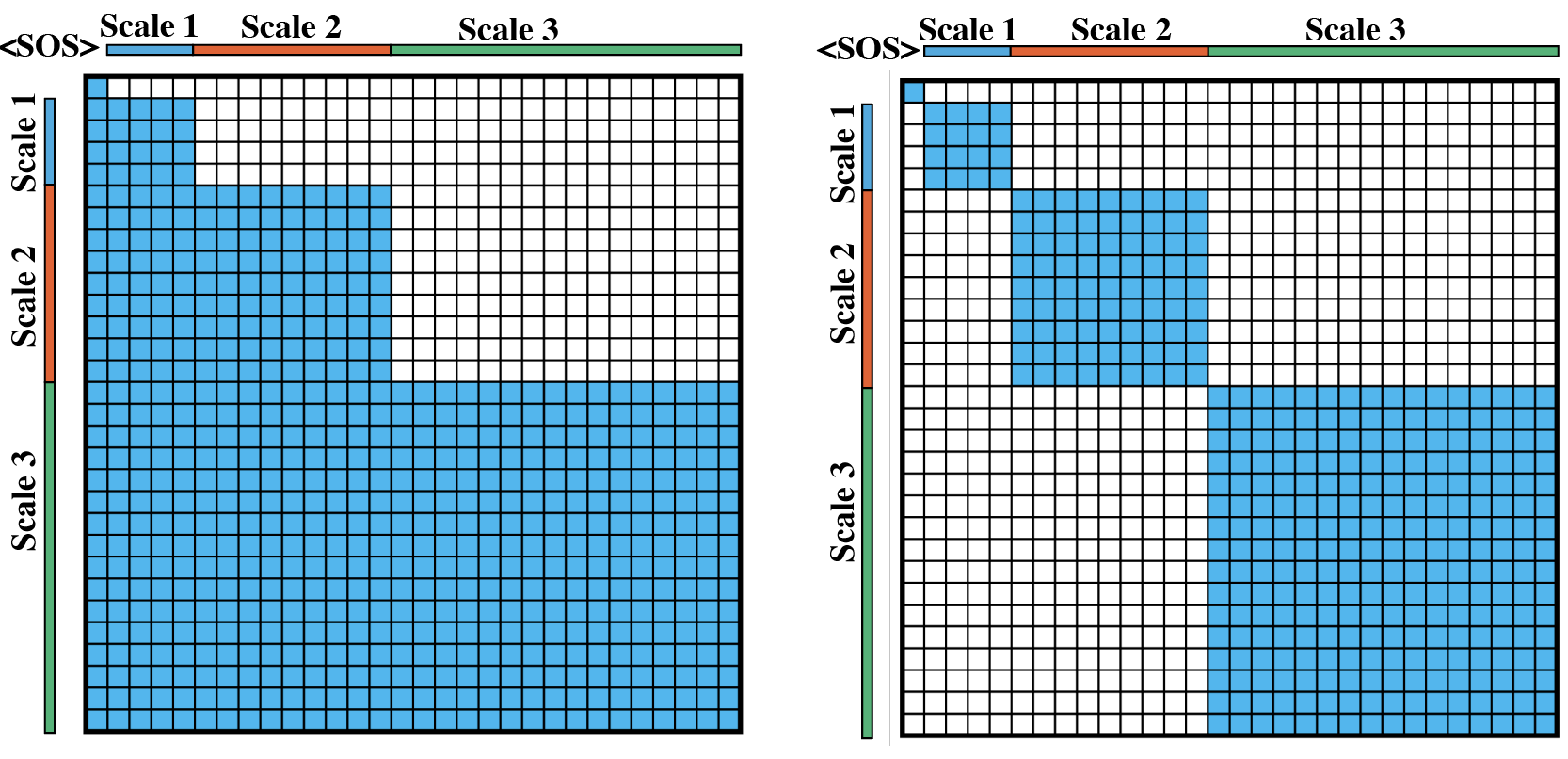}
    \caption{\textbf{Attention masks in VAR and HMAR}: Block-diagonal pattern in HMAR (right) enables more sparsity compared to the block-causal pattern in VAR(left).}
    \label{fig:attn-mask}
\end{figure}
We provide an illustration of the attention masks utilized in VAR and HMAR in Fig.~\ref{fig:attn-mask} to highlight their distinct mechanisms. VAR employs a block-causal attention mask, which allows each scale to attend not only to itself but also to all preceding scales. This design ensures a comprehensive flow of information across scales, facilitating a more global understanding of the data. In contrast, HMAR adopts a block-diagonal attention mask, where each scale is restricted to attending only to the immediately preceding scale in order to generate the next one. This results in a more localized and computationally efficient attention mechanism. As shown in Fig.~\ref{fig:attn-mask}, the block-diagonal mask is significantly sparser compared to the block-causal mask. This sparsity can be leveraged to achieve faster attention computation, particularly as the degree of sparsity increases with image resolution. Consequently, this approach becomes even more efficient for attention computation at higher resolutions.

\begin{figure*}
    \centering
    \begin{subfigure}{0.98\textwidth}
        \centering
        \begin{tabular}[c]{@{}c@{}c@{}}
           \includegraphics[scale=0.33]{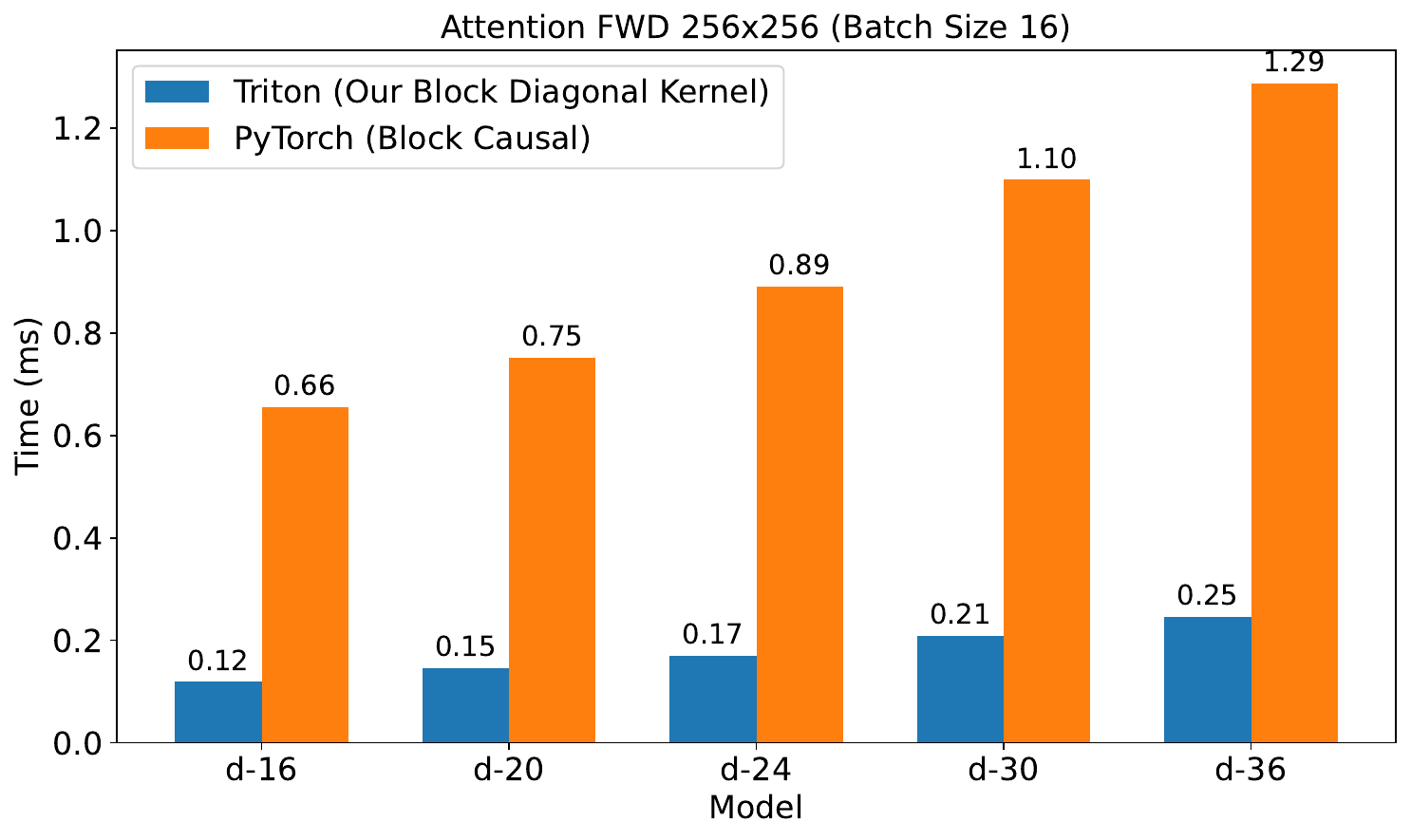}~~~~&~~~~
            \includegraphics[scale=0.33]{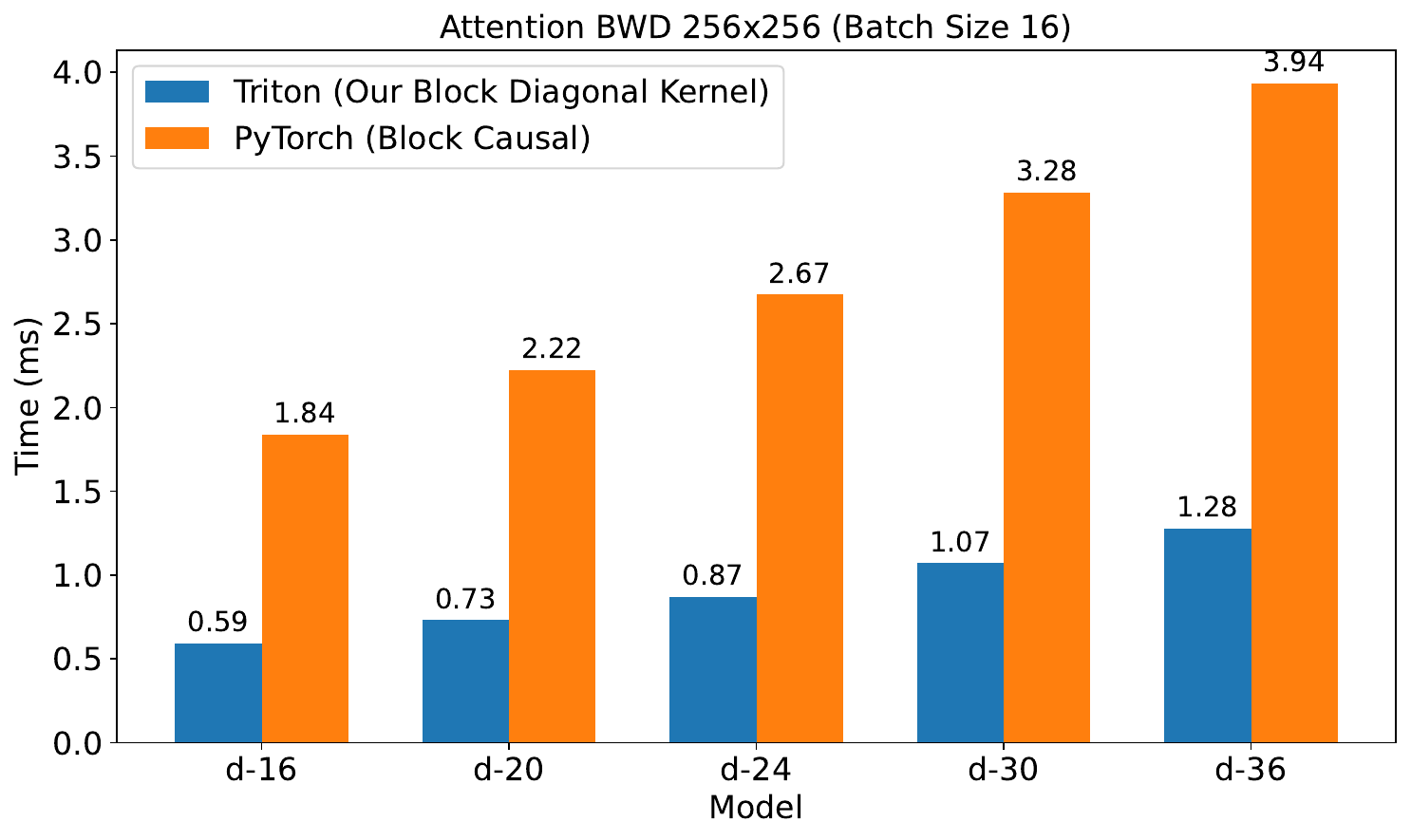} \\[-2pt]
            \includegraphics[scale=0.33]{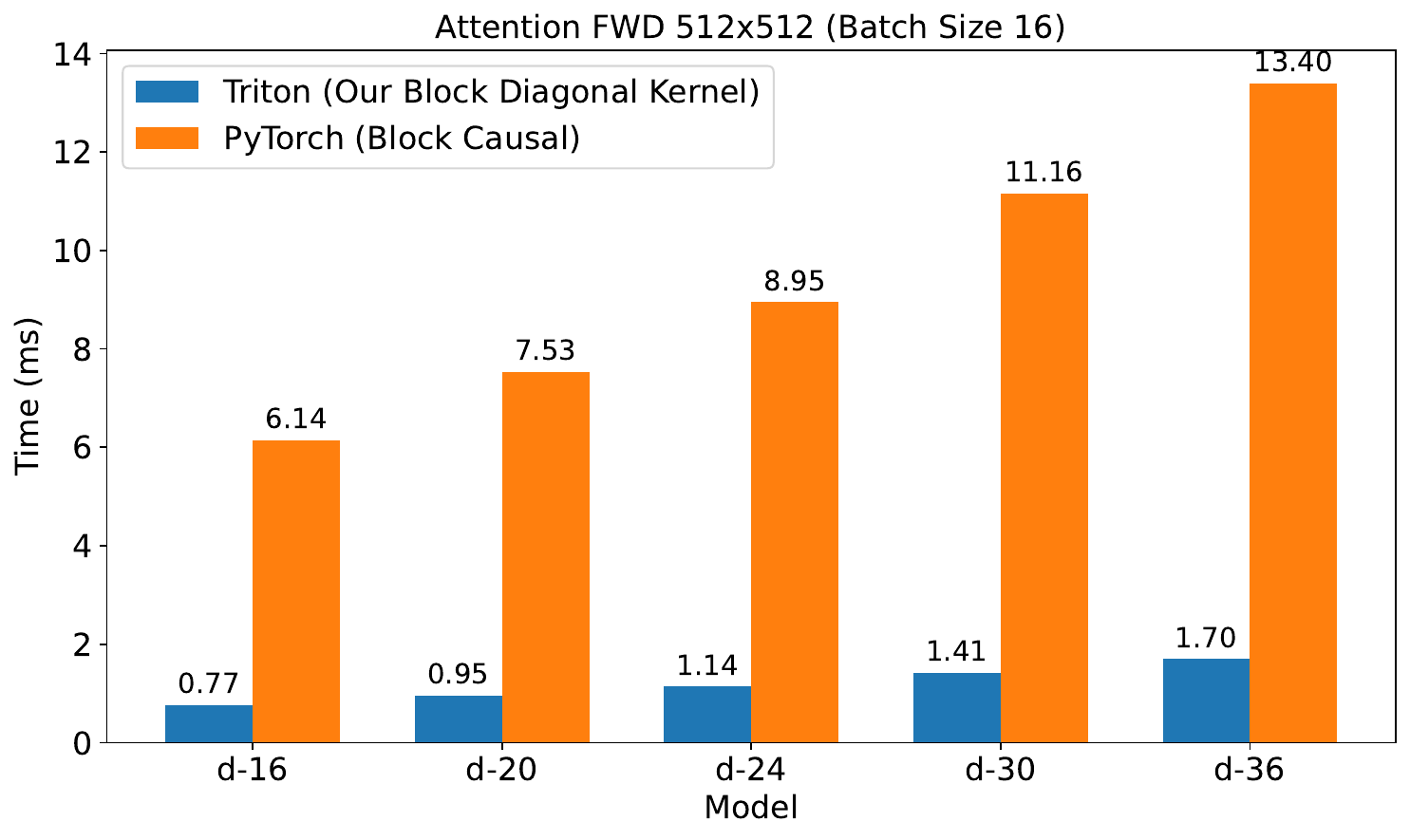} ~~~~&~~~~
            \includegraphics[scale=0.33] {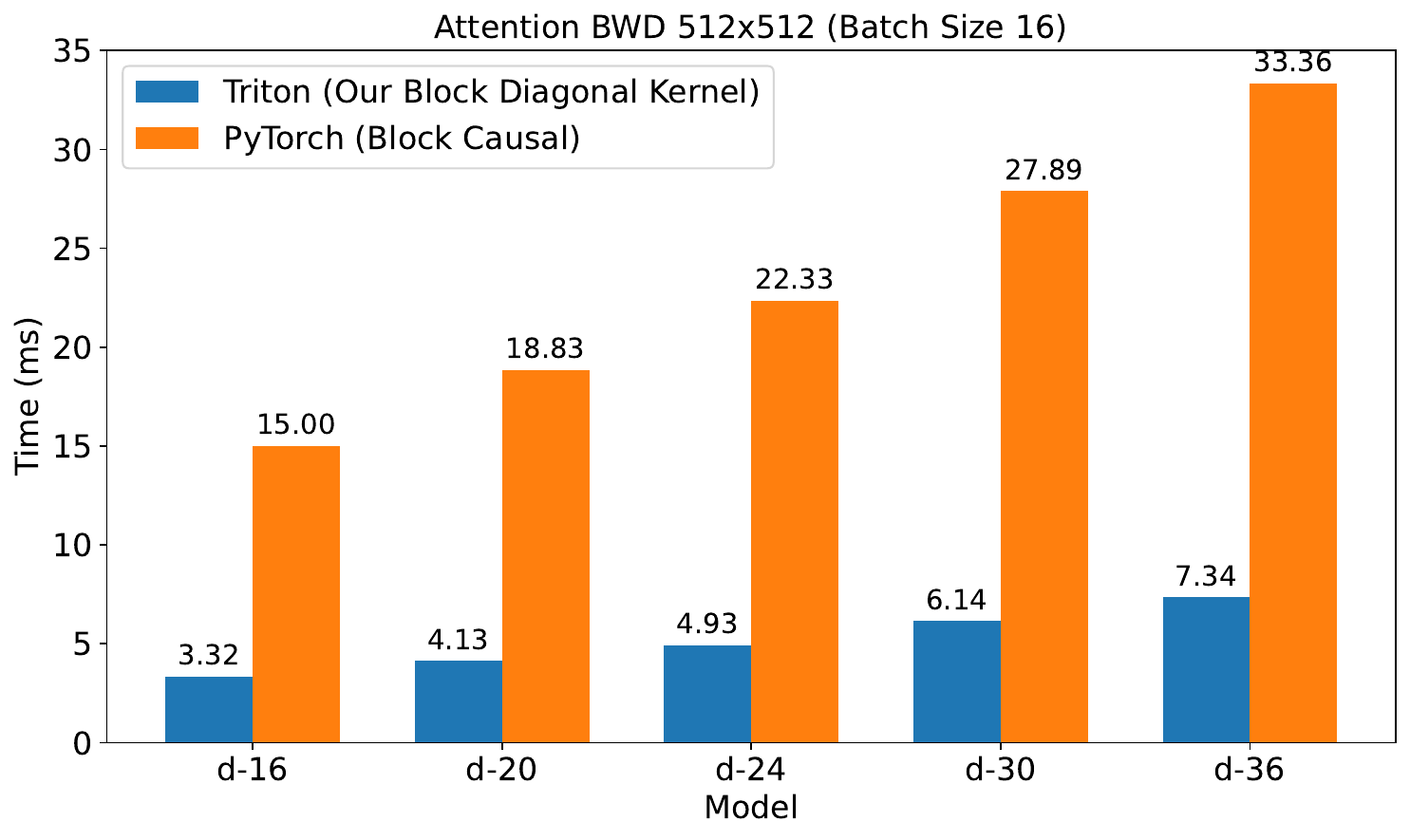} \\ 
            \includegraphics[scale=0.33]{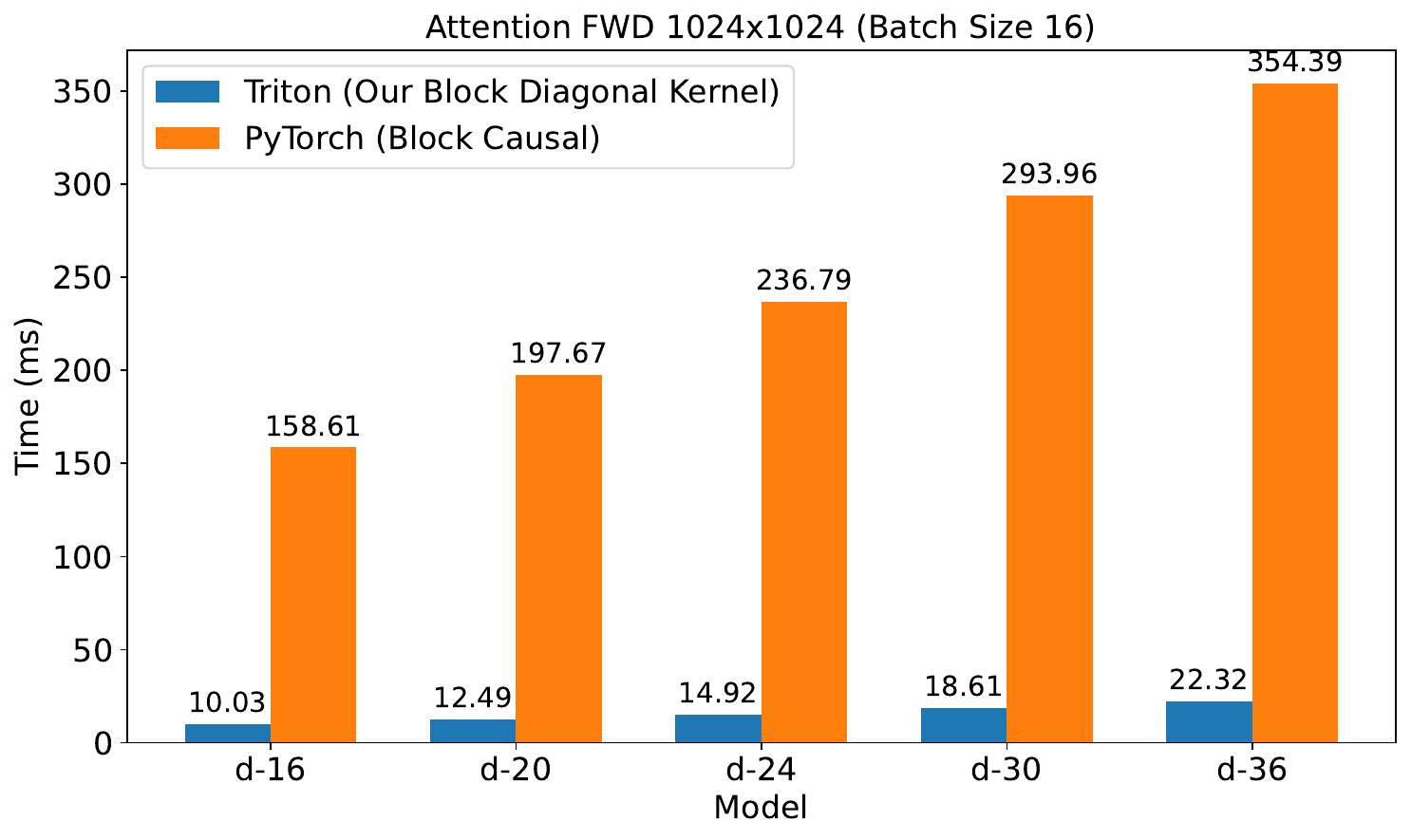} ~~~~&~~~~
           \includegraphics[scale=0.33]{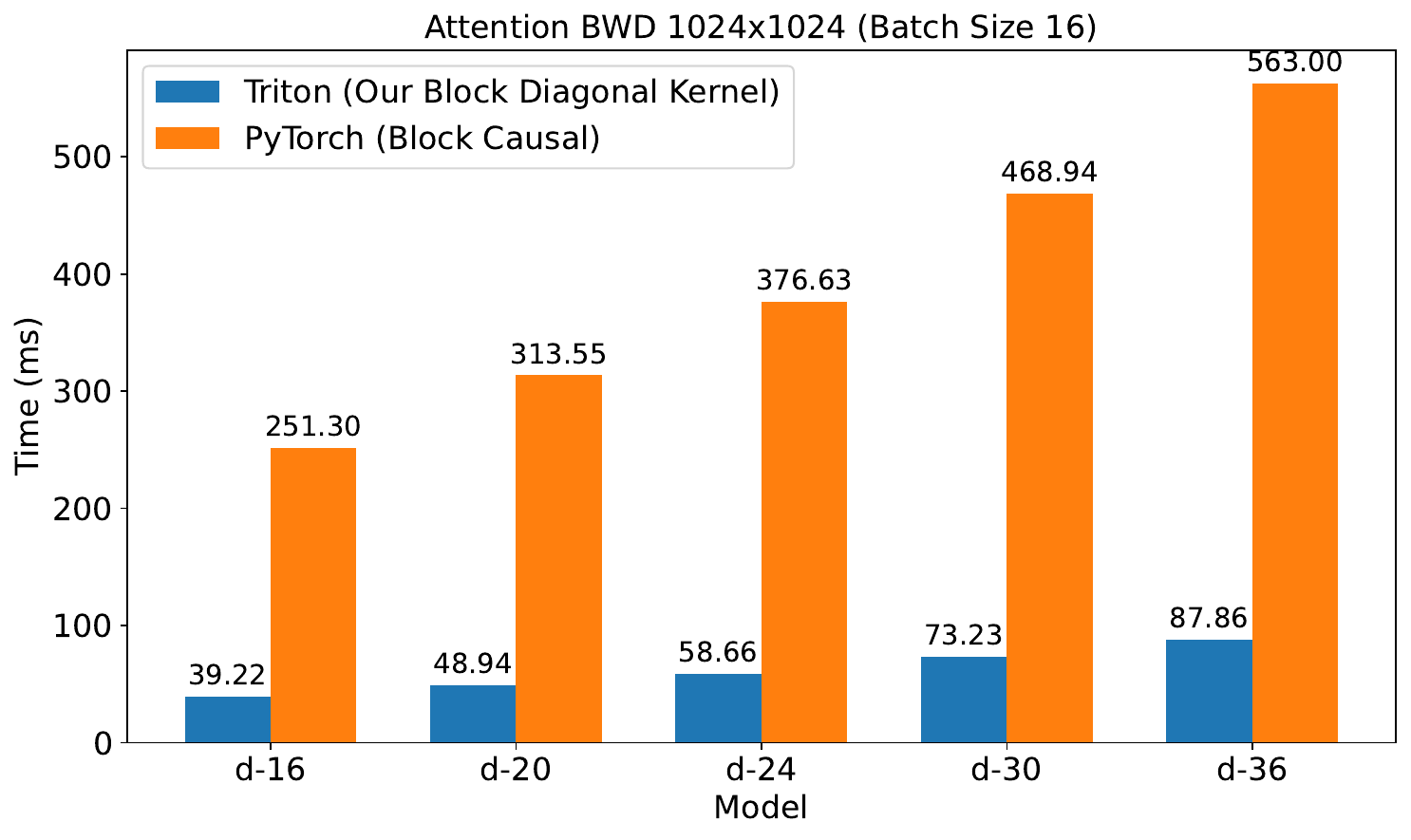}
        \end{tabular}
    \end{subfigure}%
    \caption{\textbf{Comparison of forward and backward pass speeds between our block sparse attention kernel in (HMAR) and the PyTorch block-causal attention in (VAR)} across different image resolutions (256×256, 512×512, and 1024×1024). Tests were performed on an A100 80GB GPU with batch size 16 and model dimension 64. Our implementation shows significant speedups, achieving up to 15.8× faster forward pass and 6.4× faster backward pass at 1024×1024 resolution.}
    \label{fig:efficient-attn}
\end{figure*}
\subsection{Efficient Attention Performance}
\label{sec:efficient-attention-performance}

We benchmark the performance of our block sparse attention kernel used in HMAR against the block-causal attention in VAR. For the block-causal attention, we compare against the memory-efficient attention implementation that supports different attention masks via \texttt{torch.sdpa}. The benchmarking is conducted across various image resolutions, utilizing a single A100 80GB GPU  and results are averaged over 25 repetitions. Results are shown in Fig.~\ref{fig:efficient-attn}. At each resolution, we evaluate both the forward and backward passes, ensuring a comprehensive analysis of performance. The benchmarks are conducted with a batch size ($bs$) of $16$ and a model dimension ($d$) of $64$. For each corresponding model, such as d-$24$ or d-$20$, the number of attention heads matches the model dimension—for instance, the d-$24$ model has $24$ attention heads.

On the forward pass, our efficient implementation is up to $5.2\times$ faster at $256\times256$, $7.9\times$ faster at $512\times512$, and $15.8\times$ faster at $1024\times1024$. On the backward pass, our efficient implementation is up to $3.1\times$ faster at $256\times256$,  $4.6\times$ faster at $512\times512$, and $6.4\times$ faster at $1024\times1024$, demonstrating significant performance improvements of our implementation across various resolutions. 
\clearpage
\section{Training Dynamics}
In this section, we delve into how each scale contributes to visual quality and how to focus the model's capacity during training on the most important scales that matter for visual quality. 
\subsection{Learning Difficulty Across Scales}
\label{sec:learning-difficulty}
\begin{figure}[h]
    \centering
    \includegraphics[width=7.35cm, height=4.5cm]{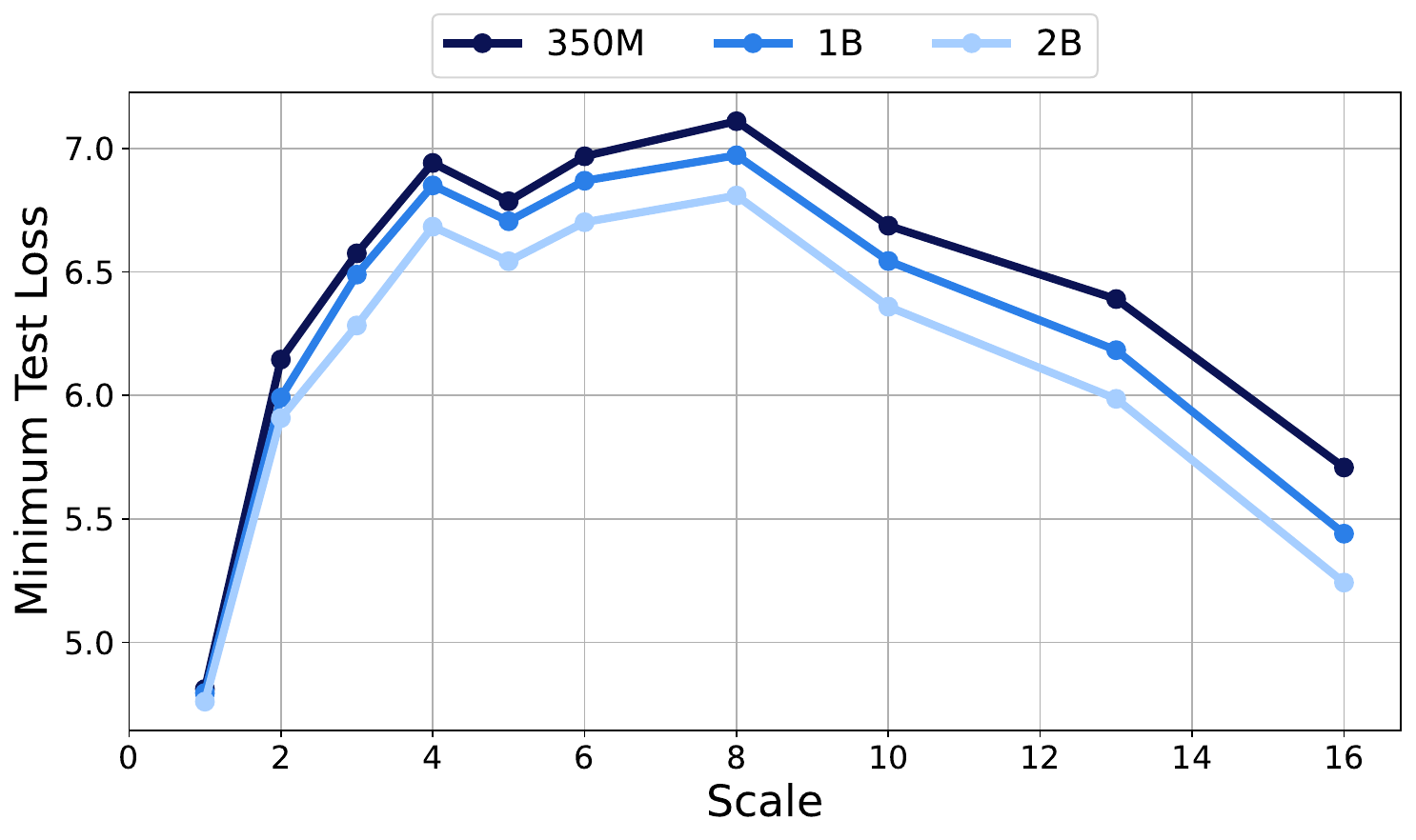}
    \caption{\textbf{Minimum Test Loss Across Scales}}
    \label{fig:loss-distribution}
\end{figure}
\noindent We use the minimum test loss at each scale as a proxy for learning difficulty. We find that this has an approximately log-normal pattern (Fig. \ref{fig:loss-distribution}),  suggesting that scales in the middle are more challenging to learn compared to those at the beginning and end.

\subsection{Loss Weighting Ablation}
\label{sec:loss-weighting-ablation}
\begin{figure}[H]
    \centering
    \includegraphics[width=7.35cm, height=4.8cm]{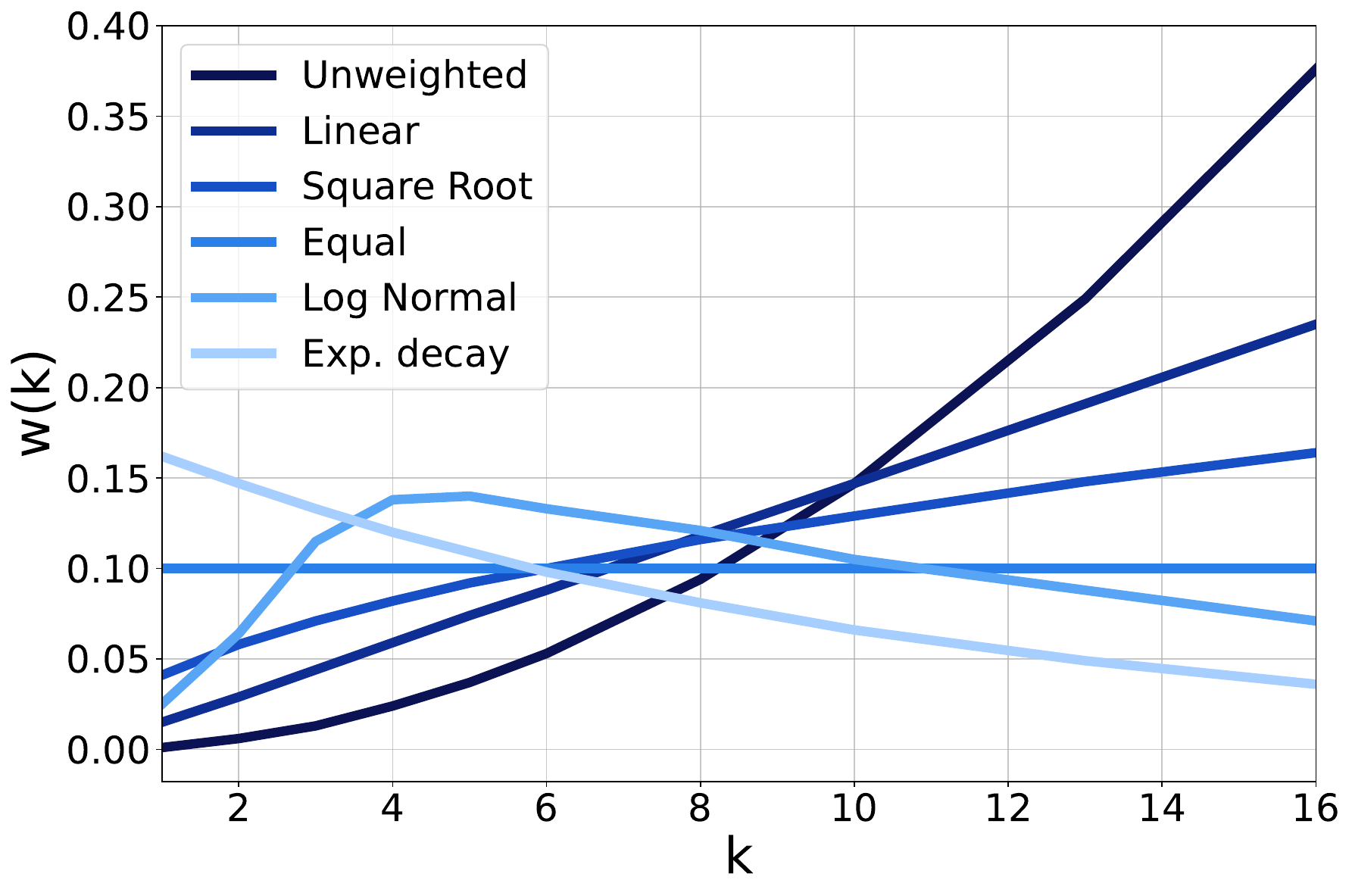}
    \caption{\textbf{Different Loss Weighting Functions}}
    \label{fig:loss-weighting}
\end{figure}
We ablate the impact of loss weighting on image quality. For each function  (Fig. \ref{fig:loss-weighting}), we train a $310$M parameter model for approximately $150$K steps and evaluate the FID, Inception Score, Precision, and Recall. Results in Table. \ref{table:loss-reweighting} demonstrate that loss weighting can significantly influence the quality, with the log-normal weighting yielding the best performance.

\begin{table}
\centering
\begin{tabular}{l|cc|cc}
\toprule
\multicolumn{5}{c}{\textbf{Loss Weighting Ablation on ImageNet-$256\times256$}}\\
\toprule
\textbf{Function} & \textbf{FID} $\downarrow$ & \textbf{IS} $\uparrow$ & \textbf{Prec} $\uparrow$ & \textbf{Rec} $\uparrow$\\
\midrule
Unweighted & 3.89 & 283.3 & 0.86 & 0.48 \\
\midrule
Equal & 3.64 & 296.5 & 0.85 & 0.50 \\
\midrule
Linear & 3.72 & 301.9 & \textbf{0.86} & 0.50 \\
\midrule
Sqrt. & 3.79 & 306.6 & 0.86 & 0.50 \\
\midrule
Exp. decay & 3.72 & 281.1 & 0.84 & 0.50 \\
\midrule
Log-normal & \textbf{3.59} & \textbf{307.4} & 0.85 & \textbf{0.50} \\
\bottomrule
\end{tabular}
\caption{\textbf{Loss Reweighting Ablation on ImageNet-$256\times256$}. (cfg=$1.5$, top-k=$900$, top-p=$0.96$) We show the impact of the choice of loss weighting on image quality. A log-normal weighting that mirrors the distribution of learning difficulty yields the best performance.}
\label{table:loss-reweighting}
\end{table}

\subsection{Loss Analysis}
Unlike autoregressive language models, where total loss correlates with downstream performance, we find this relationship doesn't hold in our setting. Due to the disproportionate number of tokens in later scales, the total loss is heavily influenced by performance on high-frequency details that are often imperceptible to human observers. As shown in Fig. \ref{fig:loss-acc-decomposition}, models with better performance in early and middle scales achieve superior FID and Inception Scores, despite potentially higher total losses. 
\begin{figure}[H]
    \centering
    \includegraphics[width=7.35cm, height=5.2cm]{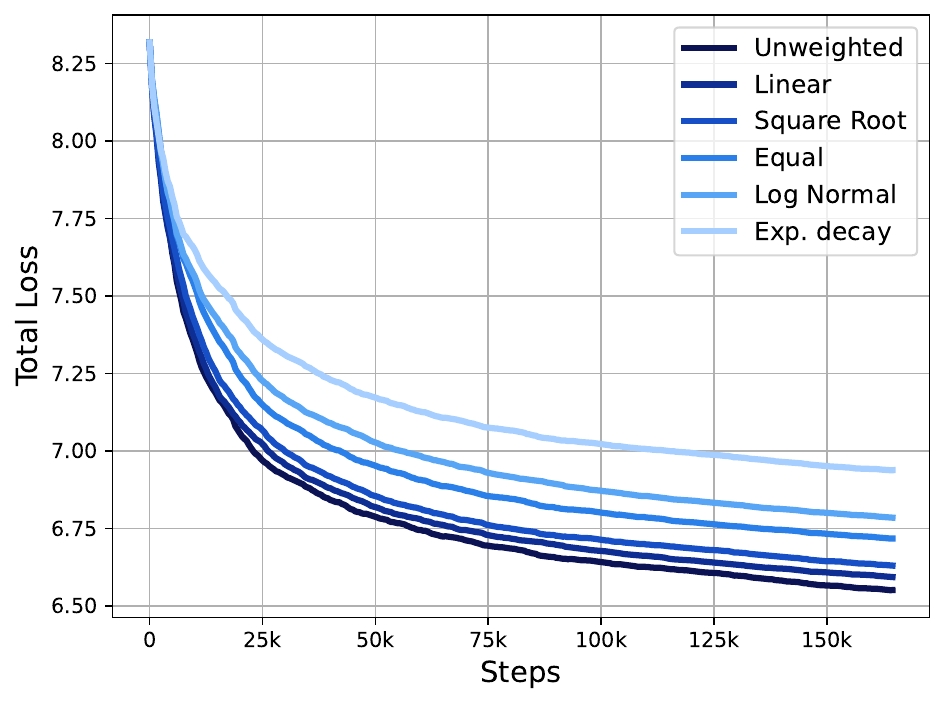}
    \caption{\textbf{Total Loss for Different Weighting Functions}}
    \label{fig:enter-label}
\end{figure}

\begin{figure}[H]
    \centering
    \includegraphics[width=7.35cm, height=5.2cm]{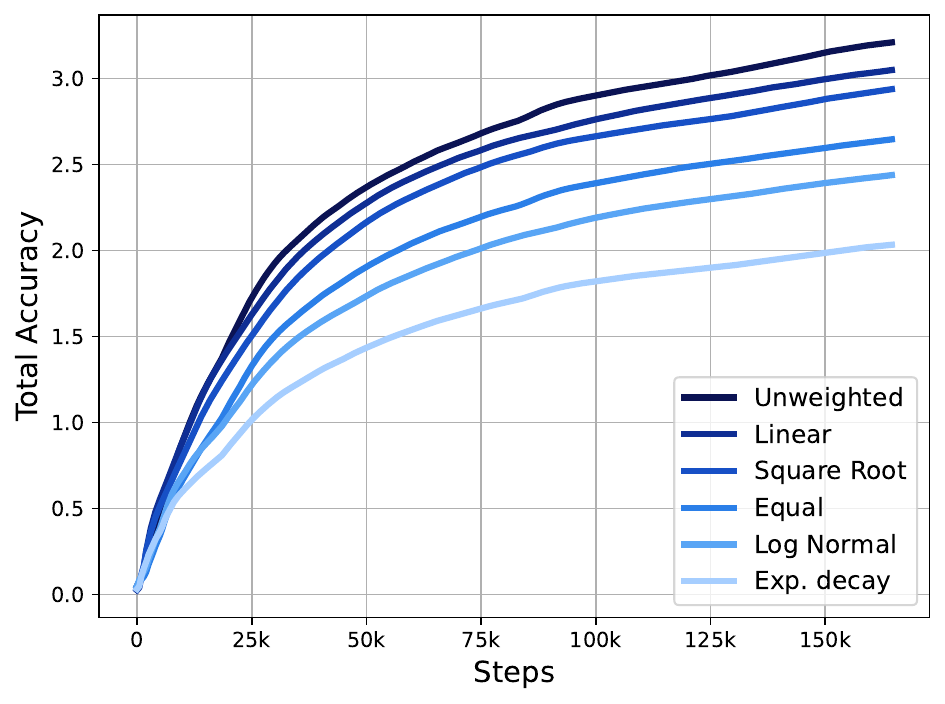}
    \caption{\textbf{Total Accuracy for Different Weighting Functions}}
    \label{fig:enter-label}
\end{figure}
\begin{figure*}
    \centering
        \includegraphics[scale=0.23]{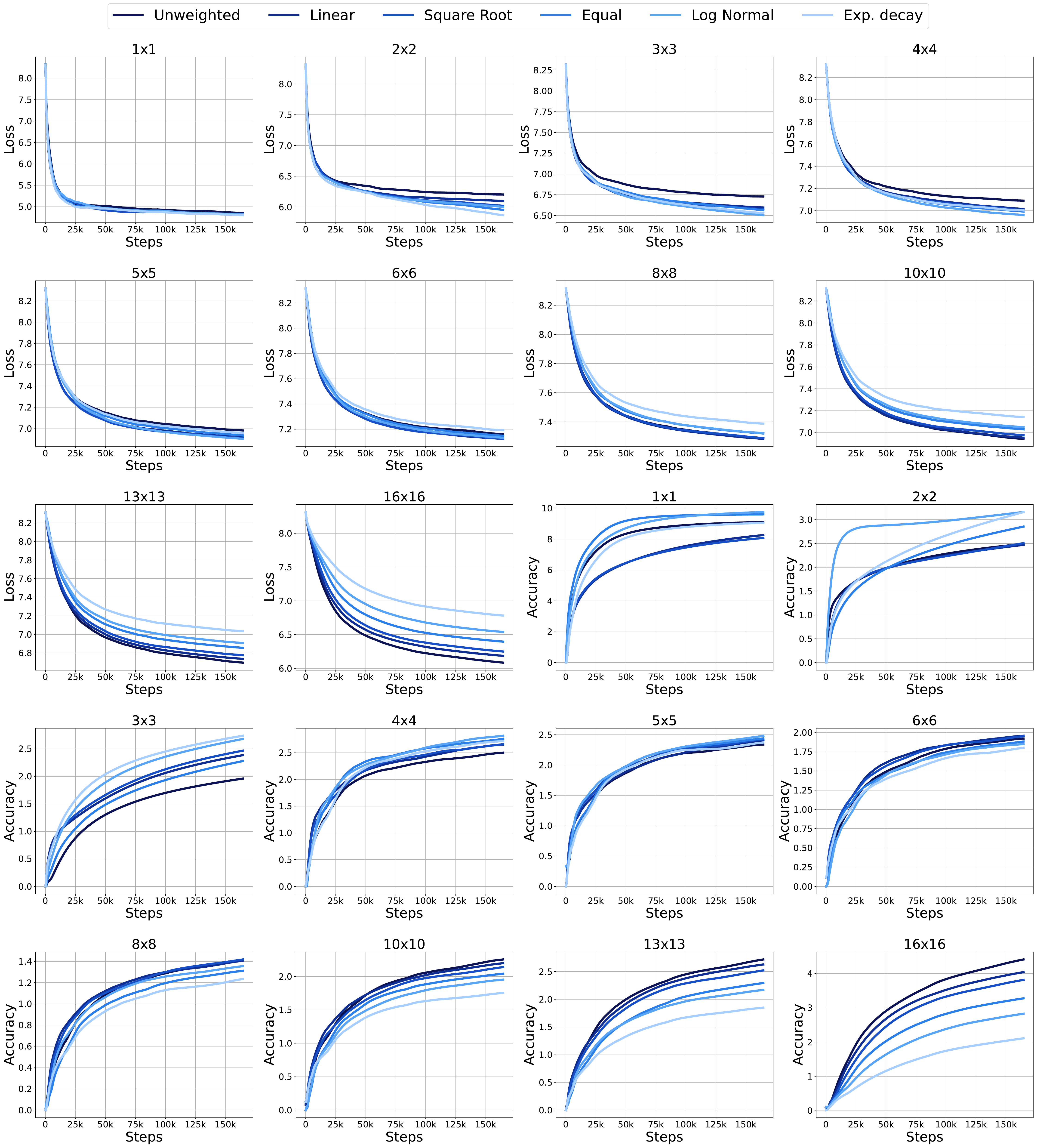}
    \caption{\textbf{Scale-wise Decomposition of Loss and Accuracy}. We analyze the impact of loss weighting across scales on generated image quality.  By decomposing both loss and accuracy metrics scale-by-scale, we reveal a key insight: prioritizing performance in early and intermediate scales, rather than simply minimizing total loss, leads to improved perceptual quality.  This is evidenced by the correlation between strong early/mid-scale performance and superior FID and Inception Scores, as detailed in Table \ref{table:loss-reweighting}. This suggests that while later scales contribute significantly to the overall loss due to their higher token count, they are less critical for generating high-fidelity images.}
    \label{fig:loss-acc-decomposition}
\end{figure*}

\clearpage
\section{Masking/ Parallel Sampling}
In this section, we start by demonstrating how parallel sampling in VAR can result in error accumulation during the image generation process. We then delve into our masked image reconstruction objective and discuss masked sampling. Finally, we present ablations to illustrate how intra-scale masking improves the generation quality. 
\subsection{Error Accumulation in Parallel Sampling}
\label{sec:error-accumulation}
\begin{figure}[H]
    \centering
    \includegraphics[scale=0.5]{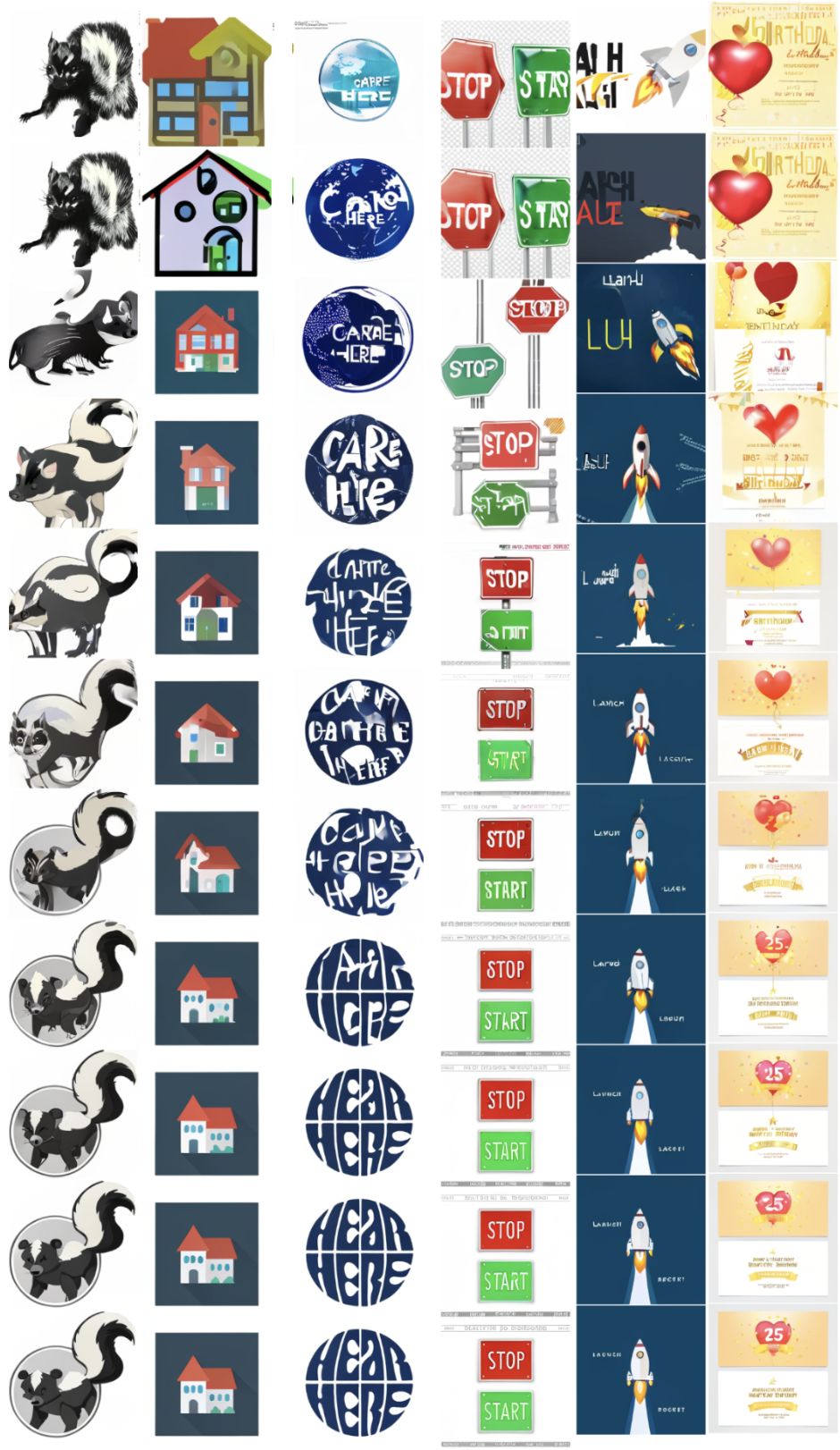}
    \caption{\textbf{Error Accumulation in Next-Scale Prediction}: Teacher forcing experiment showing error propagation across scales. The top rows show image generation starting from the ground truth at earlier scales, showing lower quality. The bottom rows show image generation starting from later scales, leading to improved quality, indicating error accumulation from earlier scales impacts quality. Model: VAR-d16.}
    \label{fig:error-accumulation}
\end{figure}
We illustrate error propagation in image generation using a VAR-d16 model through a teacher-forcing experiment. Ground truth tokens are provided at each scale, and the model generates the rest of the image beginning from that scale. As shown in Fig. \ref{fig:error-accumulation}, starting from earlier scales results in poorer quality, while later scales yield better images. This suggests that to get good image quality, it is important to get the earlier scales correctly. We hypothesize that errors introduced during parallel token sampling at earlier scales propagate during the generation process and impact the overall image quality.

\subsection{Image Reconstruction}
\begin{figure}[H]
    \centering
    \includegraphics[scale=0.5]{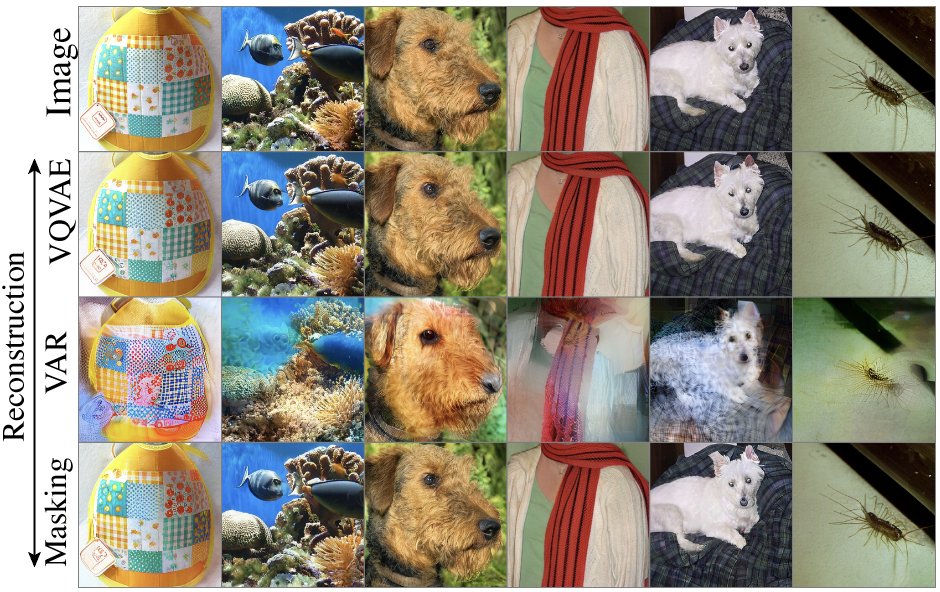}
    \caption{\textbf{Image Reconstruction} in VQ-VAE, VAR and HMAR}
    \label{fig:image-reconstruction}
\end{figure}
Fig.~\ref{fig:image-reconstruction} compares image generation in HMAR and VAR. Our HMAR model can reconstruct images to a quality comparable to the Multi-Scale VQ-VAE. We mask out 50\% of all levels in the image and use HMAR to predict the masked tokens, then combine these tokens across all scales to form a complete image. In contrast, we also demonstrate VAR's image reconstruction. For each image, we provide the ground truth at each level, as done during VAR training, and allow the VAR model to predict the next scale.

\begin{figure*}
    \centering
    \includegraphics[width=0.98\linewidth]{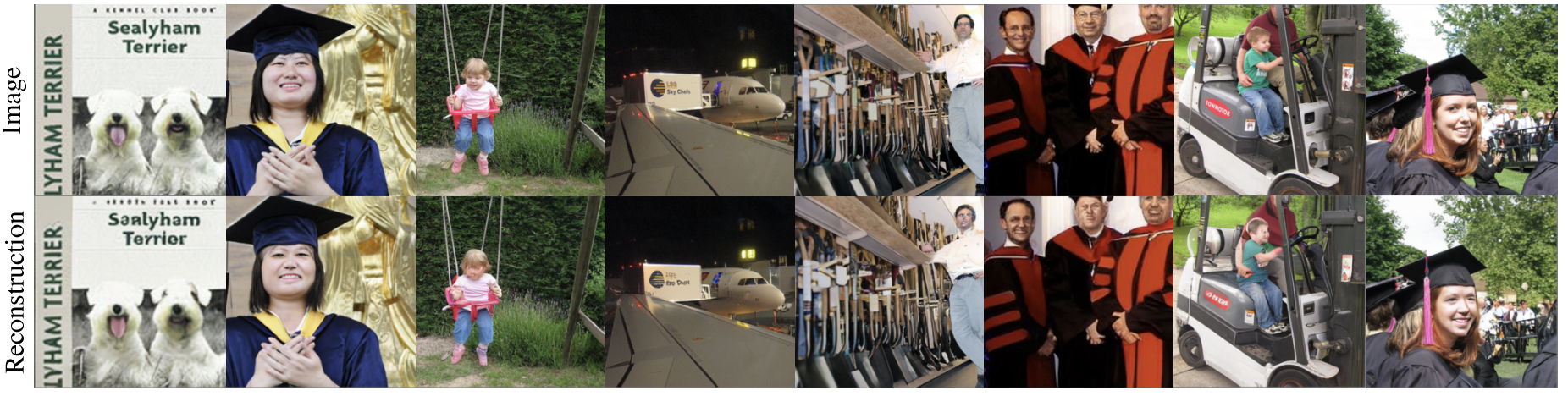}
    \caption{\textbf{Multi-Scale VQ-VAE~\citep{tian2024visual} Reconstructions}: We show some failure cases for the Multi-Scale VQ-VAE \citep{tian2024visual}. While it can capture general image structures, it struggles to reconstruct fine-grained details, particularly in complex elements such as text and facial features.}
    \label{fig:vqvae-failure}
\end{figure*}

\subsection{Masked Finetuning}
We find masked prediction significantly easier to learn than next-scale prediction, reaching accuracies of 65+\% compared to just 5\% in next-scale prediction (Fig.~\ref{fig:masking-loss}), allowing for efficient fine-tuning of a masked prediction layer atop a pre-trained next-scale model. Unlike in next-scale prediction, where we reweight the loss by each scale, we find that for masked prediction, the task is easier to learn, and as such, reweighting the loss to prioritize certain scales can actually hurt performance. We use  $25\%$ to $40\%$ of the pre-training iterations when fine-tuning. In our experiments, we fine-tune specific layers while freezing others, but alternative PEFT methods like LORA could also be used effectively to reduce the number of parameters used. Directly pre-training with both next-scale and masked-prediction objectives is another exciting direction to explore.

\subsection{Masked Sampling}
\begin{figure}[!h]
  \centering
  \vspace{-4mm}
    \includegraphics[scale=0.5]
    {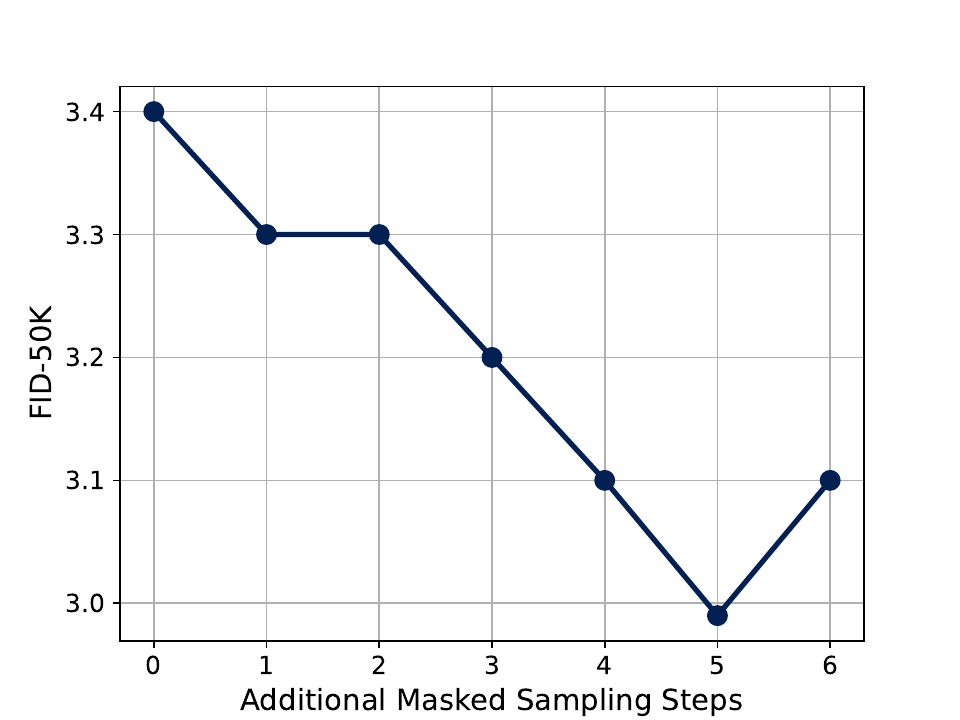}
    \vspace{-2mm}
  \caption{\textbf{Impact of Extra Masked Sampling Steps on FID for HMAR-$d16$.} Increasing the number of sampling steps at earlier scales leads to better FID.
  \vspace{-2mm}}
  \label{fig:masking-quantitative}
\end{figure}

\begin{figure}[!h]
  \centering
  \vspace{-4mm}
    \includegraphics[scale=0.385]
    {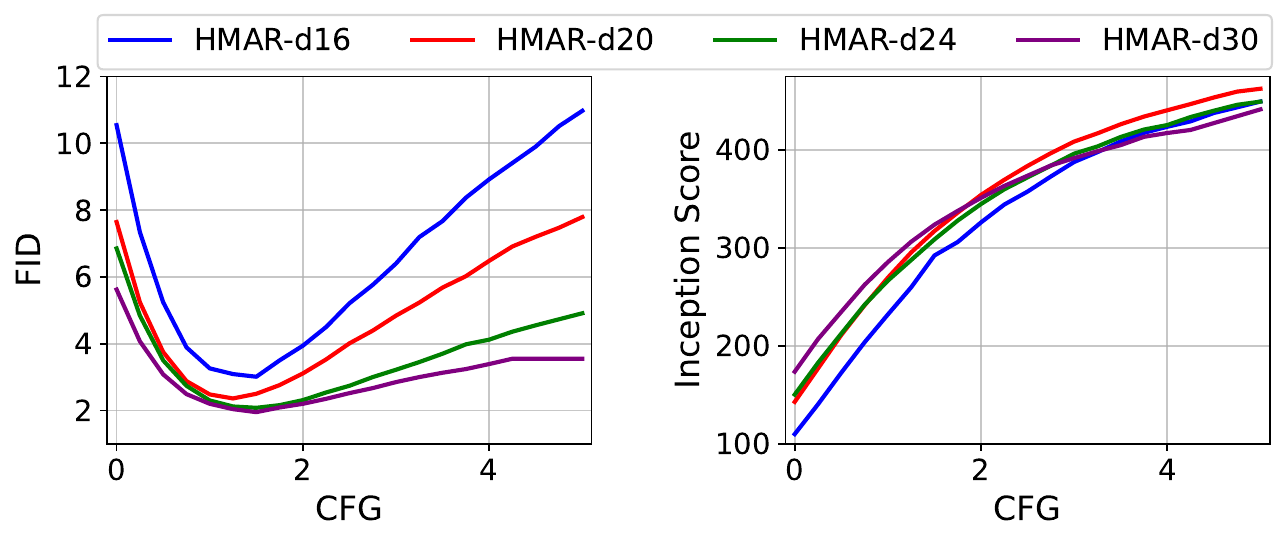}
    \vspace{-2mm}
  \caption{\textbf{Classifier-free guidance} impact on FID and Inception Score}
  \label{fig:cfg_fid_is}
\end{figure}

\begin{figure*}[t]
\begin{minipage}{\textwidth}
\centering
\includegraphics[scale=0.23]{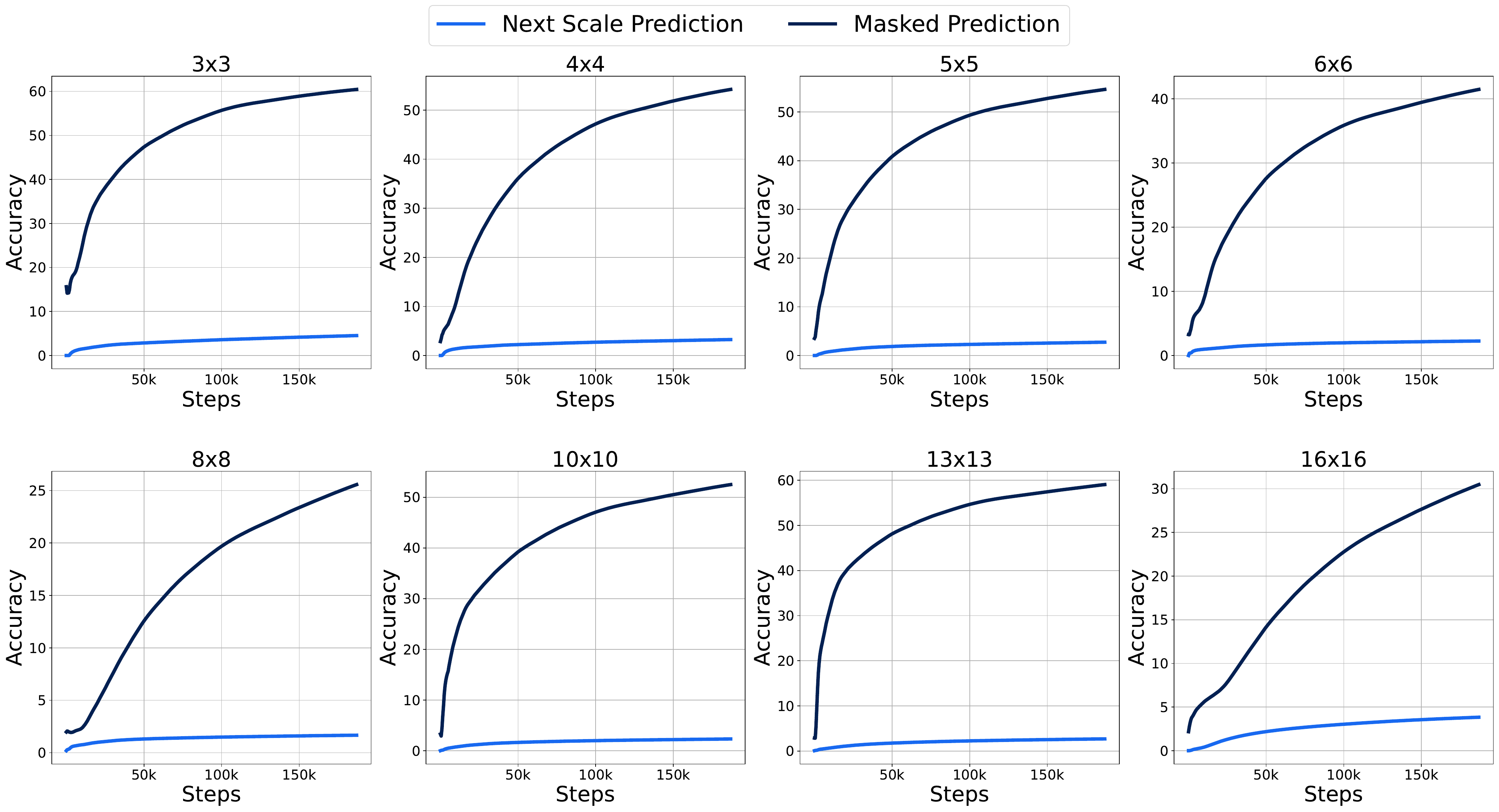}
\caption{\textbf{Comparison of Accuracy Between Next-Scale Prediction and Masked Prediction Across Scales}. Masked prediction on residuals is a considerably simpler task to learn compared to next-scale prediction.}
\label{fig:masking-loss}
\end{minipage}
\begin{minipage}{\textwidth}
\centering
\includegraphics[scale=0.55]{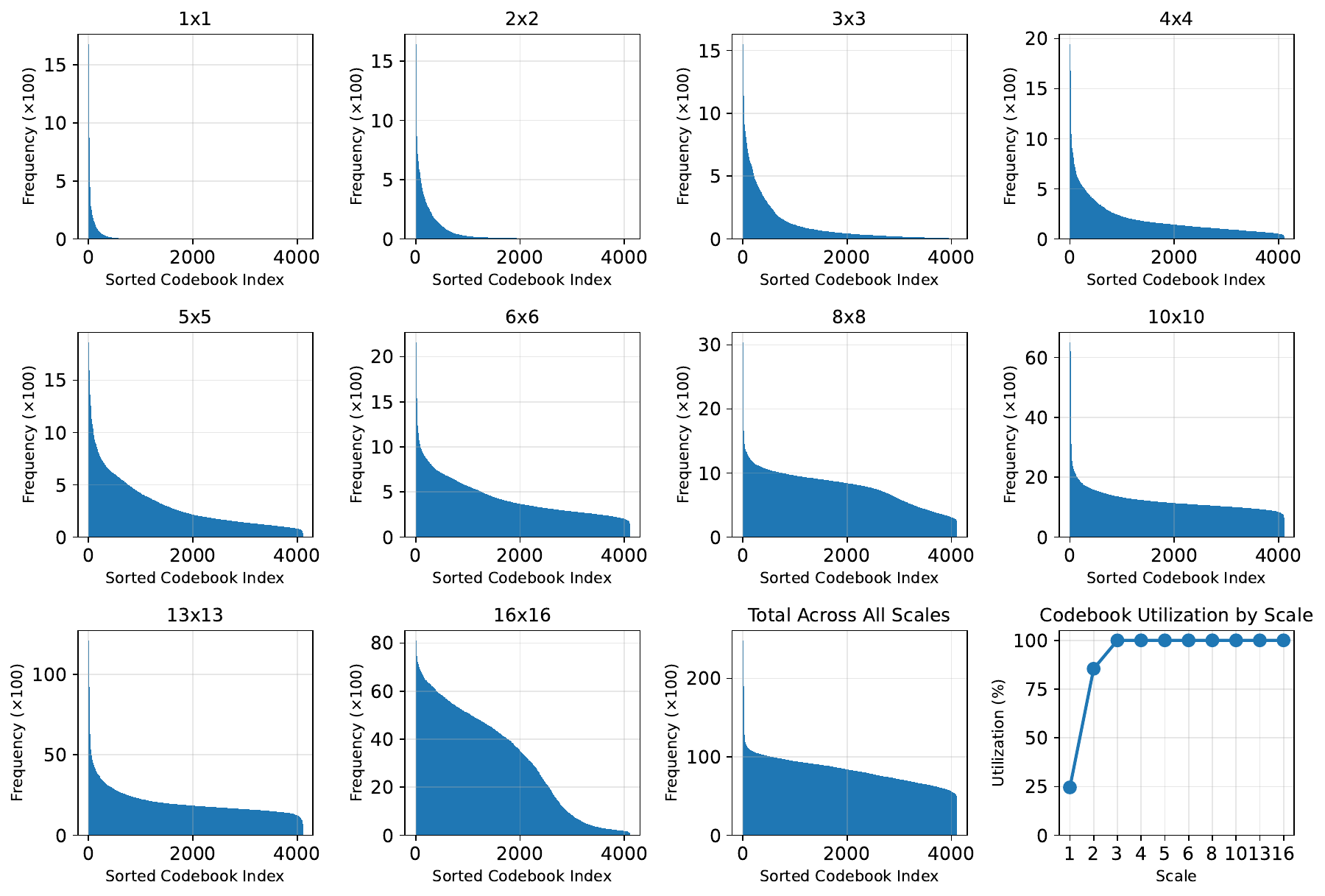}
\caption{\textbf{Analysis of Codebook Usage Patterns Across Scales}. We observe distinct patterns in how codebook entries are utilized: early scales show highly skewed distributions with only a small subset of codes being frequently accessed, indicating potential redundancy, while later scales demonstrate more uniform usage patterns. The codebook utilization rate progressively increases from less than 50\% in early scales to nearly 100\% by scale 4.}
\label{fig:codebook-utilization}
\end{minipage}
\end{figure*}

We run a hyperparameter search to choose the sampling steps and report quantitative metrics in Table \ref{table:imagenet-256} with 14 steps because this yields the best performance. In particular, additional sampling steps at the earlier scales have the highest impact on quantitative metrics. We find that additional sampling steps beyond the first five scales do not improve quantitative metrics but can improve finer details in the image (Fig. \ref{fig:masking-qualitative}). We show how adding one additional sampling step to each scale in HMAR-d16 impacts the FID in Fig.~\ref{fig:masking-quantitative}.
\newline
We use classifier-free guidance when sampling and show its effect on FID and Inception score in Fig.~\ref{fig:cfg_fid_is}.

\section{Discussion on Multi-Scale VQ-VAE Tokenizer}
\label{sec:tokenizer-failure}
We adopt the Multi-scale VQ-VAE tokenizer from VAR \citep{tian2024visual}. In this section, we provide more details and highlight some of its limitations and possible ways to improve it.

\subsection{Image Reconstruction}
We provide PSNR and rFID values for the tokenizer at different image resolutions in Table \ref{table:vqvae-metrics}.
\begin{table}[htbp]
    \centering
    \begin{tabular}{|c|c|c|}
        \hline
        \textbf{Resolution} & \textbf{rFID} & \textbf{PSNR (dB)} \\
        \hline
        256$\times$256 & 0.92 & 20.69 \\
        \hline
        512$\times$512 & 0.66 & 21.74 \\
        \hline
    \end{tabular}
    \caption{\textbf{PSNR and rFID} values for different image resolutions}
    \label{table:vqvae-metrics}
\end{table}
In Fig.~\ref{fig:vqvae-failure}, we show the tokenizer's inability to capture fine-grained details within images. Recent works like HART \citep{tang2024hart} propose a hybrid tokenization scheme, incorporating continuous tokens to better represent high-frequency details. These contributions are complementary to our work.

\subsection{Codebook Utilization}
In Fig.~\ref{fig:codebook-utilization}, we analyze the distribution of codebook usage across different scales. The overall codebook usage appears relatively uniform; however, early scales, which capture coarse structural features, exhibit highly skewed distributions, with only a small subset of codebook entries being used. In contrast, later scales that capture fine-grained details demonstrate a more uniform distribution of code usage. Future tokenizer designs could benefit from an asymmetric approach: smaller, specialized codebooks for early scales to efficiently capture essential structural features and larger, more diverse codebooks for later scales to accommodate the broader range of local details. 
\clearpage
\section{Additional Qualitative Results}
\label{sec:additional-qualitative-results}
\subsection{Qualitative Comparisons}
\begin{minipage}{0.98\textwidth}
  \strut\newline
  \centering
\setlength{\tabcolsep}{1pt}
\begin{tabular}{cccc}
    ~~~~~~~~~~~~~~MaskGIT ~~~~~~~&~~~~~~~~~~~~~~~~~~~~~~~~~~~~~DiT &~~~~~~~~~~~~~~~~~~~~~~~~~~~~~~~~~~~~~~~VAR & ~~~~~~~~~~~~~~~~~~~~HMAR \\
    \multicolumn{4}{c}{\includegraphics[width=0.925\textwidth, height=0.375\textwidth]{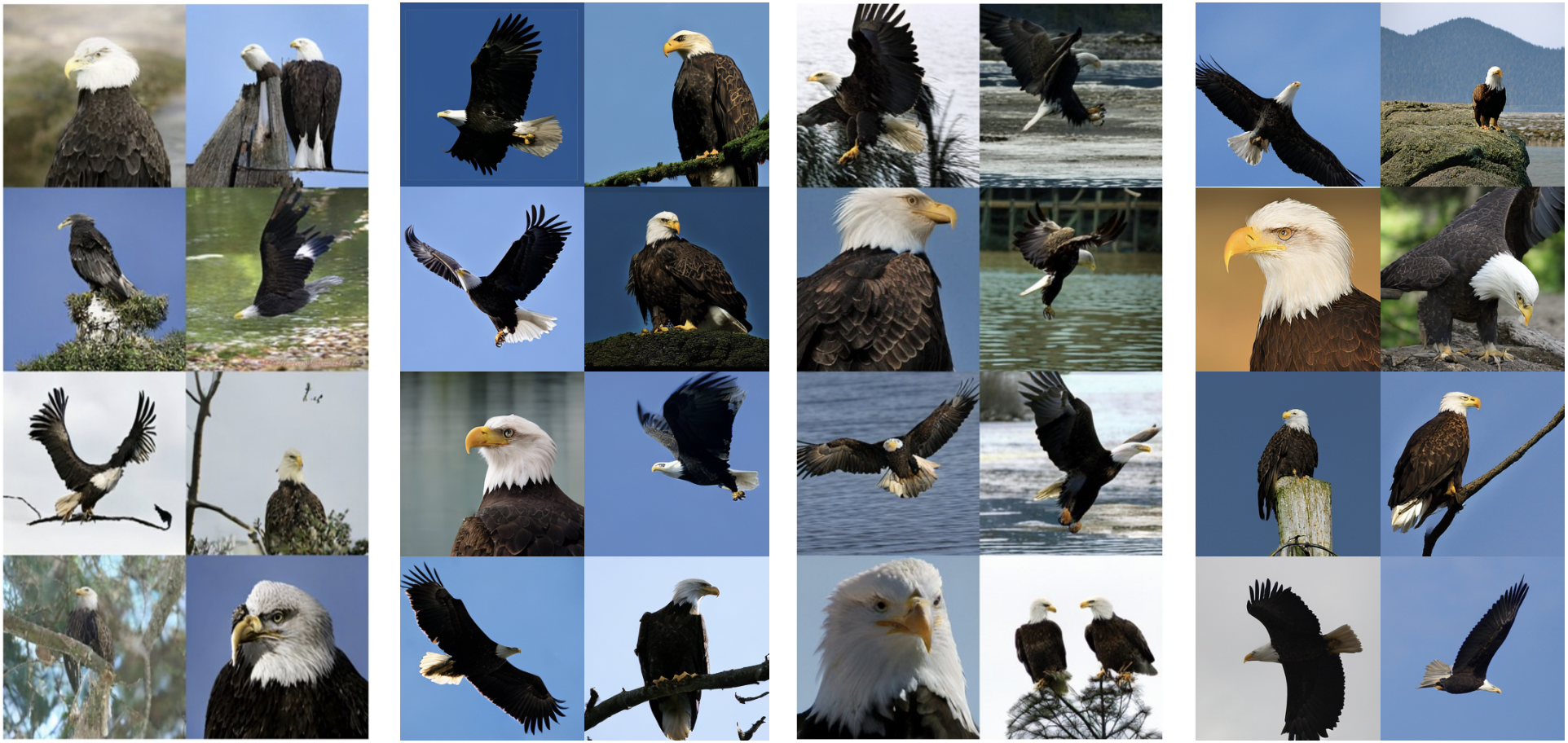}} \\
    \multicolumn{4}{c}{\includegraphics[width=0.925\textwidth, height=0.375\textwidth]{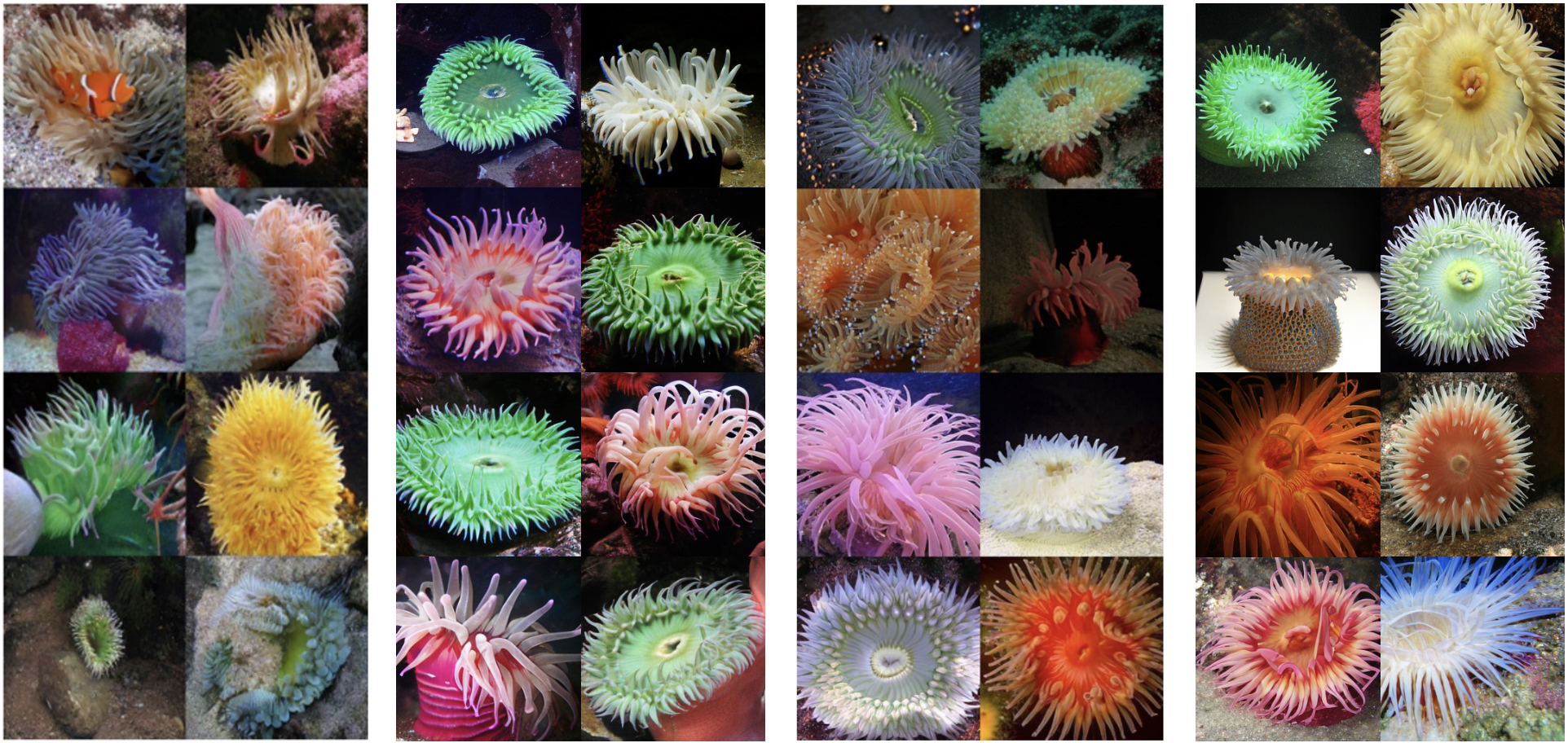}} \\
    \multicolumn{4}{c}{\includegraphics[width=0.925\textwidth, height=0.375\textwidth]{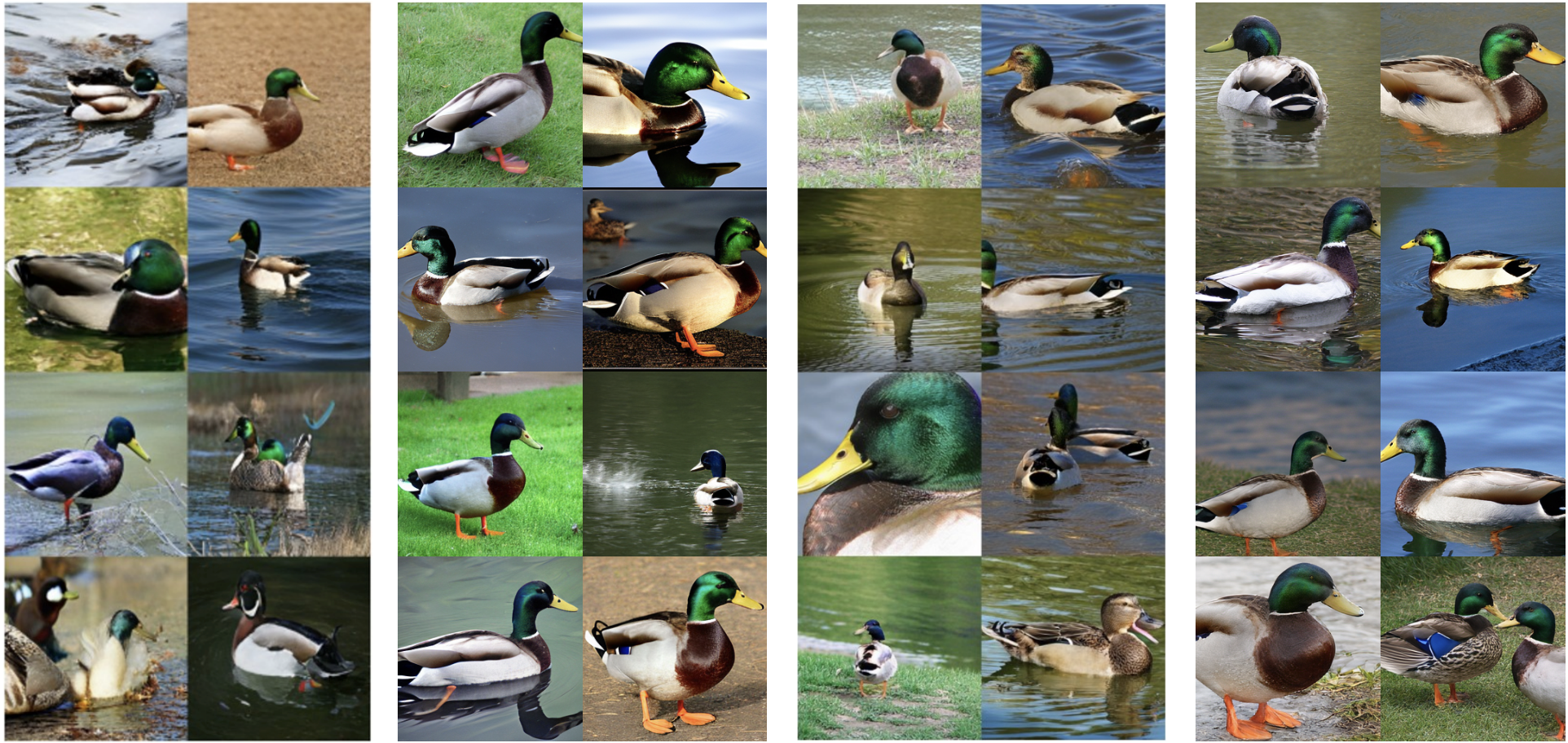}} \\
\end{tabular}
  \captionof{figure}{\textbf{Qualitative Comparisons on ImageNet $256\times256$}}\label{fig:figure1}
    \label{fig:additional-examples-3}
\end{minipage}
\clearpage
\subsection{Class Conditional ImageNet 256x256 Samples}



\begin{minipage}{0.99\textwidth}
  \strut\newline
  \centering
\setlength{\tabcolsep}{1pt}
\begin{tabular}{cc}
    \includegraphics[width=0.45\textwidth, height=0.35\textwidth]{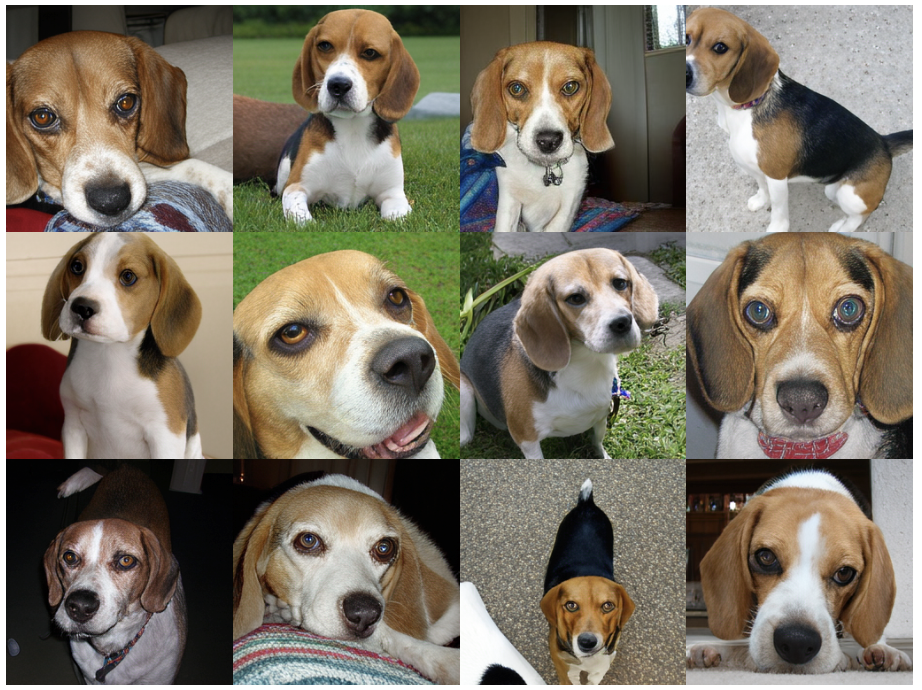} & 
    \includegraphics[width=0.45\textwidth, height=0.35\textwidth]{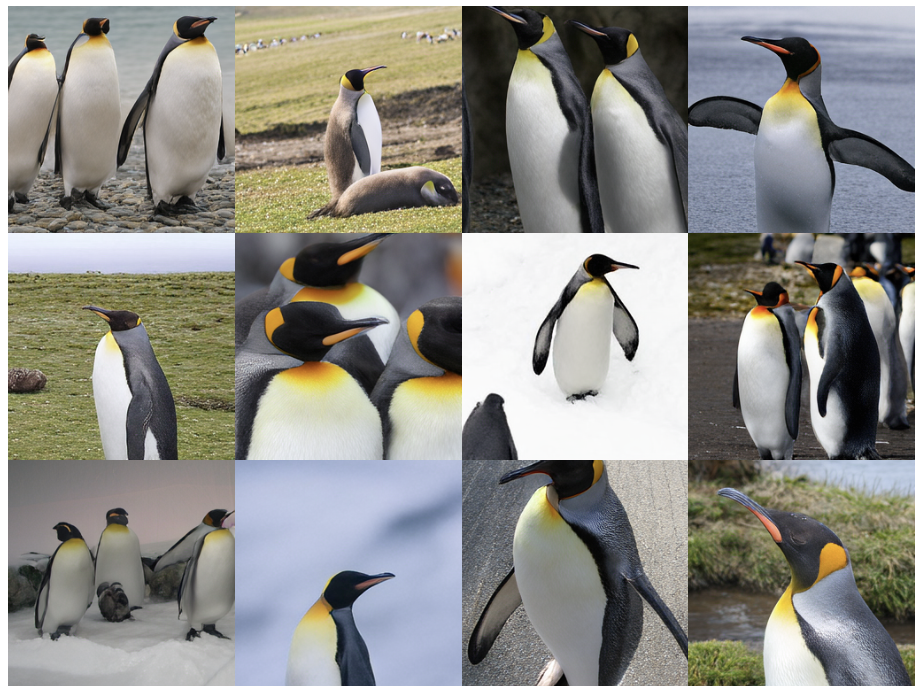} \\
    \small Class ID $162$, Beagle & \small Class ID $145$, Penguin \\[6pt]
    
    \includegraphics[width=0.45\textwidth, height=0.35\textwidth]{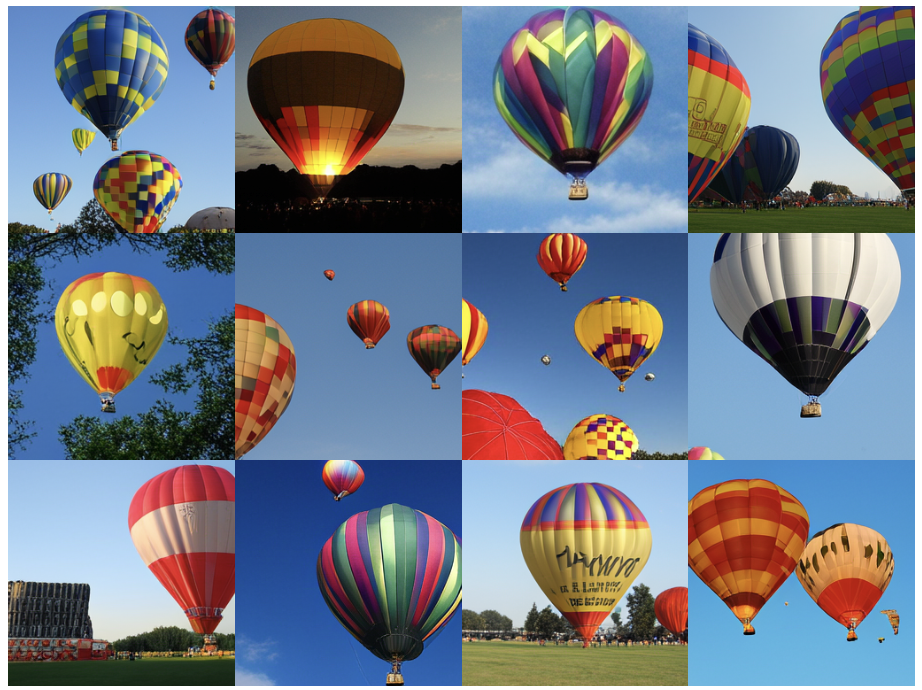} & 
    \includegraphics[width=0.45\textwidth, height=0.35\textwidth]{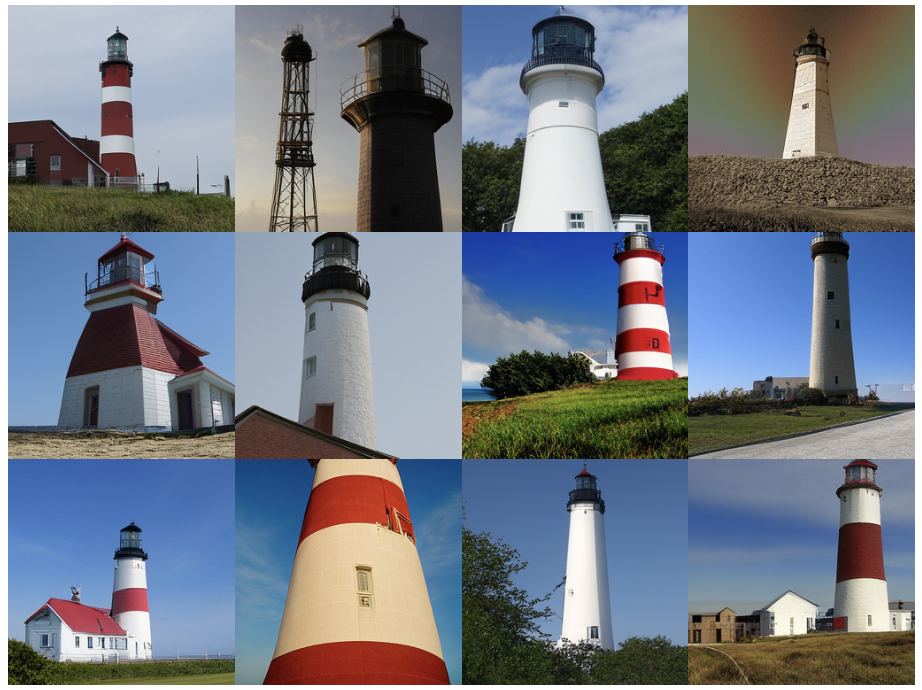} \\
    \small Class ID $417$, Balloon & \small Class ID $437$, Beacon\\[6pt]

    \includegraphics[width=0.45\textwidth, height=0.35\textwidth]{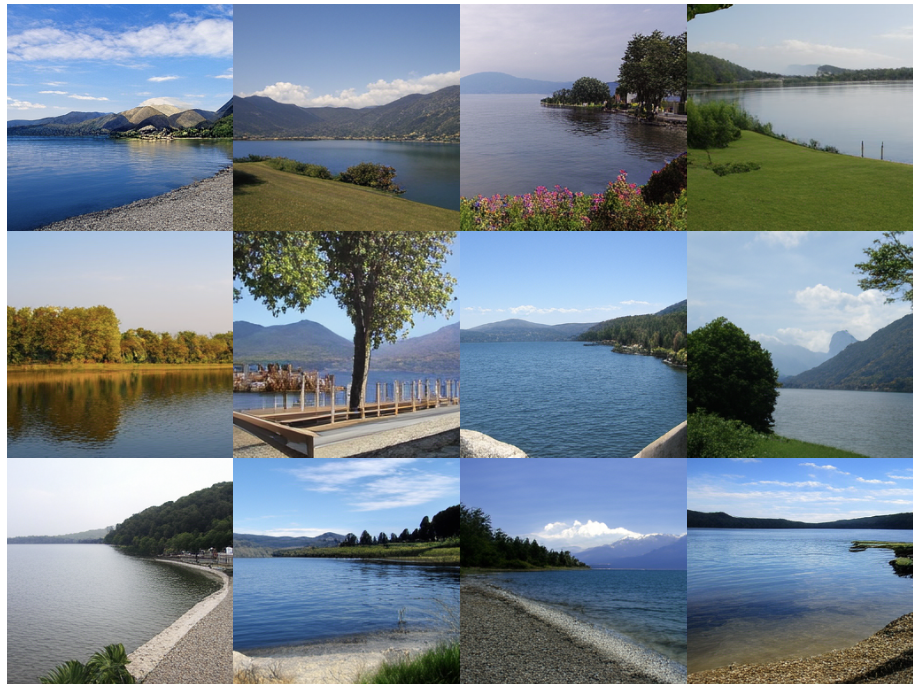} & 
    \includegraphics[width=0.45\textwidth, height=0.35\textwidth]{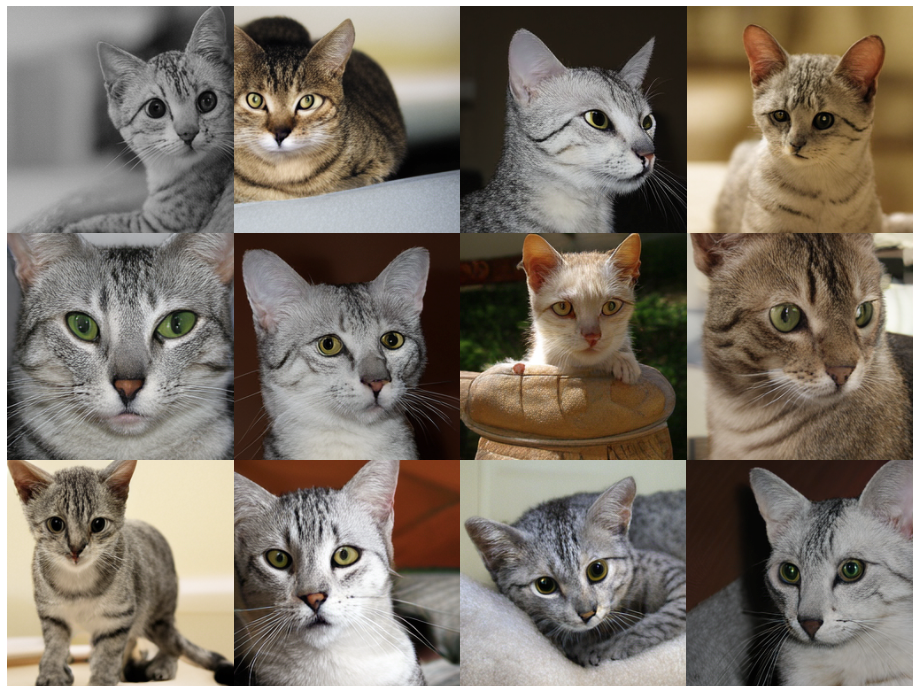} \\
    \small Class ID $975$, Lakeside & \small Class ID $285$, Egyptian Cat\\[6pt]
\end{tabular}
\vspace{-0.35cm}
  \captionof{figure}{\textbf{Additional Class-Conditional Image Generation Samples on ImageNet $256\times256$}}\label{fig:figure1}
    \label{fig:additional-examples-3}
\end{minipage}

    


\clearpage
\begin{minipage}{0.99\textwidth}
  \strut\newline
  \centering
\setlength{\tabcolsep}{1pt}
\begin{tabular}{cc}
    \includegraphics[width=0.45\textwidth, height=0.35\textwidth]{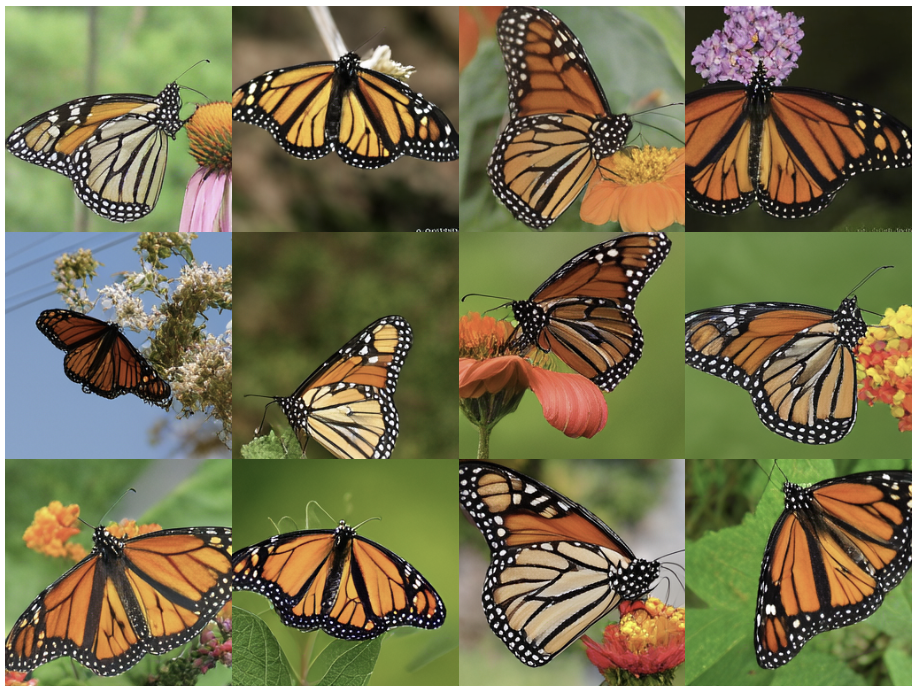} & 
    \includegraphics[width=0.45\textwidth, height=0.35\textwidth]{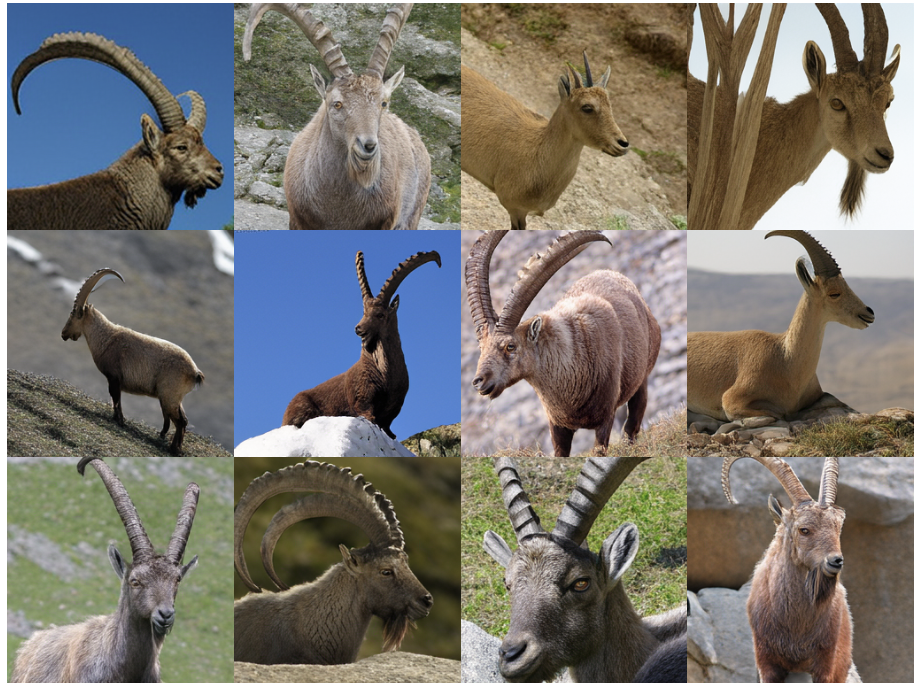} \\
    \small Class ID $323$, Monarch & \small Class ID $350$, Ibex\\[6pt]
    
    \includegraphics[width=0.45\textwidth, height=0.35\textwidth]{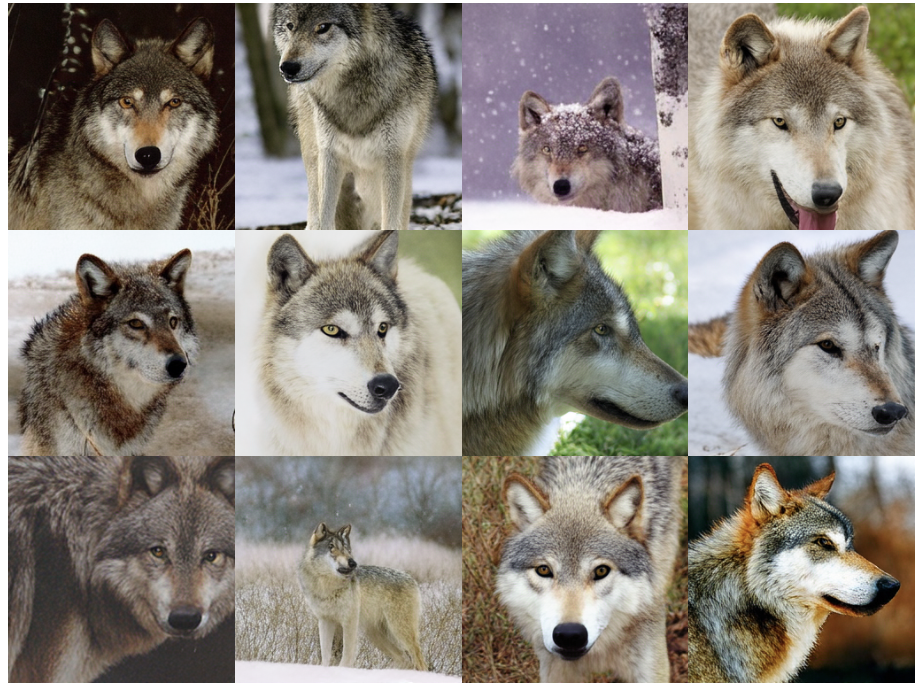} & 
    \includegraphics[width=0.45\textwidth, height=0.35\textwidth]{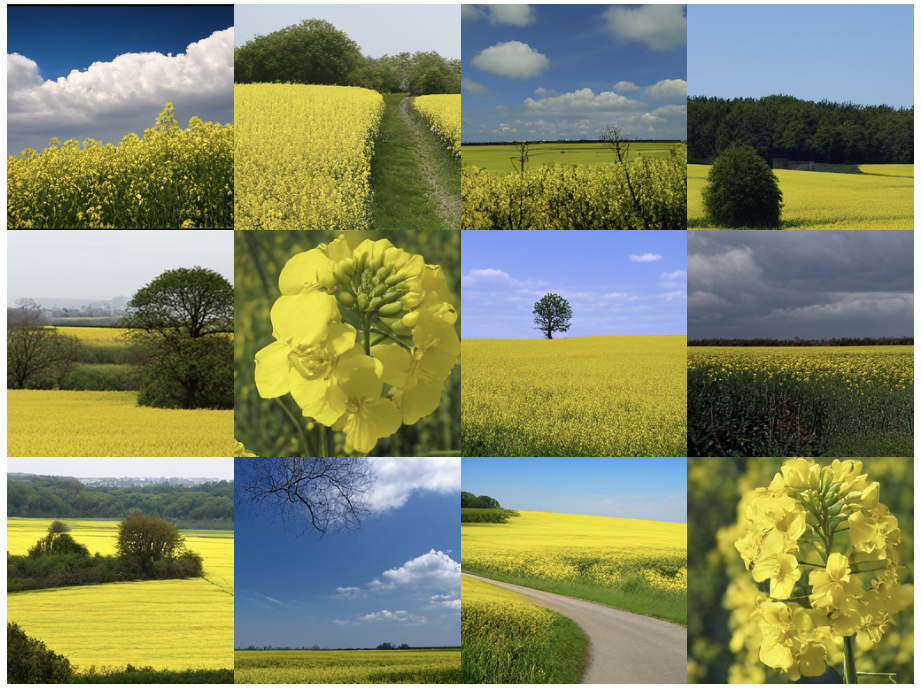} \\
    \small Class ID $269$, Timber Wolf & \small Class ID $984$, Rapeseed\\[6pt]

    \includegraphics[width=0.45\textwidth, height=0.35\textwidth]{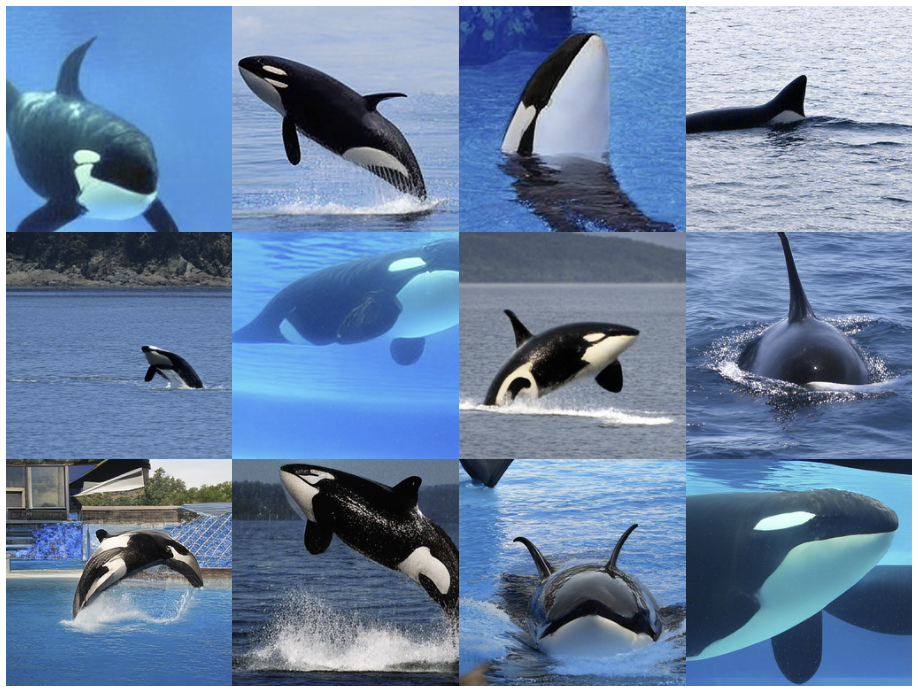} & 
    \includegraphics[width=0.45\textwidth, height=0.35\textwidth]{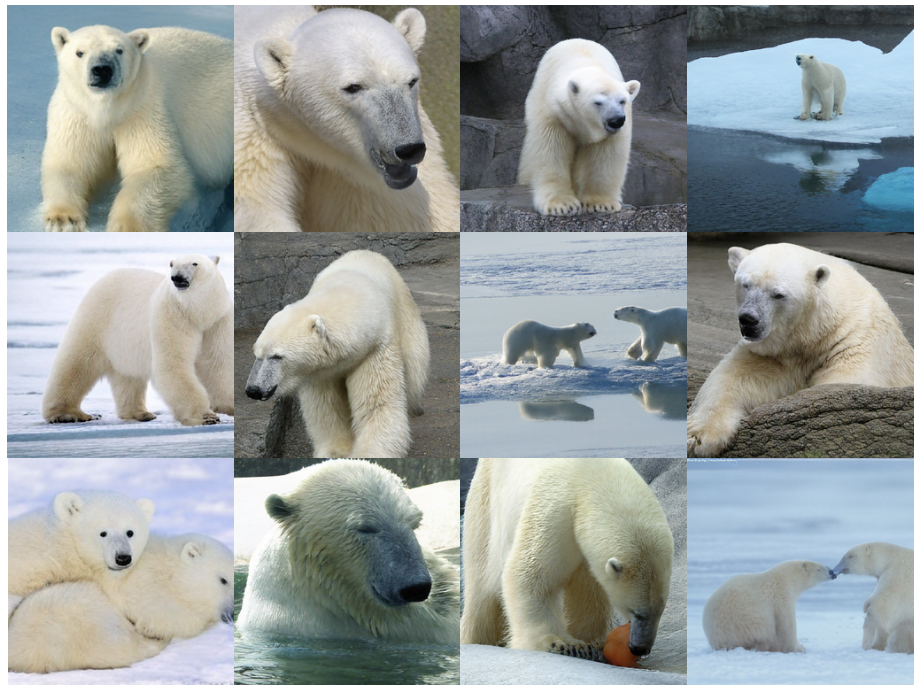} \\
    \small Class ID $148$, Killer Whale & \small Class ID $296$, Ice Bear\\[6pt]
\end{tabular}
  \captionof{figure}{\textbf{Additional Class-Conditional Image Generation Samples on ImageNet $256\times256$}}\label{fig:figure1}
    \label{fig:additional-examples-3}
\end{minipage}

\clearpage
\subsection{Class Conditional ImageNet 512x512 Samples}
    
    
\begin{minipage}{0.985\textwidth}
  \strut\newline
  \centering
\setlength{\tabcolsep}{1pt}
\begin{tabular}{cc}
    \includegraphics[width=0.45\textwidth, height=0.43\textwidth]{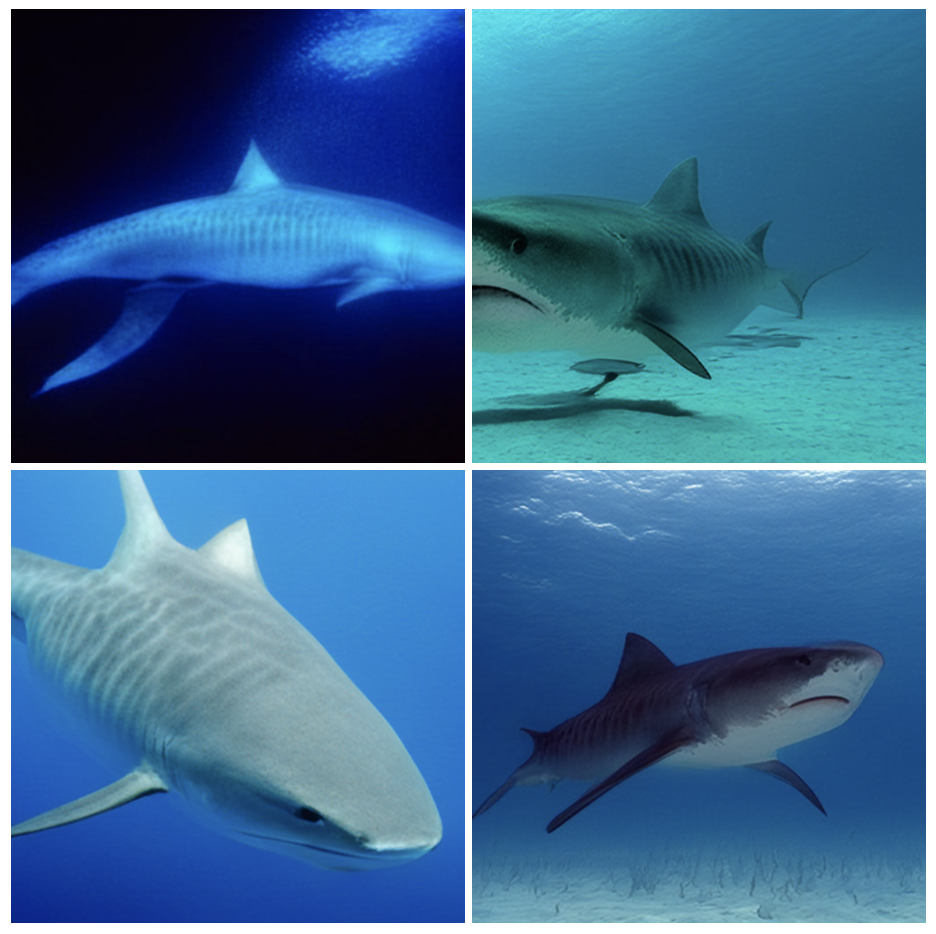} & 
    \includegraphics[width=0.45\textwidth, height=0.43\textwidth]{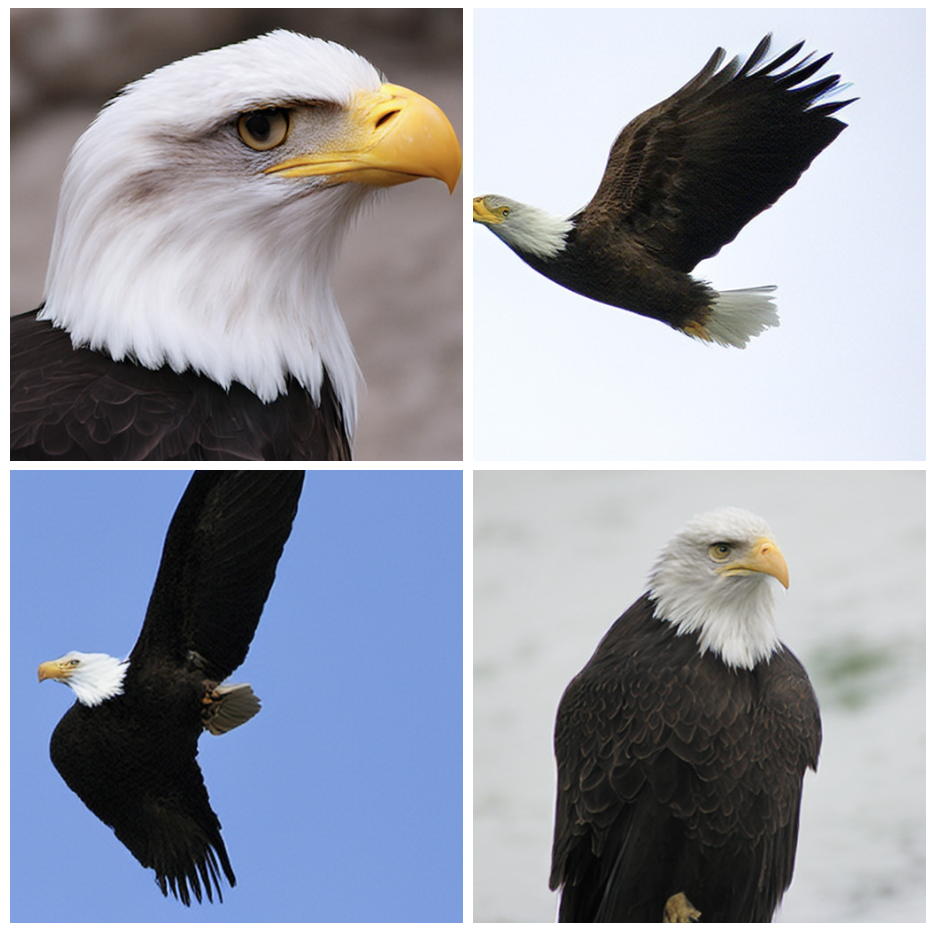} \\
    \small Class ID $3$, Shark & \small Class ID $22$, Bald Eagle \\[6pt]
    
    \includegraphics[width=0.45\textwidth, height=0.43\textwidth]{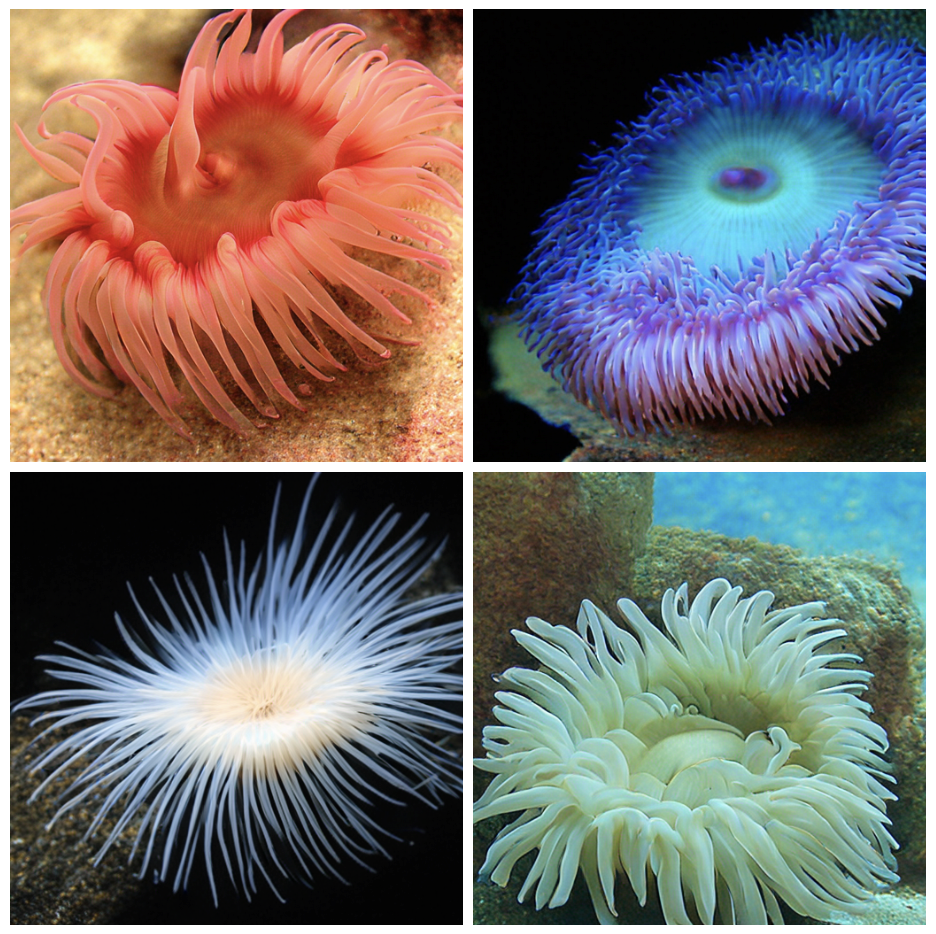} & 
    \includegraphics[width=0.45\textwidth, height=0.43\textwidth]{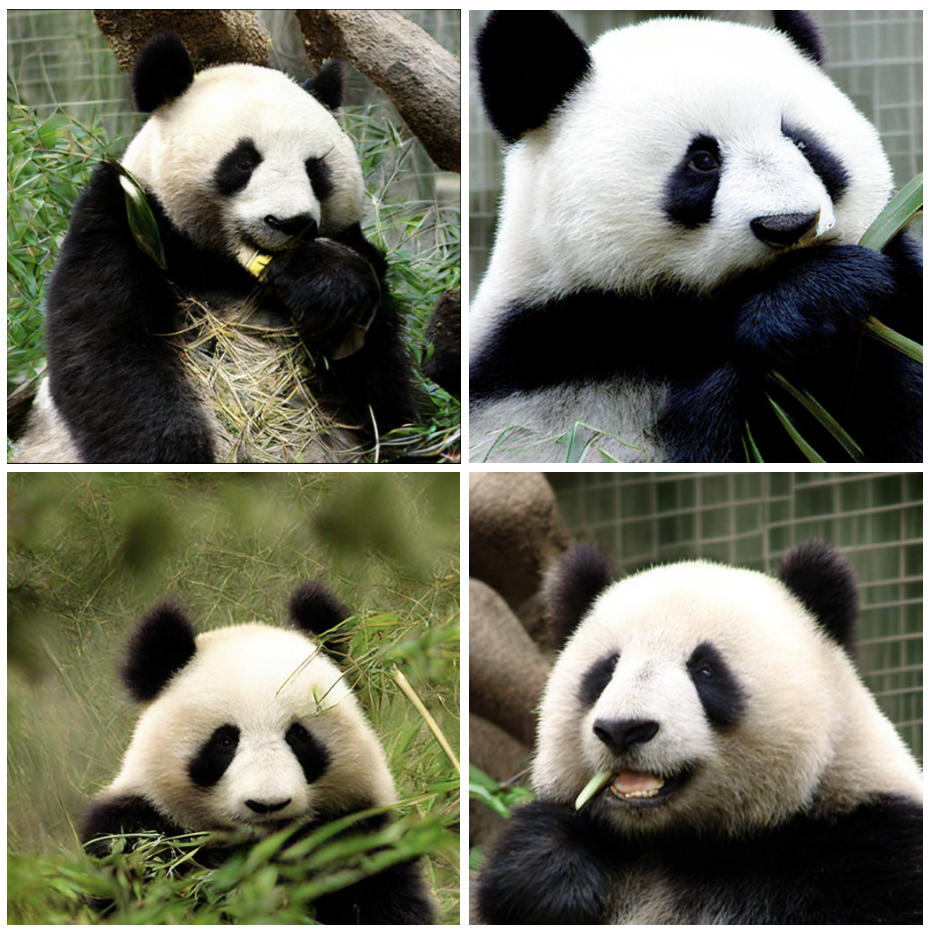} \\
    \small Class ID $108$, Sea Anemone & \small Class ID $388$, Giant Panda\\[6pt]
    
    {\includegraphics[width=0.45\textwidth, height=0.23\textwidth]{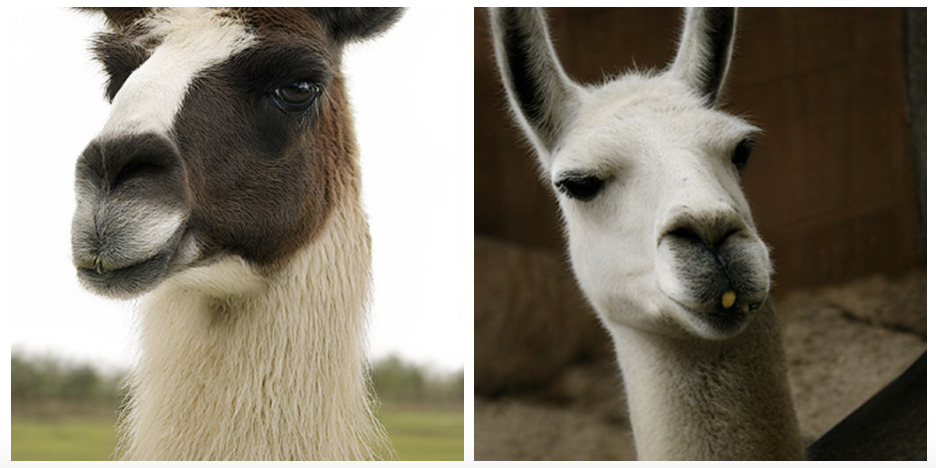}} &
     {\includegraphics[width=0.45\textwidth, height=0.23\textwidth]{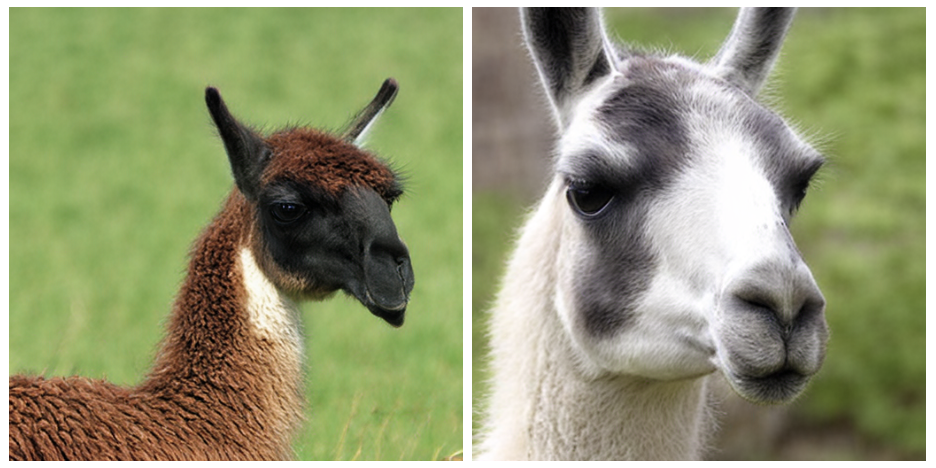}}\\
    \multicolumn{2}{c}{\small Class ID $355$, Llama}
\end{tabular}
  \captionof{figure}{\textbf{Additional Class-Conditional Image Generation Samples on ImageNet $512\times512$}}\label{fig:figure1}
    \label{fig:additional-examples-3}
\end{minipage}

\stopcontents[appendix]  

\end{document}